%% file: main_arxiv.tex
\documentclass{article}

\usepackage{arxiv}

\usepackage{amsfonts}
\usepackage{amsmath,bm}
\usepackage{amssymb}
\usepackage{lineno}
\usepackage{tabularx}
\usepackage{multirow}
\usepackage{multicol}
\usepackage{subfig}
\usepackage{graphicx}
\usepackage{caption}
\usepackage{tikz}
\usepackage{color}
\usepackage{xspace}
\usepackage{enumerate}
\usepackage{xurl}
\usepackage[colorlinks,allcolors=blue]{hyperref}
\usepackage{xspace}
\usepackage[normalem]{ulem}

\newcolumntype{L}{>{\raggedright\arraybackslash}X}
\newcolumntype{C}{>{\centering\arraybackslash}X}
\newcolumntype{R}{>{\raggedleft\arraybackslash}X}
\newcommand{\highest}[1]{{\mathbf{#1}}}%
\newcommand{\sent}{optical\xspace}
\newcommand{\weat}{weather\xspace}
\newcommand{\dem}{DEM\xspace}
\newcommand{\soil}{soil\xspace}
\newcommand{\sentsource}{S2\xspace}
\newcommand{\weatsource}{ERA5\xspace}
\newcommand{\demsource}{SRTM\xspace}
\newcommand{\soilsource}{SoilGrids\xspace}
\newcommand{\sentraw}{S2-R\xspace}
\newcommand{\sentmonth}{S2-M\xspace}

\newcommand{\lstminputfusion}{LSTM-IF\xspace}
\newcommand{\lgbminputfusion}{GBDT-IF\xspace}
\newcommand{\gatedfusion}{MVGF\xspace}
\newcommand{\GFweights}{gated fusion weights\xspace}
\newcommand{\shortGFweights}{fusion weights\xspace}

\newcommand{\mat}[1]{{\mathbf{#1}}}%
\newcommand{\vect}[1]{{\bm{\mathbf{#1}}}}%

\title{Adaptive Fusion of Multi-view Remote Sensing data for Optimal Sub-field Crop Yield Prediction}

\date{}                                         

\author{Francisco Mena\thanks{both authors contributed equally to this work.}\href{https://orcid.org/0000-0002-5004-6571}{\includegraphics[scale=0.06]{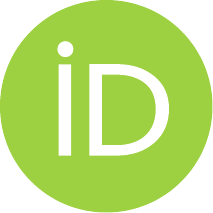}\hspace{1mm}},
        Deepak Pathak\thanks{both authors contributed equally to this work.}\href{https://orcid.org/0009-0004-4105-0466}{\includegraphics[scale=0.06]{imgs/orcid.pdf}\hspace{1mm}}, 
        Hiba Najjar\href{https://orcid.org/0000-0002-7498-794X}{\includegraphics[scale=0.06]{imgs/orcid.pdf}\hspace{1mm}}, 
        Cristhian Sanchez\href{https://orcid.org/0009-0006-3431-9142}{\includegraphics[scale=0.06]{imgs/orcid.pdf}\hspace{1mm}}
        Miro Miranda, 
        Andreas Dengel\href{https://orcid.org/0000-0002-6100-8255}{\includegraphics[scale=0.06]{imgs/orcid.pdf}\hspace{1mm}} \\
        Department of Computer Science,     University of Kaiserslautern-Landau (RPTU), Kaiserslautern, Germany; \\
        SDS, German Research Center for Artificial Intelligence (DFKI), Kaiserslautern, Germany.\\
        \{\texttt{francisco.mena, deepak\_kumar.pathak}\}\texttt{@dfki.de}
        \And
        Jayanth Siddamsetty\href{https://orcid.org/0000-0003-1820-3028}{\includegraphics[scale=0.06]{imgs/orcid.pdf}\hspace{1mm}}, 
        Marlon Nuske\href{https://orcid.org/0000-0002-0651-0664}{\includegraphics[scale=0.06]{imgs/orcid.pdf}\hspace{1mm}}, 
        Marcela Charfuelan\href{https://orcid.org/0009-0005-6886-0415}{\includegraphics[scale=0.06]{imgs/orcid.pdf}\hspace{1mm}}, 
        Diego Arenas\href{https://orcid.org/0000-0001-7829-6102}{\includegraphics[scale=0.06]{imgs/orcid.pdf}\hspace{1mm}}, 
        Michaela Vollmer \\ 
        SDS, German Research Center for Artificial Intelligence (DFKI), Kaiserslautern, Germany.
        \And
        Patrick Helber\href{https://orcid.org/0000-0001-8454-4301}{\includegraphics[scale=0.06]{imgs/orcid.pdf}\hspace{1mm}}, 
        Benjamin Bischke, 
        Peter Habelitz \\
        Vision Impulse GmbH, Kaiserslautern, Germany.
}

\renewcommand{\headeright}{A Preprint}
\renewcommand{\undertitle}{Preprint submitted to Journal}
\renewcommand{\shorttitle}{Adaptive Fusion of Multi-view RS data for Crop Yield Prediction}

\hypersetup{
pdftitle={Adaptive Fusion of Multi-view RS data for Crop Yield Prediction},
pdfauthor={Francisco Mena, Deepak Pathak, Hiba Najjar, Cristhian Sanchez, Patrick Helber, Benjamin Bischke, Peter Habelitz, Miro Miranda, Jayanth Siddamsetty, Marlon Nuske, Marcela Charfuelan, Diego Arenas, Michaela Vollmer, Andreas Dengel},
pdfkeywords={Multi-view Learning, Multi-modal Learning, Data Fusion, Remote Sensing, Deep Learning, Crop Yield Prediction}
}

\newcommand*\citep[1]{\cite{#1}}

\begin{document}

\maketitle

\begin{abstract}
Accurate crop yield prediction is of utmost importance for informed decision-making in agriculture, aiding farmers, industry stakeholders, and policymakers in optimizing agricultural practices. However, this task is complex and depends on multiple factors, such as environmental conditions, soil properties, and management practices. 
Leveraging Remote Sensing (RS) technologies, multi-view data from diverse global data sources can be collected to enhance predictive model accuracy. 
However, combining heterogeneous RS views poses a fusion challenge, like identifying the specific contribution of each view in the predictive task.
In this paper, we present a novel multi-view learning approach to predict crop yield for different crops (soybean, wheat, rapeseed) and regions (Argentina, Uruguay, and Germany).  
Our multi-view input data includes multi-spectral optical images from Sentinel-2 satellites and weather data as dynamic features during the crop growing season, complemented by static features like soil properties and topographic information. 
To effectively fuse the multi-view data, we introduce a Multi-view Gated Fusion (MVGF) model, comprising dedicated view-encoders and a Gated Unit (GU) module. The view-encoders handle the heterogeneity of data sources with varying temporal resolutions by learning a view-specific representation. These representations are adaptively fused via a weighted sum. The \textit{fusion} weights are computed for each sample by the GU using a concatenation of all the view-representations.
The MVGF model is trained at sub-field level with 10 m resolution pixels. Our evaluations show that the MVGF outperforms conventional models on the same task, achieving the best results by incorporating all the data sources, unlike the usual fusion results in the literature.
For Argentina, the MVGF model achieves an $R^2$ value of 0.68 at sub-field yield prediction, while at the field level evaluation (comparing field averages), it reaches around 0.80 across different countries. 
The GU module learned different weights based on the country and crop-type, aligning with the variable significance of each data source to the prediction task.
This novel method has proven its effectiveness in enhancing the accuracy of the challenging sub-field crop yield prediction. Our investigation indicates that the gated fusion approach promises a significant advancement in the field of agriculture and precision farming.
\end{abstract}

\keywords{Multi-view Learning \and Multi-modal Learning \and Data Fusion \and Remote Sensing \and Deep Learning \and Crop Yield Prediction}

\input{content/introduction}
\input{content/sota}
\input{content/data}
\input{content/methods}

\input{content/experiments}

\input{content/analysis}
\input{content/conclusion}

\section*{Abbreviations}
\noindent 
\begin{minipage}[t]{0.5\textwidth}\vspace{0pt}
\begin{tabular}{ll}
BN & Batch-Normalization \\ 
CNN & Convolutional Neural Network \\
DEM & Digital Elevation Model \\
GU & Gated Unit \\
LSTM & Long-Short Term Memory \\
MLP & Multi-Layer Perceptron \\
ML & Machine Learning \\
\end{tabular}
\end{minipage} \quad\hfill
\begin{minipage}[t]{0.5\textwidth}\vspace{0pt}
\begin{tabular}{ll}
MVL & Multi-View Learning \\
RS & Remote Sensing \\
RNN & Recurrent Neural Network \\
SCL & Scene Classification Layer \\
SITS & Satellite Image Time Series \\
S2 & Sentinel-2 \\
\end{tabular}
\end{minipage}

\section*{Acknowledgments}
The research results presented are part of a large collaborative project on agricultural yield predictions, which was partly funded through the ESA InCubed Programme (\url{https://incubed.esa.int/}) as part of the project AI4EO Solution Factory (\url{https://www.ai4eo-solution-factory.de/}). F. Mena and H. Najjar acknowledge support through a scholarship of the University of Kaiserslautern-Landau.
The research carried out in this work was partly funded by companies, BASF Digital Farming, John Deere, and MunichRe.

\section*{Data availability}
The authors do not have the permission to share data or code.

\bibliographystyle{apalike} 
\bibliography{content/refs}

\clearpage
\setcounter{table}{0}
\renewcommand{\thetable}{A\arabic{table}}
\appendix
\input{content/appendix.tex}

\end{document}

%% file: content/introduction.tex
\section{Introduction} \label{sec:introduction}

Phenomena observed on the Earth have complex interactions, and hence their observation requires a multi-faceted measurement approach. For instance, the development of a farm crop is affected by human practices, weather conditions, soil structure, and other aspects.
To cover some of these interacting factors, the availability of diverse and rapidly increasing Remote Sensing (RS) sources \citep{camps2021deep} has enabled multiple observations for an object of study.
In the machine learning domain, this scenario is called Multi-View Learning (MVL; \cite{yan2021deep}). However, in RS-based applications, the views can be rather heterogeneous.
The differences in spatial and temporal resolution could be significant, 
and establishing the complementary and supplementary information between sensors for a predictive task is a non-trivial problem.

In this paper, we focus on the challenging machine learning (ML) task of multi-view crop yield prediction at sub-field level. We focus on the use of temporal data from seeding to harvesting, encompassing the entire crop growing season. Notably, the beginning and end of the growing season depend on several factors, including the region, crop-type and farmer's practices.  
Furthermore, the crop productivity can be influenced by, but not limited to, environmental conditions, soil properties,  and management practices.
Therefore, the effectiveness of the predictive models relies on how well it combines the task-related information from the multi-view data.
The standard approach in the related research involves extracting a set of domain-specialized features from the views, concatenating and feeding them to a single model, such as classical ML models \citep{bocca2016effect, cai2019integrating,maimaitijiang2020soybean}.
Among the architectures used, recent research has employed Convolutional Neural Networks (CNNs) and Recurrent Neural Networks (RNNs) with the same approach \citep{gavahi2021deepyield}. 
Additionally, some works \citep{maimaitijiang2020soybean,chu2020end,shahhosseini2021corn} use a feature-level fusion strategy, where a view-encoder is used to learn a new set of features from each view, before merging them. In this way, the fusion is done at the intermediate layers of neural network models instead of at the input layer.
However, these techniques use a static fusion function (such as concatenation or average operators), which ignores the variable impact that each view has for the yield value.

Our case study consists of the crop yield prediction utilizing multiple RS views, each characterized by distinct spatial and temporal resolution. The target data is the crop yield from fields rasterized at a spatial resolution of 10 m/px, which we refer to as \textbf{sub-field} level data. 
We use multiple crops (soybean, wheat, and rapeseed) and multiple regions (Argentina, Uruguay, and Germany) grouped into four dataset combinations. 
We use multi-spectral optical images from the Sentinel-2 (S2) mission to provide the main information about the Earth surface. 
In addition, we collected different RS data sources (views) to enhance the modeling of the yield prediction task, and provide further information that the optical image might not capture directly.
We include weather features during the growing season, and static information from soil properties, and elevation maps.
For this scenario, we propose a pixel-wise approach for the crop yield prediction at sub-field level.
We use a MVL model that performs data fusion at the feature-level using Gated Units (GUs; \cite{arevalo2020gated}). The features are learned by dedicated view-encoders, allowing to handle the heterogeneous nature of input views with different temporal resolution and data distributions.
Since the RS data used is fairly diverse, we include the GU module in the MVL model to adaptively fuse the learned feature vectors for each view based on the multi-view data and \shortGFweights.
We use a cross-validation splits of the fields with the $R^2$, MAE, and MAPE metrics at {field} and {sub-field} level (pixels) computation.
We obtain improvements across all metrics when compared to previous approaches based on single-view learning, input-level fusion \citep{miranda2023}, and other conventional feature-level fusion.
In our approach, the view-encoder that learns a high-level representation for each view combined with the GU allows fusing the multi-view features in different ways for each sample (pixel).
Thus, the best results are obtained by feeding all the views, contrary to previous models in the literature \citep{bocca2016effect,kang2020comparative,miranda2023}.
The $R^2$ values are around $0.80$ across all datasets at field-level evaluation. While at sub-field level, the score is $0.68$ for Argentina and around $0.44$ for Uruguay and Germany.
As a result of this study, we present additional insights. For instance, the learned adaptive fusion depends on the evaluated country and crop-type.
The key contributions of this paper are as follows:
\begin{itemize}
    \item We propose a two-component model that (i) learns a high-level representation for each view via dedicated view-encoders, and (ii) learns to adaptively fuse this multi-view data with an attention-like mechanism (Gated Unit). To the best of our knowledge, the proposed Multi-View Gated Fusion (\textbf{\gatedfusion}) is the first model applying an adaptive fusion (via gated units) approach to the multi-view crop yield prediction task.
    \item We empirically show that the best results in the MVL scenario are consistently obtained by utilizing all views (\sent, \weat, \dem, and \soil). In addition, we obtain overall improvements compared to an input-level fusion baseline and single-view based models.
    \item We present experiments to identify the key factors that contribute to the success of our approach, such as leave-one-year-out validation, spatial coverage of S2 analysis, and \GFweights visualization. It is worth highlighting that the GU module in the proposed \gatedfusion learns different distribution of the fusion weights depending on the region and task.
\end{itemize}

The paper is organized as follows: In Section~\ref{sec:sota}, related works in crop yield prediction and adaptive fusion are presented. While Section~\ref{sec:data} describes the data and study, Section~\ref{sec:methods} explains the proposed approach.
Experimental settings and main results are described in Section~\ref{sec:experiments}. 
Additional analysis is presented in Section~\ref{sec:analysis}. Finally, the conclusion about the work are in Section~\ref{sec:conclusion}.

%% file: content/sota.tex
\section{Related Work} \label{sec:sota}

Recently, there has been an increase of MVL models applied to different RS tasks. Since the approaches usually incorporate domain knowledge, they vary from task to task \citep{mena2022common}. In the following, we briefly discuss some related works in MVL for RS.

\paragraph{MVL for crop yield prediction} 
The standard approach in crop yield prediction is to build a specialized domain-specific set of features across the growing season and then apply standard ML models \citep{bocca2016effect, cai2019integrating}, e.g. random forest and Multi Layer Perceptron (MLP), or deep learning based \citep{gavahi2021deepyield,miranda2023}, e.g. with convolutional (CNN) or recurrent (RNN) operations. 
In these cases, the input-level fusion is used to merge the information coming from multiple RS sources, i.e. all the input features are concatenated and then feed to a single model.
However, certain approaches involve the fusion of hidden features in intermediate layers of neural network models, a concept known as feature-level fusion. This strategy needs a sub-model for each view (referred to as a view-encoder), which learns features.
For instance, \cite{yang2019deep} use two views, RGB and multi-spectral images, to predict the crop yield at county level, where a 2D CNN is used on each view-encoder.
Some works apply feature-level fusion with other types of view-encoder, depending on the data used. \cite{maimaitijiang2020soybean} use MLPs for vector multi-view data, while \cite{chu2020end} incorporate an independent RNN for meteorological dynamic features and an MLP for county information. \cite{shahhosseini2021corn} use a 1D CNN (across time) for weather data, 1D CNN (across depths) for soil properties, and an MLP for vector data.
Additionally, some works group views based on their information content.
For instance, \cite{wang2020winter} group static (soil properties) and dynamic (optical and meteorological) features into a two view model that performs feature-level fusion with an MLP and Long-Short Term Memory (LSTM) as view-encoders.
Subsequent works \citep{cao2021integrating,srivastava2022winter} have applied a similar group of static and dynamic views for the yield prediction with the feature-level fusion. 
However, we model individually the multi-view data, deriving them from four RS sources, and employ an adaptive fusion approach.

\paragraph{Adaptive fusion and attention in RS}
As different variations of neural networks models have been used in literature, different forms of the {attention mechanism} \citep{vaswani2017attention} have shown state-of-the-art results in RS applications. The temporal attention pooling is used by some works \citep{lin2020deepcropnet,garnot2020lightweight,ofori2021crop} to aggregate dynamic RS views on time-series data. 
Attention mechanisms have also been used to highlight spatial and spectral features of RS data \citep{wang2022multi}. 
Motivated by attention mechanisms and mixture-of-expert models \citep{jacobs1991adaptive}, some studies have explored gated fusion approaches in pursuit of adaptively fusing multiple features \citep{zhang2020hybrid,zheng2021gather,hosseinpour2022cmgfnet}. The main idea is to highlight (apply an adaptive weight to) the most relevant information of each view and aggregate them, e.g. with a linear sum \citep{arevalo2020gated}. 
Furthermore, recent literature is exploring the connection between attention and explainability. For instance, explanation through attention is used for  
land-use and land-cover classification \citep{meger2022explaining} and crop-type mapping \citep{russwurm2020self, obadic2022exploring}. 
This is mainly because the data dependent weights operate as a proxy for feature importance by identifying which features are most used for model prediction. 
Our work considers a tailored of the adaptive fusion based on a gating mechanism. To the best of our knowledge, this is the first work applying the gated fusion approach to the multi-view crop yield prediction task.

\paragraph{MVL for land-use and land-cover task}
In the widely studied land-use and land-cover mapping, the target data focuses on a specific and limited time frame. Typically, the views consist of static views captured by a variety of sensors, including multi-spectral optical or radar images, or elevation maps.
\cite{chen2017deep} propose one of the first models that use deep neural networks on the view-encoders (concretely 2D CNN) for two sensors and perform feature-level fusion.
Later, to fuse sensors with different resolution, \cite{benedetti2018m} include an auxiliary classifier for each view that is feed with the learned features and acts as a regularization. 
\cite{wu2021convolutional} use a variation of this approach with auxiliary reconstructions of the learned features.
On the other side, \cite{audebert2018beyond} propose to fuse across all layers in 2D CNN view-encoders with a central model. 
They use skip-connections between individual views and post-fusion layers, where \cite{hosseinpour2022cmgfnet} later extend the merge across all skip-connections layers.
To account for different levels of information that views have, \cite{wang2022multi} merge the learned features in a hierarchical way.
In addition, the decision-level fusion (merge class predictions) has been explored without significant advantages compared to feature fusion \citep{audebert2018beyond,ofori2021crop}. 
These works usually consider a pixel-wise prediction as our application. However, we are considering a time-dependent target with dynamic features.

Recent works in crop yield prediction focuses on feature fusion at field level by grouping dynamic and static features. However, the standard merger are static functions, such as simple concatenation, while in other RS applications more sophisticate approaches have been used. 
Since the gated fusion (through the adaptive weighted sum) suits the dynamic significance that each view could have for prediction, we use it for a sub-field crop yield prediction case study.

%% file: content/data.tex
\section{Data} \label{sec:data}
Our case study consists of the crop yield prediction task based on multiple RS views (with different spatial and temporal resolutions). The target data is the crop yield at {sub-field level} (pixel-wise yield values at 10 m spatial resolution) over different countries and crop-types.

\subsection{Crop Yield Data}
\begin{figure}[!t]
    \centering
    \includegraphics[width=0.7\linewidth]{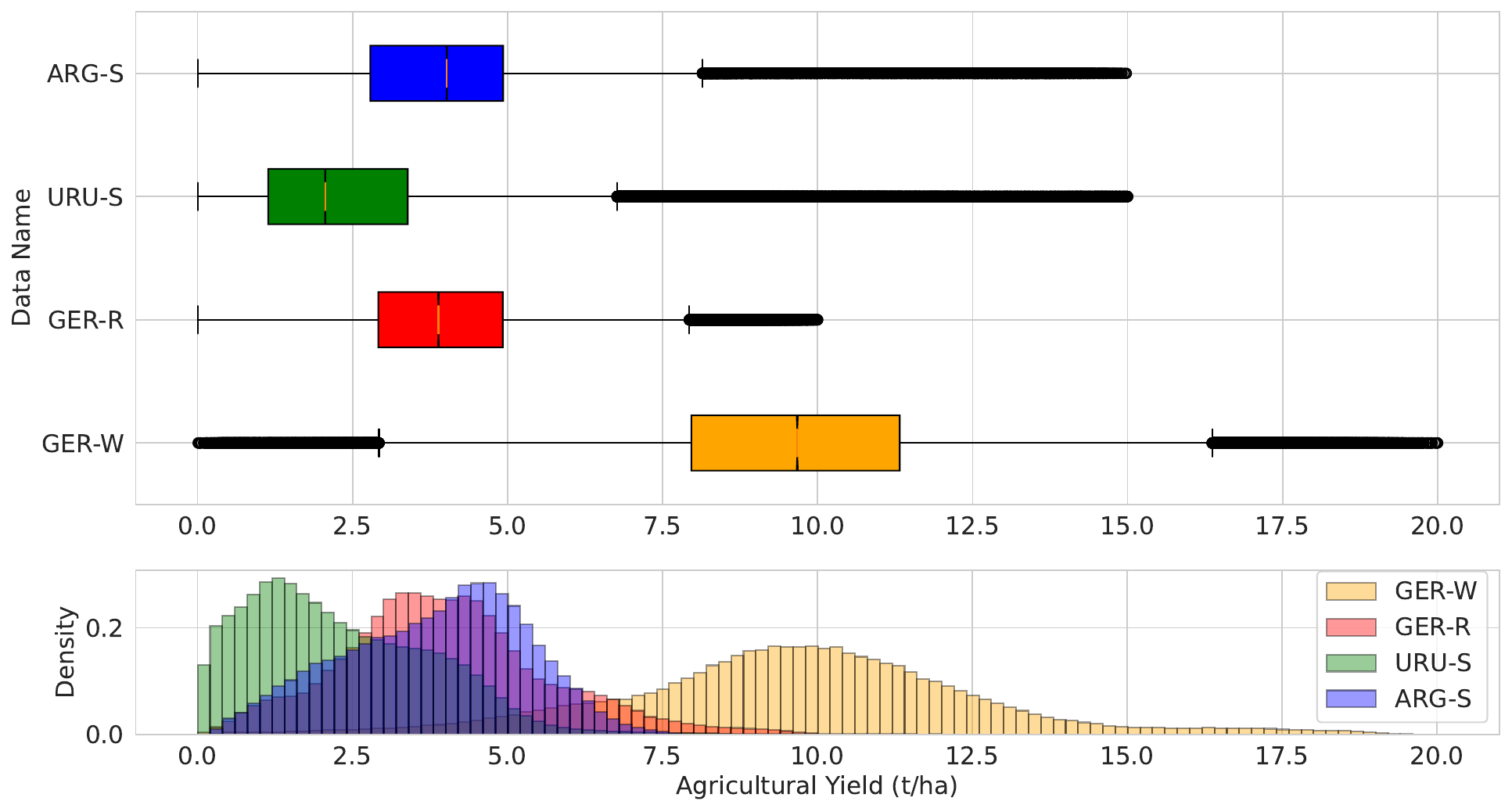}
    \caption{Crop yield distribution per pixel (at 10 m resolution) in the four datasets considered in this study.}
    \label{fig:yield_dist}
\end{figure}

The crop yield data corresponds to the target (ground-truth) variable to be predicted by machine learning models.
This yield data comes from combine harvesters at a sub-field level.
The combine harvester equipped with yield monitors records data at consistent intervals with a high spatial resolution as it moves across the field during the harvesting process.
All records collected are characterized by different features such as the geolocation, yield moisture, and amount of yield. 
We preprocessed the records by re-projecting the reference coordinate system, removing wrong values\footnote{An example of a wrong yield record will be a biological infeasible crop yield.} (based on position, timestamp, yield, and moisture), and filtering based on a statistical threshold (three standard deviations around the mean of the field distribution). 
The resulting geolocated data for each field is rasterized into 10 meter per pixel (m/px), which we named yield maps. The unit of these yield maps is tons per hectare (t/ha). 
Fig.~\ref{fig:yield_dist} shows the variability of the yield data from the different datasets used in this study. The distribution of the yield, e.g. mean and standard deviation, changes between country and crop-type. 

\begin{table*}[t!]
    \centering
    \caption{Descriptive factors of the four datasets in this study, with different combinations of country and crop-type.} \label{tab:region_distribution}
    \scriptsize
    \begin{tabularx}{\linewidth}{C|C|C|c|c|c|c|c|C|C} \hline
        \multirow{3}{*}{Name} & \multirow{3}{*}{Country} & \multirow{3}{*}{Crop-type} & \multirow{3}{*}{Years} & \multirow{3}{*}{Total Area} & \multirow{3}{*}{Fields} & \multirow{3}{*}{Pixels}  & \multirow{2}{*}{Yield value} & Total Area & Growing season \\ \cline{8-10}
        & & & & & & &  mean $\pm$ std. & \multicolumn{2}{c}{average across field} \\ \hline
        ARG-S & Argentina & soybean & 2017--2022 & 15351 ha & 190 & $\sim 1.4$ M & $3.86 \pm 1.49$ & 79.5 ha & 156 days\\
        URU-S & Uruguay & soybean & 2018--2021 & 28358 ha & 486 & $\sim 1.8$ M & $2.35 \pm 1.59$ & 58.3 ha & 169 days \\
        GER-R & \multirow{2}{*}{Germany} & rapeseed & 2016--2022 & 3221 ha &  111 & $\sim 0.3$ M & $4.01 \pm 1.67$ &  29.0 ha & 335 days\\
        GER-W &  & wheat & 2016--2022 & 3240 ha & 188 & $\sim 0.3$ M & $9.64 \pm 2.95$ & 17.2 ha & 306 days \\
        \hline
    \end{tabularx}
\end{table*}
\paragraph{Region}
Unlike usual crop-region specific use-cases in the literature of crop yield prediction, we use field data across different countries (Argentina, Uruguay, and Germany), crop-types (soybean, rapeseed, and wheat), and years (from 2016 until 2022). Table~\ref{tab:region_distribution} presents the total number of fields and yield pixels in each dataset considered in this study. 
While there are a larger number of pixels of soybean crops (in Argentina and Uruguay), the crops in Germany have fewer number of  pixels.
The field data of this study is fairly diverse, as the area covered per field is different in each combination, and, over the years, the fields are distributed in different geographical locations within each country.
Furthermore, the seeding and harvesting dates are different across the fields and countries. We present this in more detail in the Appendix~\ref{fig:farm_seedtoharv}.

\subsection{Multi-View Input Data} \label{sec:data:mv}
We use dynamic (time-dependent) and static input data collected from different RS sources.  Since there are numerous RS sources available nowadays, this work is limited to the selected sources. Our selection criteria hinge on the global availability of the RS data, coupled by its usefulness in estimating crop yield.
The aim is to have a better representation and modeling of the crop yield drivers through the growing season, and therefore improve the crop yield prediction.
Table~\ref{tab:summary:multiview} displays an overview of the multi-view RS data used in this study.

\subsubsection{S2-based Optical Image} \label{sec:data:mv:s2} 
We use multi-spectral optical information coming from the Sentinel-2 (S2) mission. Specifically, we use the surface reflectance imaging product (level-2A) from the S2 data,  which is available from 2016. 
For this, we collected a Satellite Image Time Series (SITS) from seeding to harvesting date of each field, with approximately 5-days revisit time. 
We use the 12 spectral bands of the L2A product: B1, B2 (blue), B3 (green), B4 (red), B5, B6, B7, B8 (near-infrared), B8A, B9, B11, B12, where the bands with lower spatial resolutions (B1, B5-B7, B8A, B9-B12) are up-sampled to the ones with higher (at 10 m/px).
Finally, the number of images in the SITS per field ranges from 11 to 78 for Argentina, from 21 to 82 for Uruguay, and from 17 to 140 for Germany, depending on the crop growing season. 

\begin{figure*}[t!]
\centering
\subfloat[Fields in ARG-S data.\label{fig:coverage_time:a}]{\includegraphics[width=0.48\linewidth]{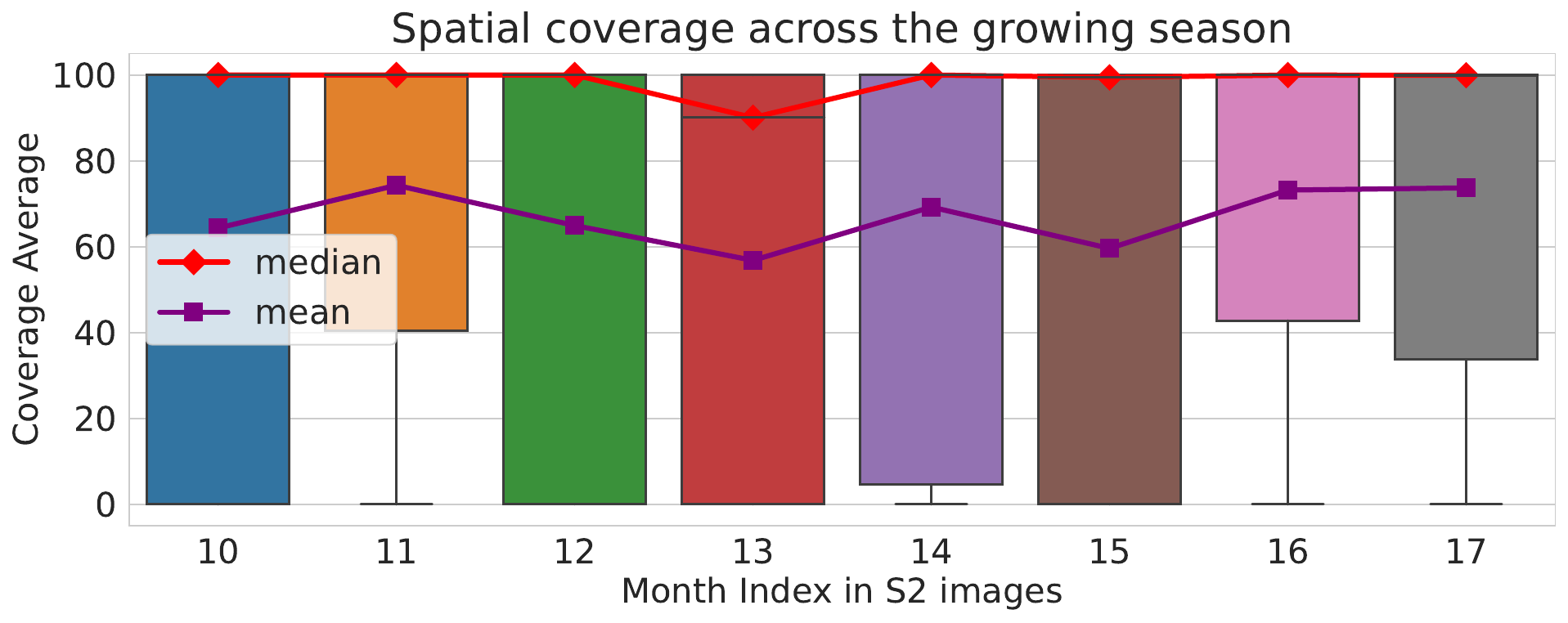}}
\hfill
\subfloat[Fields in URU-S data.\label{fig:coverage_time:b}]{\includegraphics[width=0.48\linewidth]{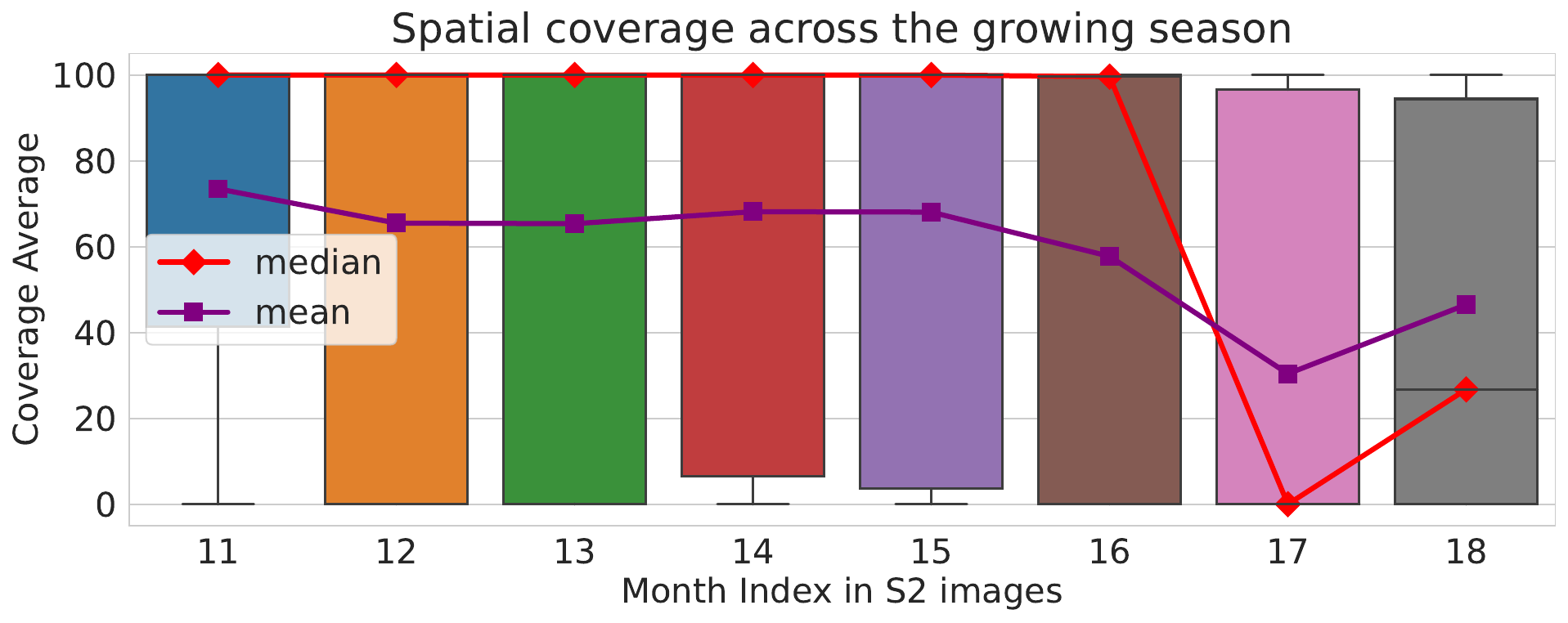}}\\
\subfloat[Fields in GER-R data.\label{fig:coverage_time:c}]{\includegraphics[width=0.48\linewidth]{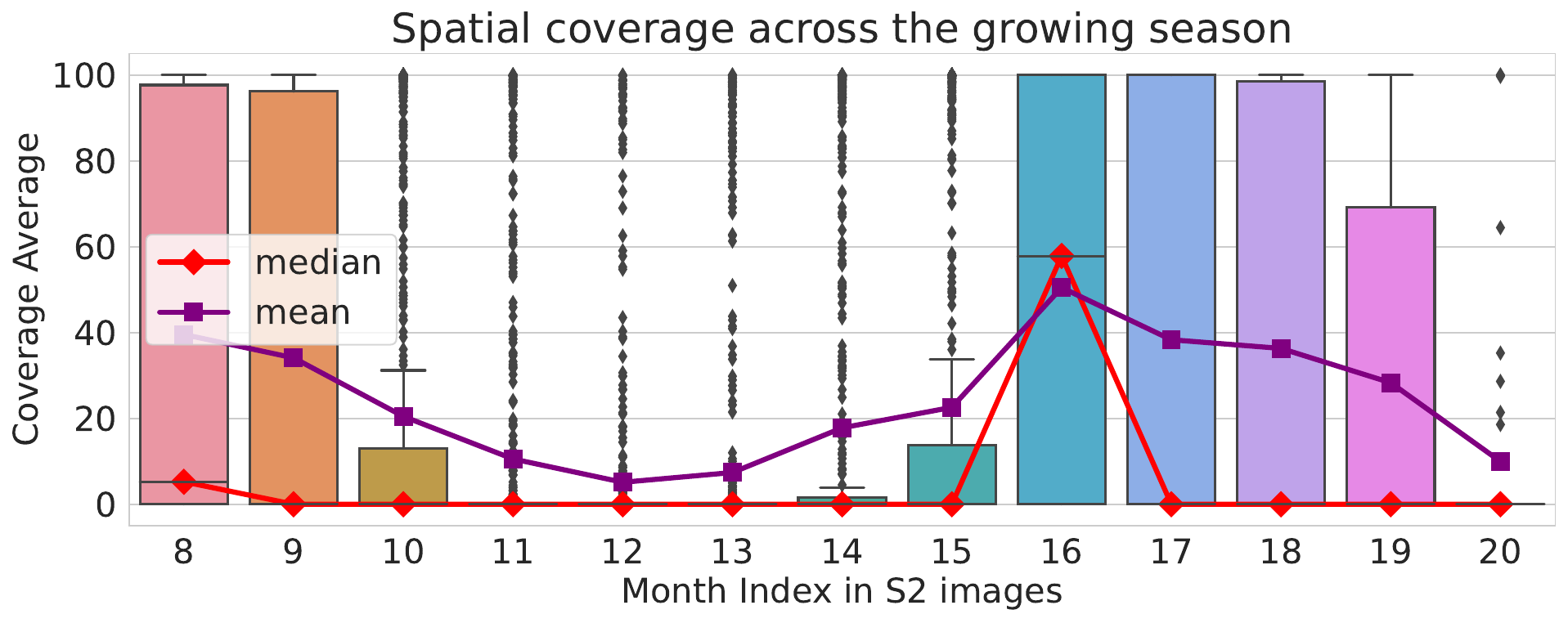}}
\hfill
\subfloat[Fields in GER-W data.\label{fig:coverage_time:d}]{\includegraphics[width=0.48\linewidth]{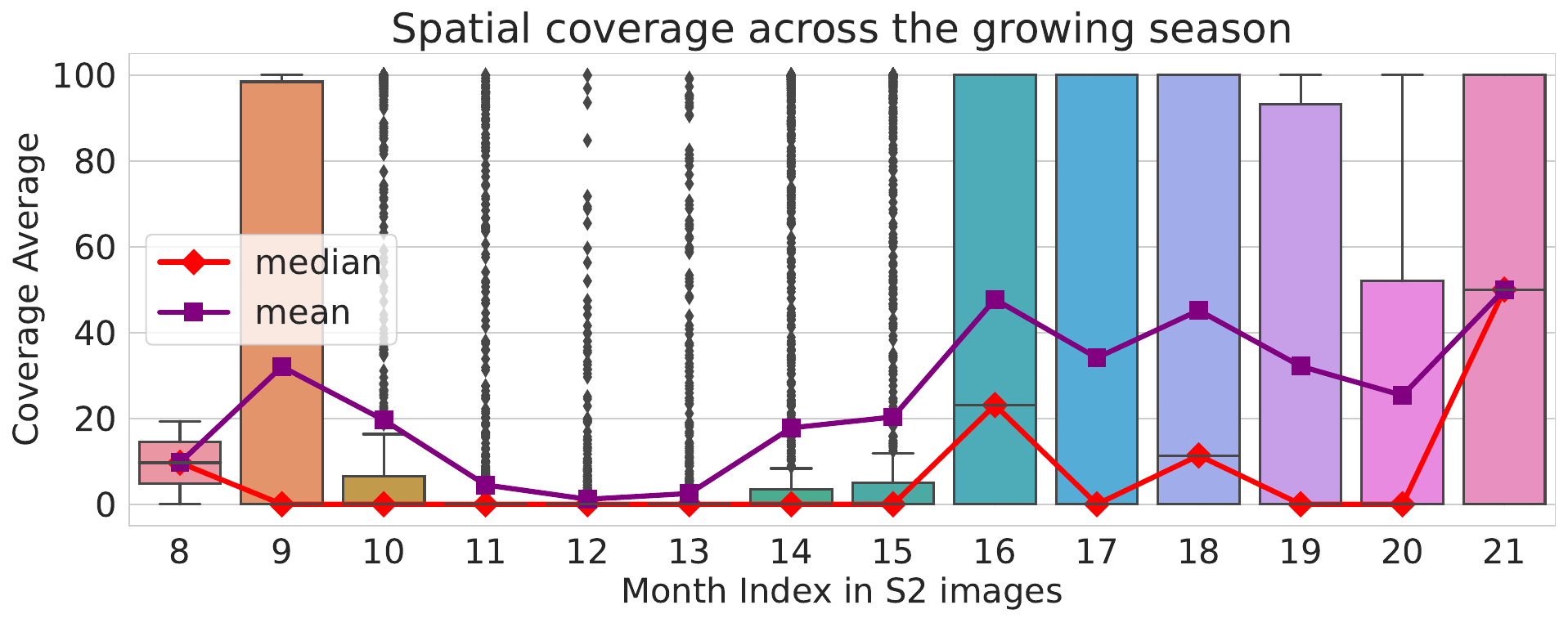}}\\
\caption{Field spatial coverage (labels 4 and 5 in S2-based SCL) across the growing season on different fields. Different sub-figures show different datasets used in this study. Each point in a boxplot represents the spatial coverage of a field in the corresponding month. The spatial coverage of the fields is grouped by month for display. The month index goes from January in the seeding year (1) to December of the following year (24). }\label{fig:coverage_time}
\end{figure*}

Thanks to the information contained in the Scene Classification Layer (SCL) mask from the L2A product, we can calculate the spatial coverage of the S2-based SITS. 
The SCL is an additional layer on the S2 data that assigns a class within 12 options\footnote{SCL labels: no data (0), saturated/defective (1), dark area pixel (2), cloud shadows (3), vegetation (4), not vegetated (5), water (6), unclassified (7), cloud medium probability (8) or high probability (9), thin cirrus (10), snow (11).} for each pixel in the image.
By ignoring labels related to external factors (snow, water, and errors) a percentage of pixels related to a clean field observation is computed for each SITS. We refer to this as \textbf{field spatial coverage}. 
In Fig.~\ref{fig:coverage_time}, we show the monthly spatial coverage for the fields across the growing season in the different datasets. In Germany, there is a longer growing season since it contains winter crops\footnote{Winter crops are a type of crop planted in early winter and harvested in early summer that could handle cooler temperatures, frost, and shorter days.}, where we see a low spatial coverage for the winter months (from December to February of the following year).
These charts show the diversity of the \sentsource optical images across the growing season for different countries and months.
For further visualization, Fig.~\ref{fig:coverage_time_year} in the appendix displays the spatial coverage over different years.

\subsubsection{Weather} \label{sec:data:mv:weather}
We utilize meteorological factors obtained from the climate source, ECMWF ERA5 \citep{hersbach2020era5}. 
This source is based on the assimilation of various observations from satellites, ground-based weather stations, and other sources into a consistent numerical weather model, which has been available from 1979.
The raw data collected for each field is at hourly temporal resolution, from which we extract 4 daily features: mean temperature, maximum temperature, minimum temperature, and cumulative precipitation.
The temperature values are the air temperature at 2 m above the land surface in a raw grid with a spatial resolution of 30 km/px.

\subsubsection{Digital Elevation Model (DEM)} \label{sec:data:mv:dem}
We use topographic information collected from the NASA source, Space Shuttle Radar Topography Mission (SRTM) \citep{farr2000shuttle}. Elevation information was collected by bouncing radar signals to the Earth's surface. 
We use 5 features extracted with the RichDEM tool\footnote{\url{http://github.com/r-barnes/richdem}} at a spatial resolution of 30 m/px: aspect, curvature, digital surface model, slope, and topographic wetness index.
This information is static across time.

\subsubsection{Soil Map} \label{sec:data:mv:soil}
We utilize soil properties obtained from the global source, SoilGrids \citep{poggio2021soilgrids}. 
This data source integrates ground-based and satellite measurements, as well as ML-based predictions. 
We select the following features: cation exchange capacity, volumetric fraction of course fragments, clay, nitrogen, soil pH, sand, silt, and soil organic carbon, from 3 soil depths: 0-5, 5-15, and 15-30 cm. 
The values are static across time, with a spatial resolution of 250 m/px.

\begin{table}[t!]
    \centering
    \caption{Summary of the collected multi-view RS input data. It is worth noting that there are significantly more features in some RS sources, such as weather and soil, however we are displaying the number used in our study.} \label{tab:summary:multiview}
    \footnotesize
    \begin{tabularx}{\linewidth}{C|C|C|C|C} \hline
        View & RS Source & Spatial Resolution & Temporal Resolution & Number of Features ($B_v$) \\ \hline
        Optical & \sentsource & 10 m & 5 days & 12 \\
        & S2-based SCL & 10 m & same as S2 & 1 (categorical) \\
        Weather & \weatsource & 30 km & daily & 4\\
        \dem & \demsource & 30 m & $-$ & 5 \\
        Soil & \soilsource & 250 m & $-$ & 8 ($\times$ 3 depths) \\
        \hline
    \end{tabularx}
\end{table}

%% file: content/methods.tex
\section{Method Description} \label{sec:methods}

We use a pixel-wise approach for the prediction task.
The approach uses a feature-level fusion with an adaptive fusion approach, which shows good results in RS applications (Sec.~\ref{sec:sota}).

\begin{figure*}[t!]
    \centering
    \includegraphics[width=0.95\textwidth]{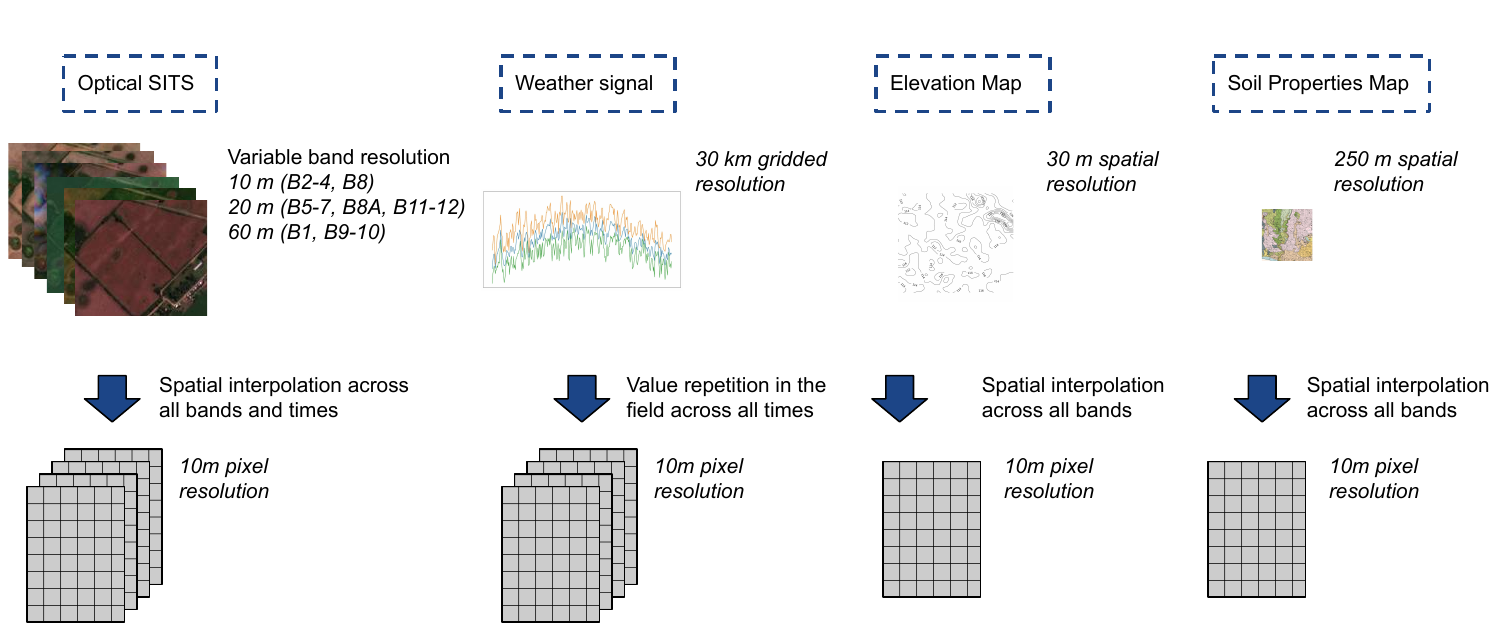}
    \caption{Illustration of spatial alignment applied to the four input views for a specific field. After this process, all views have a spatial resolution of 10 m/px.}  \label{fig:spatial_illustration}
\end{figure*}
\paragraph{Spatial alignment}
To harmonize the different spatial resolutions for the pixel-wise approach, we spatially align the multi-view data before feeding them to ML models.
For \dem and \soil views, a cubic spline method is used to interpolate to the spatial resolution of the \sentsource images (10 m/px). 
While for \weat, the value from the centroid of the field is repeated across all the field pixels to match the same spatial resolution. See Fig.~\ref{fig:spatial_illustration} for an illustration.

\subsection{Data Formulation and Notation}
Consider the following MVL scenario with $N$ labeled pixels, $\mathcal{D} = \{ \mathcal{X}^{(i)}, y^{(i)} \}_{i=1}^N$, $y^{(i)} \in \mathbb{R}+$ the ground-truth crop yield for the $i$-th pixel, and $\mathcal{X}^{(i)} = \left\{ \mat{X}_{\text{S2}}^{(i)}, \mat{X}_{\text{W}}^{(i)}, \vect{x}_{\text{D}}^{(i)}, \vect{x}_{\text{S}}^{(i)} \right\} $ its corresponding multi-view input data. 
Let $B_v$ be the number of bands or features in each view $v \in \left\{ \text{S2},\text{W},\text{D}, \text{S} \right\}$ (see Table~\ref{tab:summary:multiview}). 
The \sentsource-based \sent view for the $i$-th pixel is a multivariate time-series of length $T_{\text{S2}}^{(i)}$: $\mat{X}_{\text{S2}}^{(i)} \in \mathbb{R}^{T_{\text{S2}}^{(i)} \times B_{\text{S2}}^{}}$, 
and the \weat view is a multivariate time-series of length $T_{\text{W}}^{(i)}$: $\mat{X}_{\text{W}}^{(i)} \in \mathbb{R}^{T_{\text{W}}^{(i)} \times B_{\text{W}}}$.  
Note that the temporal resolution between views and pixels have not been aligned.
On the other hand, \dem and \soil views are constant variables over time for the $i$-th pixel, $\vect{x}_{\text{D}}^{(i)} \in \mathbb{R}^{B_{\text{D}}}$ and $\vect{x}_{\text{S}}^{(i)} \in \mathbb{R}^{B_{\text{S}}}$.
Additionally, $\mathsf{M}: \mathbb{R}^d \times \mathbb{R}^d \times \mathbb{R}^d \times \mathbb{R}^d \rightarrow \mathbb{R}^{m}$ is a merge function, e.g. concatenation ($m=4d$) or average ($m=d$), and $\mathsf{S}: \mathbb{R}^d \times \mathbb{R}^d \times \mathbb{R}^d \times \mathbb{R}^d \rightarrow \mathbb{R}^{4 \times d}$ the stacking function, with $d$ a vector dimensionality.

\subsection{Feature-level Learning}
In order to fuse the views with different temporal resolution and number of features, we learn a single high-level vector representation for each view on a shared\footnote{Since the vectors are element-wise added in \eqref{eq:weighted_sum}, the ``shared'' space means that each dimension interacts aligned.} $d$-dimensional space, named view-representation. We use one dedicated encoder model for each view built with artificial neural networks,  named view-encoder, see Fig.~\ref{fig:mvgf:illustration} for an illustration. This view-encoder is a function $E_{\theta_v}: X_v \rightarrow \mathbb{R}^d$, with the corresponding learnable parameters $\theta_v$ for the $v$ view.

\begin{align}
    \vect{z}_{\text{S2}}^{(i)} &= E_{\theta_{\text{S2}}}\left(\mat{X}_{\text{S2}}^{(i)}\right) \in \mathbb{R}^d \\ 
    \vect{z}_{\text{W}}^{(i)} &= E_{\theta_{\text{W}}}\left( \mat{X}_{\text{W}}^{(i)} \right)  \in \mathbb{R}^d \\
    \vect{z}_{\text{D}}^{(i)} & = E_{\theta_{\text{D}}}\left( \vect{x}_{\text{D}}^{(i)} \right) \in  \mathbb{R}^d \\
    \vect{z}_{\text{S}}^{(i)} & = E_{\theta_{\text{S}}}\left( \vect{x}_{\text{S}}^{(i)}\right) \in \mathbb{R}^d 
\end{align}
These learned view-representations ($\vect{z}_v$) allow the model to handle the heterogeneous nature of the views (different temporal resolutions, magnitudes, and data distributions).
The view-encoder gives the model the chance to extract information with a specific and dedicated model. Furthermore, the view-encoder might use different types of architecture (\textit{asymmetric} network). 

\paragraph{Temporal view-encoders} For multivariate time-series $\mat{X}  =  \{ \vect{x}_1, \vect{x}_{2}, \ldots, \vect{x}_{T} \} \in \mathbb{R}^{T \times B}$, with the $t$-th observation $\vect{x}_{t} \in \mathbb{R}^{B}$, 
recurrent layers (or RNN) could be used to extract temporal high-level representations $\vect{h}_t \in \mathbb{R}^d$. With $H^{(l)}$ a recurrent unit (e.g. LSTM) at layer $l$, and $\vect{h}_0^{(l)} = \vect{0} \ \forall l$ , the hidden state at a time $t$ and layer $l$ can be expressed by
\begin{equation}
    \vect{h}_t^{(l)} = \left\{ \begin{array}{ll}
             H^{(l)} \left(\vect{x}_t, \vect{h}_{t-1}^{(l)} \right) , & l=1 \\
             H^{(l)} \left(\vect{h}_t^{(l-1)}, \vect{h}_{t-1}^{(l)} \right) , & l \in \{ 2, 3, \ldots  L \}
             \end{array} \right. .
\end{equation}
Then, the last hidden state could be used to extract a single vector representation $\vect{a}^{(L)} = \vect{h}_T^{(L)} \in \mathbb{R}^d$. Additionally, attention-based approaches, such as temporal attention pooling ($\vect{a}  = \sum_t \alpha_t  \vect{h}_t$) could be used (in Sec.~\ref{sec:analysis:ablation} we show an empirical comparison).
Given the dynamic features of \sent and \weat views, we use these types of architectures in $E_{\theta_{\text{S2}}}$, $E_{\theta_{\text{W}}}$.

\paragraph{Static view-encoders} For vector data $\vect{x} \in \mathbb{R}^{B}$, fully connected layers (or MLP) are used to extract a high-level representation at the output layer $\vect{a}^{(L)} \in \mathbb{R}^d$. With $H^{(l)}$ 
a linear projection followed by a nonlinear activation function on layer $l$, the output of a layer $l$ could be written as 
\begin{equation}
    \vect{a}^{(l)} = \left\{ \begin{array}{ll}
             H^{(l)} \left(\vect{x} \right) , & l=1 \\
             H^{(l)} \left(\vect{a}^{(l)} \right) , & l \in \{ 2, 3, \ldots  L \}
             \end{array} \right. .
\end{equation}
Given the static features of \dem and \soil views, we use these types of architectures in $E_{\theta_{\text{D}}}$, $E_{\theta_{\text{S}}}$.

\subsection{Adaptive Fusion with Gating Mechanism (Gated Fusion)} \label{sec:methods:gatedfusion}

The learned view-representations could be fused with the concatenation merge function ($\mathsf{M}$), as usual in yield prediction \citep{wang2020winter,shahhosseini2021corn,srivastava2022winter}:
\begin{equation} \label{eq:static_merge}
    \vect{z}_{\text{F}}^{(i)} = \mathsf{M} \left( \vect{z}_{\text{S2}}^{(i)}, \vect{z}_\text{W}^{(i)}, \vect{z}_\text{D}^{(i)}, \vect{z}_\text{S}^{(i)}\right) \ .
\end{equation}
However, this static merge function does not align with the variable contributions that each view has in predicting crop yield \citep{kang2020comparative,miranda2023}. Additionally, it ignores the real-time environment of EO, where different phenomena (e.g. clouds in optical images or noise in measurement) can affect the data quality \citep{ofori2021crop,ferrari2023fusing}.
Therefore, inspired by the Gated Units (GUs, \cite{arevalo2020gated}), we propose an adaptive fusion approach via gating mechanisms, the so-called gated fusion.

\begin{figure}[t!]
    \centering
    \includegraphics[width=0.6\textwidth]{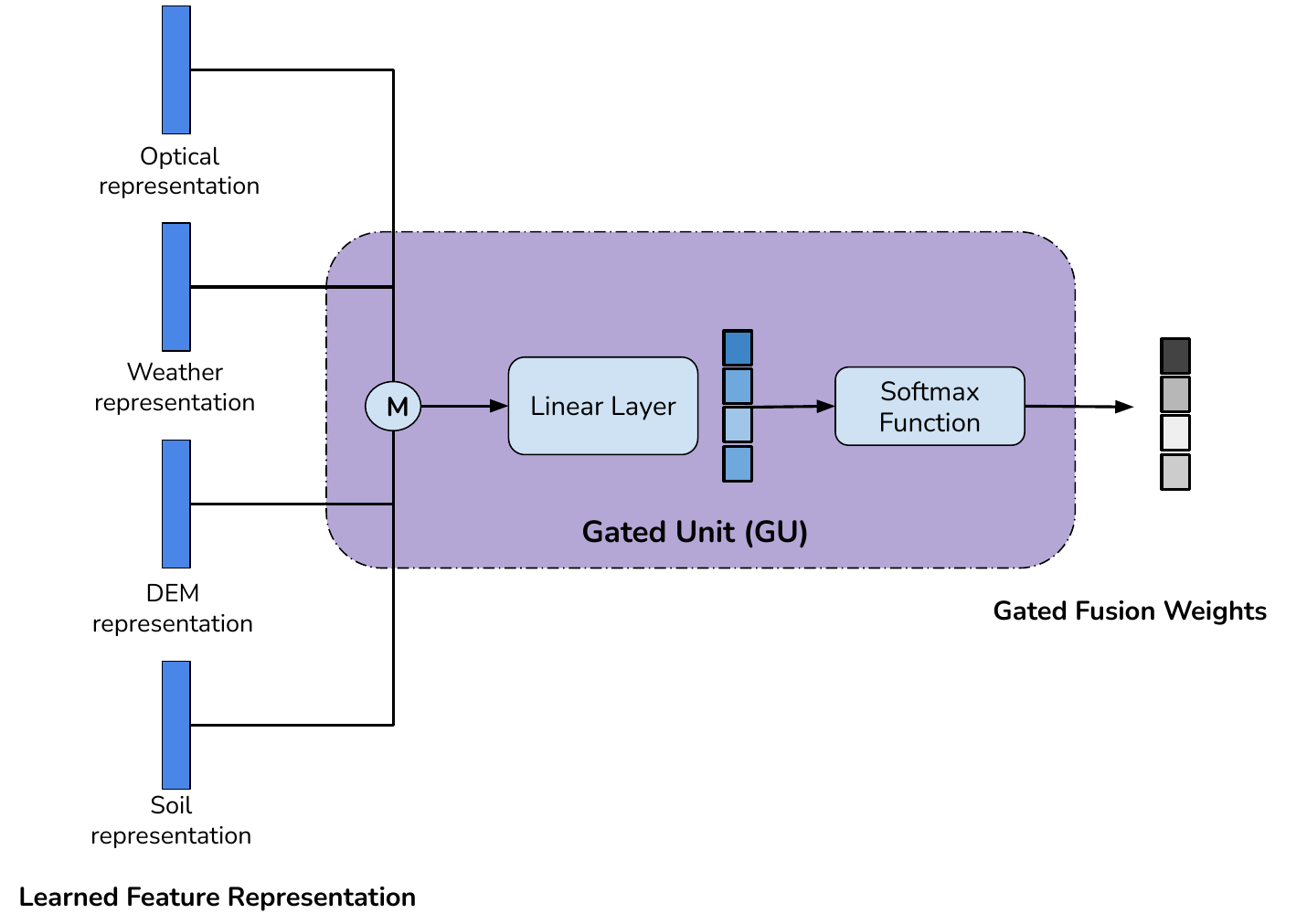}
    \caption{Illustration of the used gating mechanism (GU). The four view-representation are merged with $\mathsf{M}$ and linearly projected to a four-dimensional vector. Then, softmax function is applied and normalized fusion weights are obtained.}  \label{fig:gu_illustration}
\end{figure}
We use a gating function $G_{\theta_{\text{G}}}: \mathbb{R}^d \times \mathbb{R}^d \times \mathbb{R}^d \times \mathbb{R}^d \rightarrow \mathbb{R}^4 $ that takes the four view-representations, and generates four pixel-specific values ${\alpha}_v^{(i)} \in [0,1]$ with $v \in \{\text{S2}, \text{W}, \text{D}, \text{S}\}$, which we refer to as \textbf{\GFweights.} The following computation is used
\begin{equation}\label{eq:attention_weights}
\vect{\alpha}^{(i)} = G_{\theta_{\text{G}}}\left(  \vect{z}_{\text{S2}}^{(i)},  \vect{z}_\text{W}^{(i)}, \vect{z}_\text{D}^{(i)}, \vect{z}_\text{S}^{(i)}  \right) = \text{softmax}\left( \mathsf{M} \left( \vect{z}_{\text{S2}}^{(i)}, \vect{z}_\text{W}^{(i)}, \vect{z}_\text{D}^{(i)}, \vect{z}_\text{S}^{(i)}\right)^{\top} \mat{\theta}_{\text{G}} \right) \ ,
\end{equation}
where the learnable parameters $\theta_{\text{G}}$ depend on the choice of $\mathsf{M}$, when $\mathsf{M}$ is concatenation: $\mat{\theta}_{\text{G}} \in \mathbb{R}^{4d \times 4} $, and when $\mathsf{M}$ is average, $\mat{\theta}_{\text{G}} \in \mathbb{R}^{d \times 4} $. For our primary experimentation, we employ the concatenation, however we present a comparison in Sec.~\ref{sec:analysis:ablation}.
This GU module learns a distribution over the views: $ \sum_{v \in \{\text{S2},\text{W},\text{S},\text{D}\}} \alpha_{v}^{(i)} = 1$. 
See Fig.~\ref{fig:gu_illustration} for an illustration.
Then, these \shortGFweights are applied to the stacked vectors to obtains the fused representation $\vect{z}_{\text{F}}^{(i)} \in \mathbb{R}^d$: 
\begin{equation} \label{eq:weighted_sum}
\vect{z}_{\text{F}}^{(i)} = \mathsf{S}\left( \vect{z}_{\text{S2}}^{(i)}, \vect{z}_{\text{W}}^{(i)}, \vect{z}_{\text{D}}^{(i)}, \vect{z}_{\text{S}}^{(i)} \right)^{\top} \vect{\alpha}^{(i)}  = \sum_{v \in \{\text{S2},\text{W},\text{D},\text{S}\} } \alpha_{v}^{(i)} \cdot \vect{z}_{v}^{(i)} \ .
\end{equation}

Then, the model learns the \GFweights which are applied to the view-representations as an adaptive weighted sum \eqref{eq:weighted_sum}. 
The gating mechanism plays a pivotal role in both weight learning and enabling adaptive fusion for each sample.
In our approach, a pixel is a sample. Therefore, the model highlights the views depending on the information contained in the pixel-level representation.
For example, for a cloudy pixel, the model could learn to assign a lower weight to the optical view $\alpha_{\text{S2}}$ and higher to the other views ($\alpha_{\text{W}}, \alpha_{\text{S}}, \alpha_{\text{D}}$). While, for a cloudless pixel, learn to assign a higher weight to the optical view and distribute the rest to the complementary views.

In the gated fusion approach introduced by \cite{arevalo2020gated}, a sigmoid activation function is used in the GU module, feeding it with previous layers of the view-encoders $E_v$, and using concatenation in  $\mathsf{M}$. This is different from our work, where we use the softmax function and feed the GU with $\vect{z}_v$.
Additionally, our approach presents similarities to attention mechanisms, which use the similarity between \textit{queries} ($\mat{Q}$) and \textit{keys} ($\mat{K}$) to provide attention to \textit{values}. In \cite{vaswani2017attention}, attention weights ($\alpha$) are computed with a dot-product: $\mat{Q}^\top \mat{K} / \sqrt{d}$.
Analogously for our case, we use the representation concatenation (in $\mathsf{M}$) as a single \textit{master query} \citep{garnot2020lightweight}, and encode the keys as model parameters (in $\mat{\theta}_{\text{G}}$) as a \textit{key-as-parameter} approach \citep{garnot2020lightweight} 
\footnote{In this work, a \textit{query-as-parameter} was used. 
However, the core 
concept is to encode the components as parameters.}, which provide attention weights to the views \eqref{eq:weighted_sum}.

\subsection{Prediction and optimization}
\begin{figure*}[t!]
    \centering
    \includegraphics[width=0.9\textwidth]{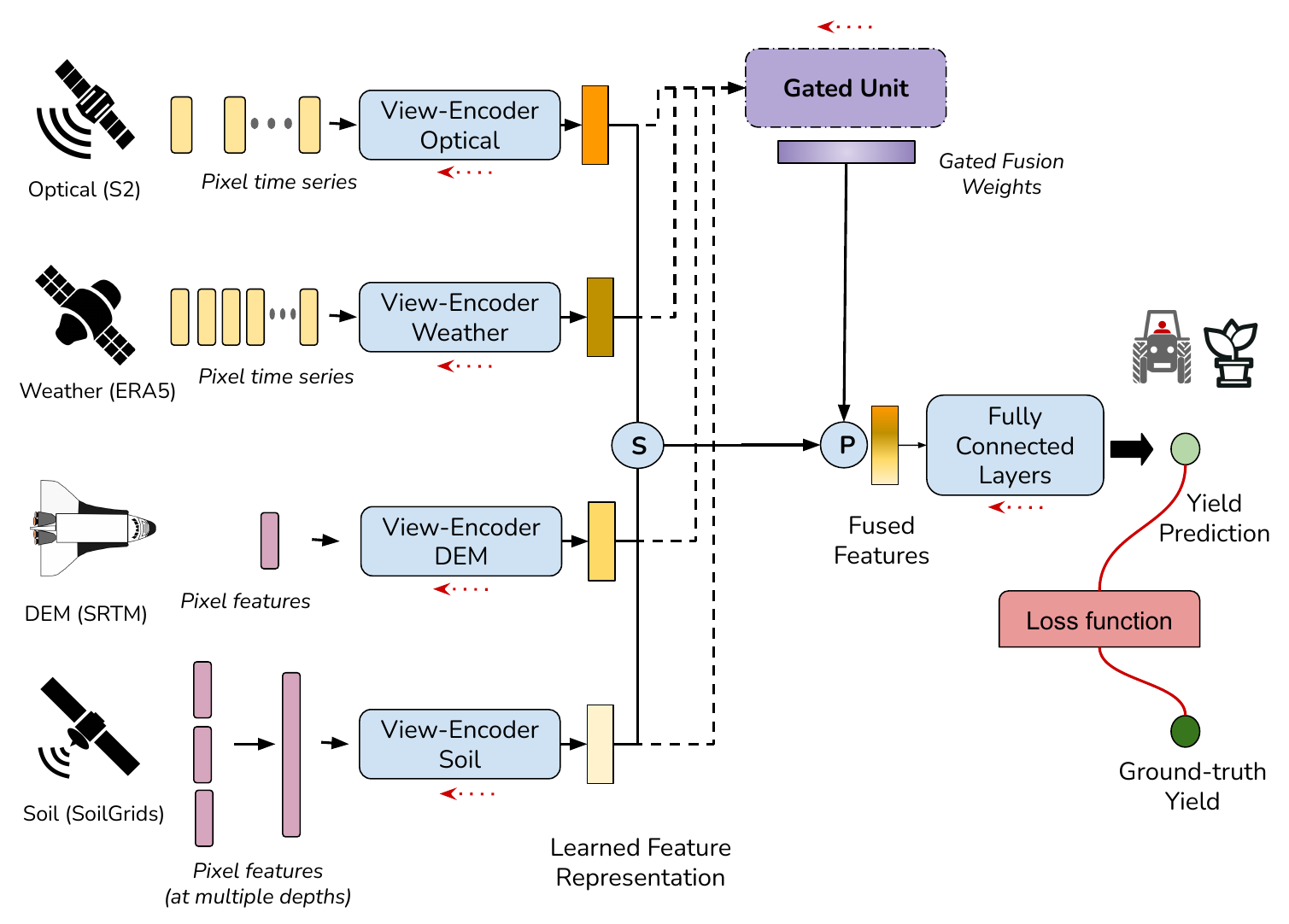}
    \caption{Illustration of the proposed Multi-View Gated Fusion (\gatedfusion) model with the four views used. ``S'' represents the vector stacking operation, and ``P'' the dot product. The forward pass is shown with a black arrow, while the dotted arrow shows the additional connections for the GU. The model is learned end-to-end by comparing the prediction with the ground truth. The red dotted arrows illustrate the backward pass of the loss function through the model components.} \label{fig:mvgf:illustration}
\end{figure*} 
After obtaining the fused representation, it is fed to fully connected neural network layers. These layers, represented by a function $F_{\theta_{\text{F}}} : \mathbb{R}^d \rightarrow \mathbb{R}^1 $ parameterized by $\theta_{\text{F}}$, serve as a prediction head to estimate the crop yield for each pixel $i$: $\hat{y}^{(i)} = F_{\theta_{\text{F}}}\left(\vect{z}_{\text{F}}^{(i)}\right)$. 
Since we are designing a predictive model that is fed with the multi-view data and has all the previous components, $ \hat{y}^{(i)} = P_{\Theta} (\mathcal{X}^{(i)})$ with $\Theta = \left\{ \theta_{\text{S2}}, \theta_{\text{W}}, \theta_{\text{D}}, \theta_{\text{S}}, \theta_{\text{G}}, \theta_{\text{F}} \right\}$, we could minimize a loss function $\mathcal{L}$ over training pixels to learn it end-to-end. We use the following
\begin{equation}
\mathcal{L}(\Theta; \mathcal{D}) = \frac{1}{N} \sum_{i=1}^N \text{MSE}\left(y^{(i)}, P_{\Theta}(\mathcal{X}^{(i)}) \right) \ ,
\end{equation}
with the mean squared error (MSE) as a loss function, $\text{MSE}(y, \hat{y})=({y} - \hat{y})^2$. 
We named our model that unifies these different components (view-encoders, the GU, and a prediction head) as Multi-View Gated Fusion (\gatedfusion), see Fig.~\ref{fig:mvgf:illustration} for an illustration.

%% file: content/experiments.tex
\section{Experiments} \label{sec:experiments}

\subsection{Experimental Settings} \label{sec:experiments:set}
\subsubsection{Data Preparation}
To avoid overfitting to the different magnitudes and scale of the multi-view data, we re-scale each numerical band or feature in the input data into a [0,1] range \citep{zhang2021information}: $x_{\text{norm}} = (x - \text{min}(x)) / (\text{max}(x) - \text{min}(x))$. For the \sentsource \sent images, we calculate the maximum and minimum across time and samples, omitting values associated with clouds, snow, and noise. In addition, for dynamic views (\sent and \weat), we pad the sequences with a masked value (-1). 
For the categorical information contained in the \sentsource-based SCL, we codify the 12 labels as a one-hot encoding vector, with one additional category for the padding. 
In the \sentsource data, the sequence length is 150 including padding, while for the \weat is 500.

For the \sentsource \sent data, we use two different versions. \textbf{\sentraw} as the raw SITS (including the SCL) up to 5-days temporal resolution with the padding previously mentioned. 
\textbf{\sentmonth} as a monthly-based sampling SITS to use with the input-level fusion as described in \cite{miranda2023}. 
In the second setting, a sample for each month is selected based on the lowest cloud coverage, across a two calendar year (24 time-steps) period. The two years are defined by seeding and harvesting, so that the harvesting month always falls in the second year \citep{helber_crop_2023}.
Additionally, we mask samples outside the growing season in the 24 time-step representation.

\subsubsection{Compared Methods} \label{sec:experiments:base}
The Input-level Fusion (IF) approach proposed for sub-field crop yield prediction in \cite{miranda2023} is used for comparison. 
For this, the \sentmonth is used, where the static views (\dem and \soil features) are vectorized (\textit{flattened}) and repeated for each month along the 24 time-step representation. The \weat data is aggregated by summing the daily features between the dates of the selected \sent images in each month.
This generates a multivariate time-series data, where each time-step and pixel has the features from \weat, \dem, and \soil concatenated with the \sentmonth features.
Since \cite{miranda2023} showed that a subset of the views have to be used for a better prediction performance, we follow the best combination that they identified for each dataset.
It is important to note that we trained IF with \sentmonth instead of \sentraw as the \gatedfusion. We do not see a simple and efficient way to perform IF with \sentraw view for a fair comparison.

To compare the fusion effect, two models based on IF are selected: LSTM for IF (\lstminputfusion) and Gradient Boosting Decision Tree for IF (\lgbminputfusion). The \lgbminputfusion model is fed with flattened (across time) vectors.
In addition, to evaluate the benefit of using a MVL scenario, two single-view models are used by feeding either \sentmonth or \sentraw to a LSTM view-encoder.

\subsubsection{Implementation Details} \label{sec:experiments:imp}
The \gatedfusion model handles the four views using different view-encoders.
For \sentsource and \weat views we use RNN-based view-encoders consisting of 2 LSTM layers with 128 units each. We show a few experiments with the Transformer model in Sec.~\ref{sec:analysis:ablation} without achieving significant improvements.
For \dem and \soil we use MLP view-encoders, with 1 hidden layer of 128 units. 
On each view-encoder, there is an output linear layer projecting the data to $d=128$ dimensions.
For the prediction head, we use an MLP with 1 hidden layer of 128 units and an output layer with a single unit. 
As suggested by previous works \citep{chen2017deep,maimaitijiang2020soybean}, we include 30\% of dropout on the view-encoders during training and Batch-Normalization (BN) on all the MLPs.
With this implementation, the compared models have a different number of learnable parameters coming from the view-encoders, since we use the same prediction head. 
For the single-view model with \sentraw, the view-encoder has 228K parameters. 
For the \lstminputfusion, the view-encoder parameters increased to 238K. 
For the \gatedfusion, the parameters of the 4 view-encoders reach to 483K, and 2.1K in the GU. 
These models are trained a maximum of 50 epochs, with an early stopping criterion of 14 epochs patience in the loss function (MSE). This function is optimized with ADAM \citep{kingma2014adam}, a batch-size of 1024, a learning rate of $10^{-3}$, and a weight decay of $10^{-4}$. The used hyperparameters are tuned manually in the ARG-S data.

\subsection{Evaluation}
For the evaluation, we use a stratified-group $K$-fold cross validation ($K=10$). The grouping is based on the field identifier and stratified at farm identifier\footnote{Four our study, a farm represents either a set of fields operated by a farmer or geographically nearby fields.}.  This means that all pixels within a field are used either for training or validation. 
We quantitatively evaluate model performance by using standard regression error metrics. The Mean Absolute Error (MAE) in t/ha, Mean Absolute Percentage Error (MAPE) in $\%$, and coefficient of determination ($R^2$) are measured between ground-truth yield values, $y$, and model predictions, $P_{\Theta}(\mathcal{X})$. 
\begin{align}
    MAE &= \frac{1}{N_{\text{val}}} \sum_{i=1}^{N_{\text{val}}} | y^{(i)} - P_{\Theta}(\mathcal{X}^{(i)}) | \\
    MAPE &= \frac{1}{N_{\text{val}}} \sum_{i=1}^{N_{\text{val}}} \frac{| y^{(i)} - P_{\Theta}(\mathcal{X}^{(i)})  |}{y^{(i)}} \\
    R^2 &= 1 - \frac{ \sum_{i=1}^{N_{\text{val}}} ( y^{(i)} - P_{\Theta}(\mathcal{X}^{(i)}) )^2 }{\sum_{i=1}^{N_{\text{val}}} ( y^{(i)} - \bar{y})^2} 
\end{align}
$N_{\text{val}}$ are the number of samples and $\bar{y}$ is the field average in the validation split. These metrics are evaluated at {sub-field level} (pixel-wise comparison, $N_{\text{val}}$= number of pixels) and {field level} (comparison of field-averaged values, $N_{\text{val}}$= number of fields). The aggregated results with the standard deviation across folds are presented. 

Additionally, we conducted a leave-one-year-out (LOYO) cross-validation experiment to compare the crop yield prediction performance for each year. In this experiment, all fields with a harvesting date\footnote{Please note that the growing season of a crop could extend from one year to the following.} in a specific year were chosen for validation. 
In addition, it is important to note that not every crop field is available for a farm throughout all years within the dataset (refer to Fig.~\ref{fig:farm_seedtoharv}). 

\subsection{Overall Results} \label{sec:experiments:res}

\begin{table*}[!t]
    \centering
    \caption{\textbf{Field-level performance}. $R^2$ of the crop yield prediction at {field level} for different models and combination of input views. $^*$The input views for IF models are: \sentmonth and \dem in ARG-S, \sentmonth and \soil in GER-R, and all the views in URU-S and GER-W.
    The highest mean and lowest standard deviation are in bold.}\label{tab:performance:field}
    \small
    \begin{tabularx}{\linewidth}{cc|C|C|C|C} \hline
         Model & Input Views & ARG-S & URU-S & GER-R & GER-W \\ \hline
         LSTM & \sentmonth &$0.74 \pm 0.12$ & $0.69 \pm 0.14$ & $0.65 \pm 0.18$ & $0.60 \pm 0.25$ \\  
         \lgbminputfusion & varies$^*$ & $0.72 \pm 0.14$ & $0.77 \pm 0.08$ & $0.69 \pm \highest{0.09}$ & $0.68 \pm 0.12$ \\ 
         \lstminputfusion & varies$^*$ & $0.82 \pm 0.12$ & $0.74 \pm 0.12$ & $0.78 \pm \highest{0.09}$ & $0.68 \pm 0.28$ \\ \hline
         LSTM & \sentraw & $\highest{0.84} \pm \highest{0.08}$ & $0.77 \pm 0.09$  & $0.77 \pm 0.13 $ & $0.72 \pm 0.11 $\\ 
        \textbf{\gatedfusion} & \sentraw, \weat, \dem, \soil & $\highest{0.84} \pm 0.11$ & $\highest{0.78} \pm \highest{0.07}$ & $\highest{0.80} \pm 0.13$ & $\highest{0.80} \pm \highest{0.09}$ \\ 
         \hline
    \end{tabularx}
\end{table*}

\begin{table*}[!t]
    \centering
    \caption{\textbf{Sub-field level performance}. $R^2$ of the crop yield prediction at {sub-field level} for different models and combination of input views.  $^*$The input views for IF models are: \sentmonth and \dem in ARG-S, \sentmonth and \soil in GER-R, and all the views in URU-S and GER-W.
    The highest mean and lowest standard deviation are in bold.}\label{tab:performance:subfield}
    \small
    \begin{tabularx}{\linewidth}{cc|C|C|C|C} \hline
         Model & Input Views & ARG-S & URU-S & GER-R & GER-W \\ \hline
         LSTM & \sentmonth & $0.61 \pm 0.11$ & $0.38 \pm 0.08$ & $0.35 \pm 0.13$ & $0.32 \pm 0.09$  \\  
         \lgbminputfusion & varies$^*$ & $0.58 \pm 0.11$ & $\highest{0.42} \pm 0.07$ & $0.42 \pm 0.08$ &  $0.37 \pm 0.10$\\ 
         \lstminputfusion & varies$^*$ & $0.65 \pm 0.08$ & $0.41 \pm 0.07$ & $0.45 \pm \highest{0.10}$ & $0.37 \pm 0.12$ \\ \hline
         LSTM & \sentraw &  $0.67 \pm \highest{0.05}$ & $0.41 \pm \highest{0.06}$ & $\highest{0.46} \pm 0.11$ & $0.41 \pm \highest{0.08}$ \\
        \textbf{\gatedfusion} & \sentraw, \weat, \dem, \soil & $\highest{0.68} \pm \highest{0.05}$ & $\highest{0.42} \pm \highest{0.06}$ & $\highest{0.46} \pm 0.15$ & $\highest{0.44} \pm 0.10$ \\ 
         \hline
    \end{tabularx}
\end{table*}
The aggregated $R^2$ results for all datasets are displayed in Table~\ref{tab:performance:field} for field level and in Table~\ref{tab:performance:subfield} for sub-field level.
We observe that the proposed \gatedfusion obtains the best performance across all datasets and metrics regarding the compared methods. The same evidence is observed in the other regression metrics (See Tables~\ref{tab:performance:arg_s}, \ref{tab:performance:uru_s}, \ref{tab:performance:ger_r},  \ref{tab:performance:ger_w} in~\ref{sec:app_results:perdata}).
The $R^2$ is around $0.80$ across all datasets at the field level, while at sub-field level it is $0.68$ for ARG-S and around $0.44$ for the rest. 
The lower values of sub-field level $R^2$ compared to field level reflects the complexity of predicting the crop yield to the high spatial resolution of 10 m/px.

For the \gatedfusion, the best-performing combination of views is obtained by using all the views (Table~\ref{tab:ablation:modalities}), in contrast to IF, where the best-performing combination is obtained with a sub-set of all the views, depending on the country and crop-type \citep{miranda2023}. For instance, for ARG-S the best combination is \sentmonth and \dem views, and for GER-R it is \sentmonth and \soil. 
These results exhibit the effect of the \gatedfusion to adaptively fuse the features based on the information of each sample. We discuss this in more detail in Sec.~\ref{sec:analysis:ablation}.  
Additionally, this shows the effectiveness of the proposed MVL approach compared to the other approaches based on IF.
When comparing the \gatedfusion to the single-view model with \sentraw, we observe high and small prediction improvements, reflecting that the benefit obtained from the MVL scenario relies on each setting.

Different results are obtained in each dataset. GER-W fields is where \gatedfusion obtains the greatest improvements as compared to the IF-based models, with $0.12$ and $0.07$ points for field and sub-field $R^2$, while in ARG-S and GER-R minor improvements are obtained, around $0.02$ and $0.01$ points in $R^2$. A similar pattern is observed when comparing the \gatedfusion with the single-view models. 
The datasets also influence the model convergence. Whereas \lstminputfusion converges in average (across folds) in less than $23$ epochs, 
\gatedfusion need $29$, $25$, $39$, and $36$ for ARG-S, URU-S, GER-R and GER-W respectively. 


\subsection{Validation Results Per Spatial Coverage}  \label{sec:experiments:coverage}

\begin{figure*}[t!]
\centering
\subfloat[Fields in ARG-S data.\label{fig:perf_coverage:a}]{\includegraphics[width=0.49\linewidth]{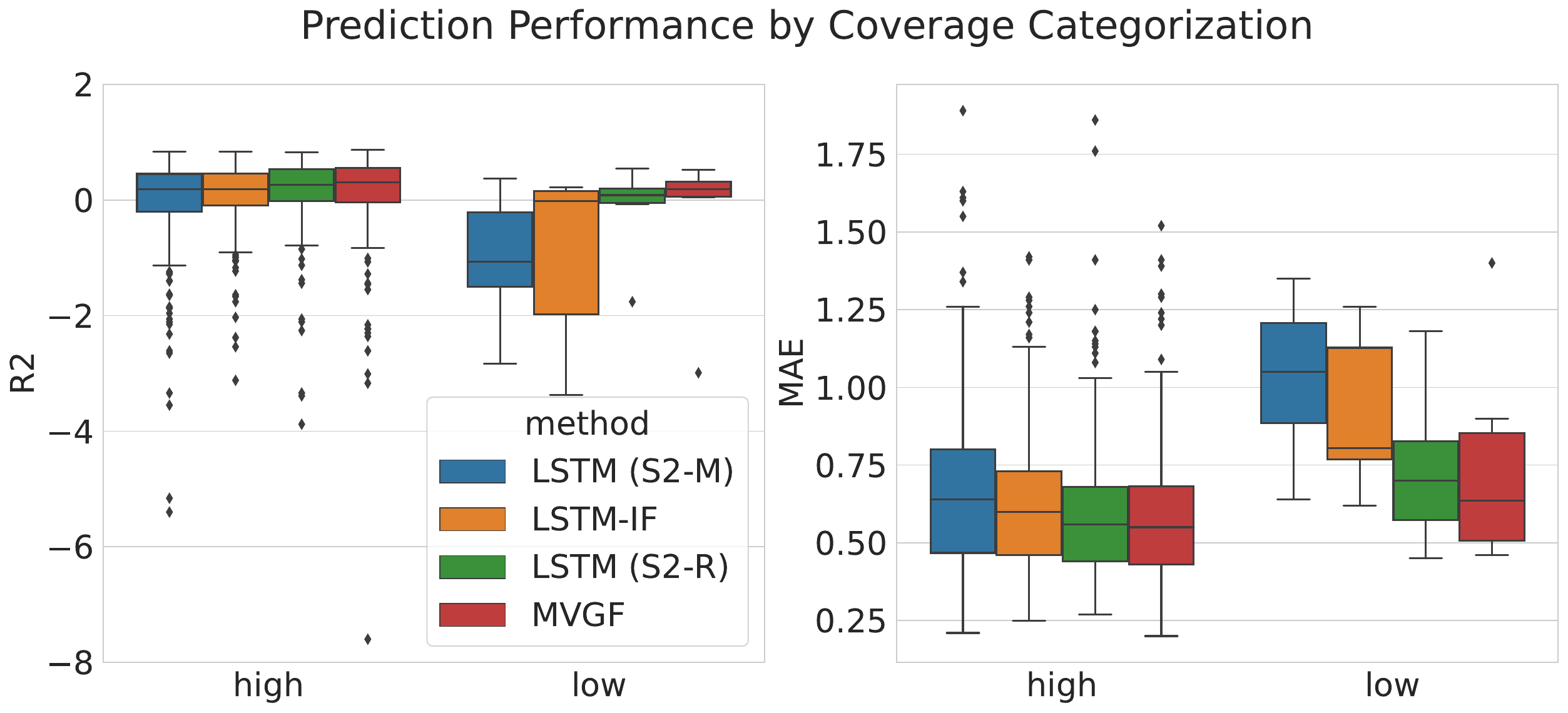}} 
\hfill
\subfloat[Fields in URU-S data.\label{fig:perf_coverage:b}]{\includegraphics[width=0.49\linewidth]{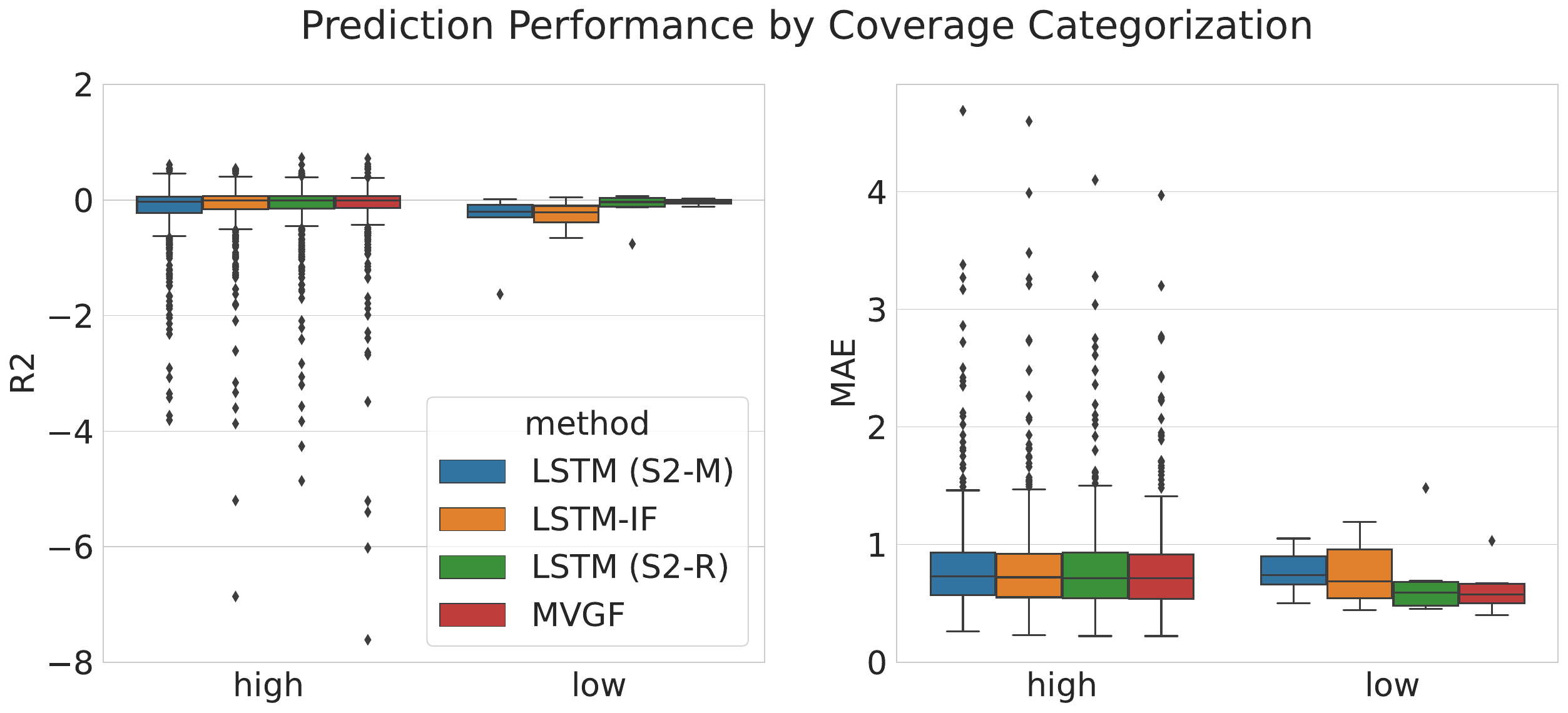}}\\ 
\subfloat[Fields in GER-R data.\label{fig:perf_coverage:c}]{\includegraphics[width=0.49\linewidth]{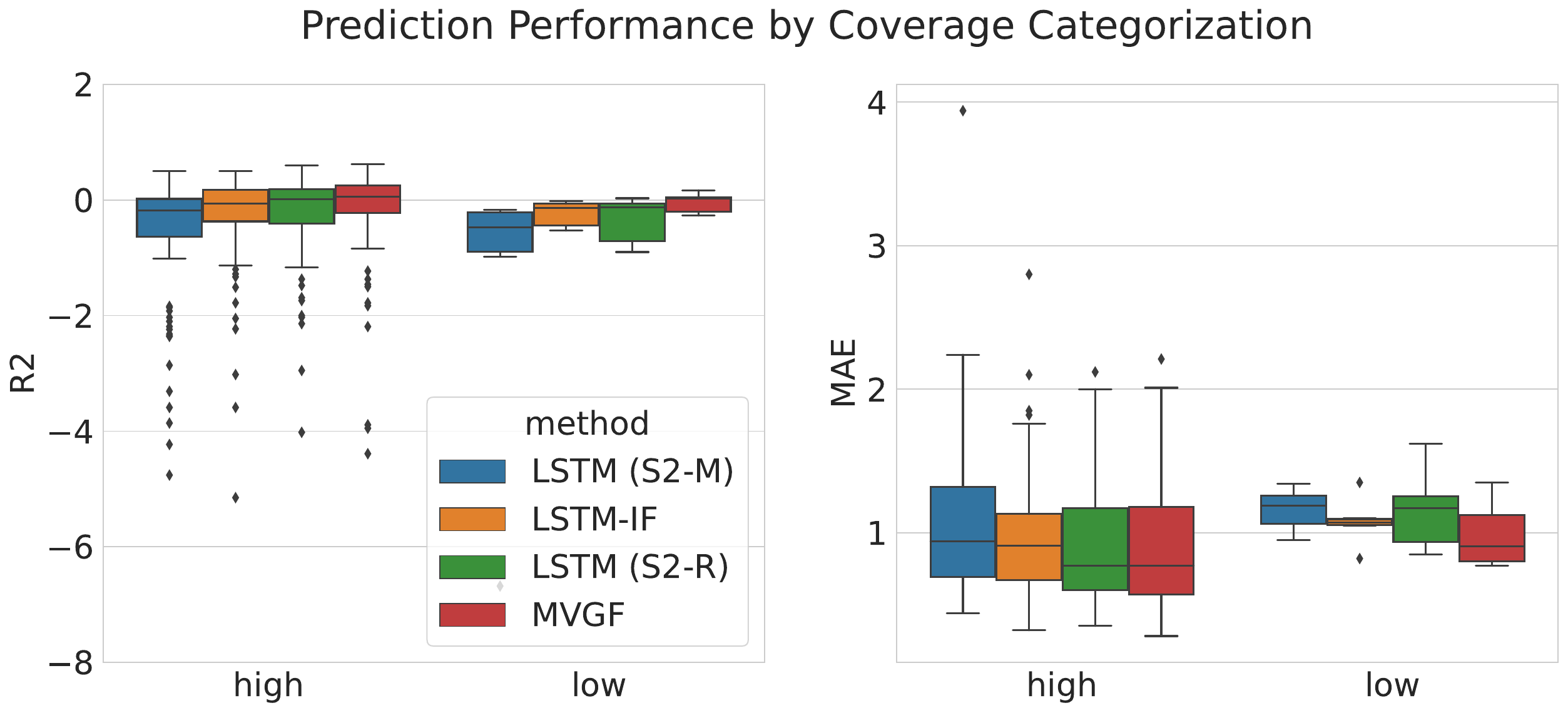}} 
\hfill
\subfloat[Fields in GER-W data.\label{fig:perf_coverage:d}]{\includegraphics[width=0.49\linewidth]{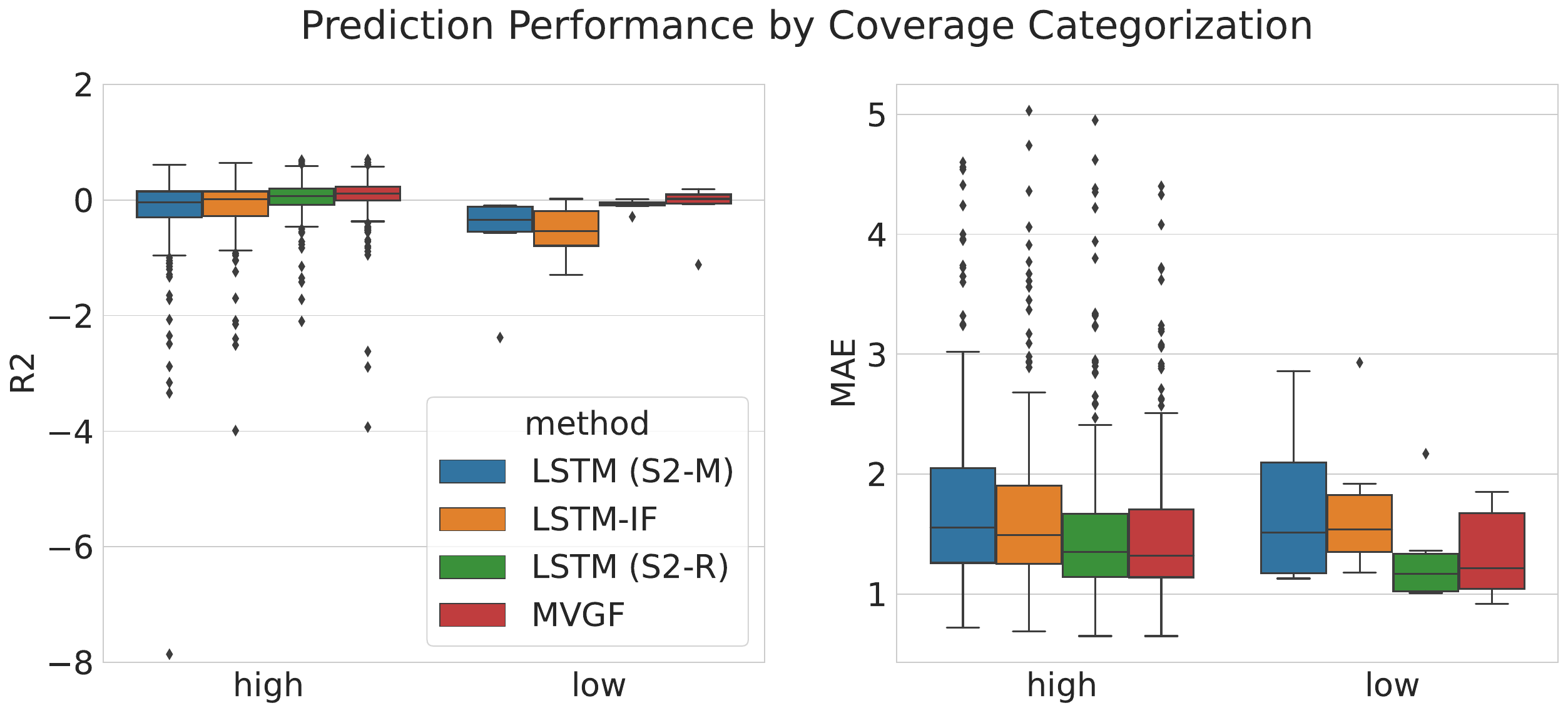}}\\ 
\caption{Prediction performance of field-level $R^2$ and MAE metrics on two {spatial coverage categorization} (high and low). ``High'' considers fields with spatial coverage above the 5th lowest spatial coverage, while ``Low'' considers fields with a spatial coverage same or below that threshold.}\label{fig:perf_coverage}
\end{figure*}  
We split the validation fields into high and low coverage based on the field spatial coverage calculated with the \sentsource-based SCL (see Sec.~\ref{sec:data:mv:s2}). The field with the 5th lowest spatial coverage\footnote{Since some fields have the same spatial coverage, there could be more than 5 fields categorized as low coverage.} is selected as a threshold in each dataset, therefore the high/low categorization is relative to the country and crop-type. These \sent images with low coverage caused by external factors (e.g. cloudy conditions and snow, and noises) are commonly mentioned as data with missing information \citep{shen2015missing}.
Fig.~\ref{fig:perf_coverage} displays $R^2$ and MAE metrics for these grouped fields, comparing the LSTM-based models to our model. 
For all the datasets, it can be seen that the prediction performance in high coverage fields is similar between the models, and that the main difference (and greater error) is actually coming from the fields with more missing optical information.
In these ``difficult'' fields for crop yield prediction is where the proposed \gatedfusion takes most advantage of the additional information given in the multi-view data, obtaining errors among the lowest for the compared methods.
This pattern is more clear in ARG-S fields, which aligns with the higher \shortGFweights given by the proposed model to the \sentsource-based \sent view  (Sec.~\ref{sec:analysis:weights}).
This suggests that a better way of combining the multi-view data particularly benefits the case where one of the views has a higher chance of missing information.
For the case study, the additional views (\weat, \soil, and \dem) could be supplementing the missing \sent information which is prone to be affected by external factors \citep{shen2015missing}.

\subsection{Validation Results Per Year} \label{sec:experiments:loyo}

\begin{figure*}[t!]
    \centering
    \includegraphics[width=0.85\linewidth]{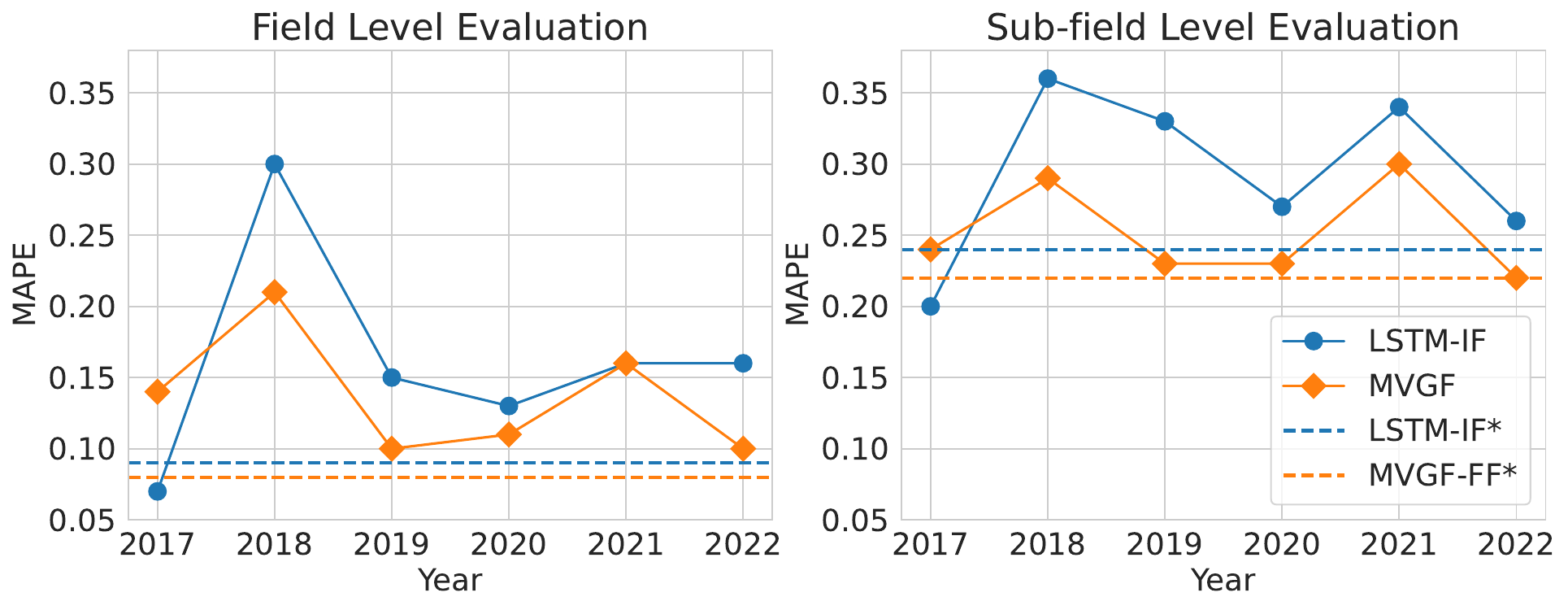} 
    \caption{\textbf{LOYO performance}. MAPE of the crop yield prediction across years in the leave-one-year-out evaluation for ARG-S fields. $^*$As a comparison, the dashed lines show the model performance in the stratified 10-fold cross-validation. The \lstminputfusion use \sentmonth with \dem views, while the \gatedfusion use all views (\sentraw, \weat, \dem, \soil).}
    \label{fig:performance:arg_s:yearly}
\end{figure*}
For the LOYO cross-validation, we focus on comparing \lstminputfusion and \gatedfusion in the ARG-S fields, since the best sub-field level performance was obtained in that data (Sec.~\ref{sec:experiments:res}). In ARG-S data, there are 8, 12, 19, 50, 73, and 27 (validation) fields respectively for each year from 2017 to 2022.
The best input views configuration is used for \lstminputfusion, which is \sentmonth with \dem, while for \gatedfusion is all the views (\sentraw, \weat, \dem, \soil). 
Since each year has a different crop yield distribution, we focus on MAPE as a metric that normalizes the magnitude of the target variable. A summary of the MAPE is in Fig.~\ref{fig:performance:arg_s:yearly}, and Table~\ref{tab:performance:arg_s:yearly} includes additional metrics in the appendix. 

In Fig.~\ref{fig:performance:arg_s:yearly} we observe that the \gatedfusion model performs better than the \lstminputfusion, as the previous overall evaluation reflects. This holds true for the MAPE across all evaluated years except for 2017 and 2021 at field level metric. 
Whereas the \gatedfusion model obtains the best yield prediction performance in the most recent year (2022), the most difficult years for prediction are 2018 at field level and 2021 at sub-field level.
As a reference, the 10-fold cross-validation mean performance is shown with dashed lines. Since the models perform better in the standard cross-validation, it illustrates the difficulty of the LOYO evaluation. This means that, for our case study, learning to predict an unseen year is harder than focusing on a random (stratified) prediction.
We expect this variable pattern, since each agricultural year could have a complete different behavior.

%% file: content/analysis.tex
\section{Analysis} \label{sec:analysis}

In the following, we present some visualizations of the \gatedfusion model in Section~\ref{sec:analysis:modelvis}, an analysis of the \GFweights in Section~\ref{sec:analysis:mvgf_lr}, and an ablation study in Section~\ref{sec:analysis:ablation}.

\subsection{Model-based Visualization} \label{sec:analysis:modelvis}

\begin{figure*}[t!]
\centering
\subfloat[Predictions of \lstminputfusion model with \sentmonth and \dem input-views.\label{fig:fieldex:1:if}]{\includegraphics[width=0.9\linewidth]{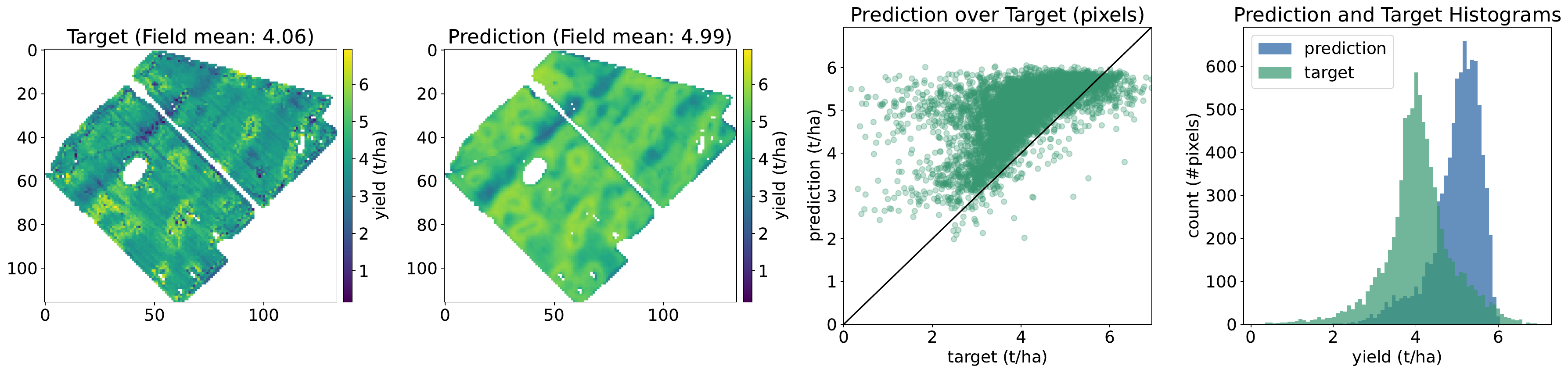}} \\
\subfloat[Predictions of \gatedfusion model with \sentraw, \weat, \dem, and \soil input-views.\label{fig:fieldex:1:mvgf}]{\includegraphics[width=0.9\linewidth]{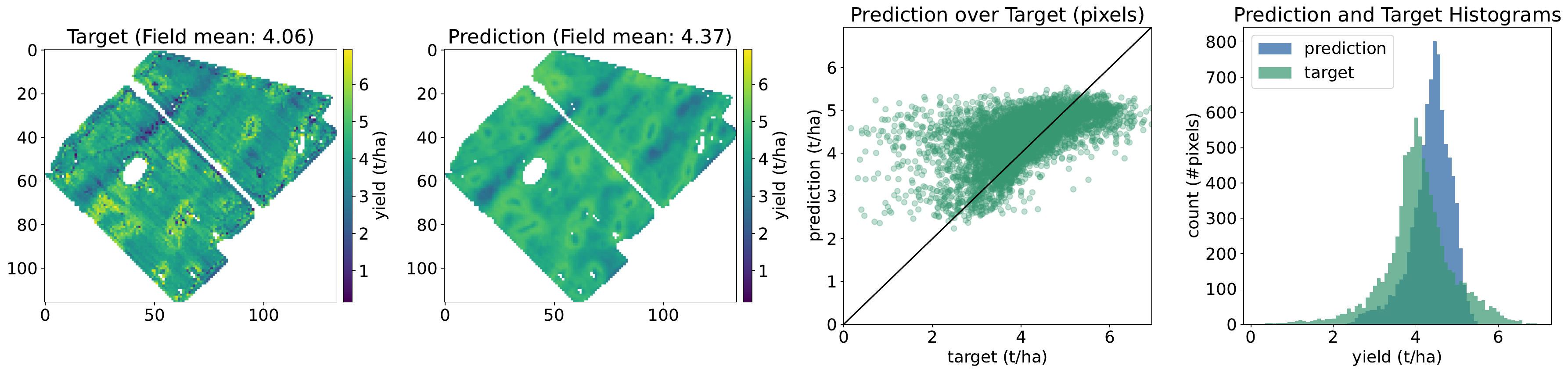}}
\caption{Crop yield prediction map for a sample field in the ARG-S data. Two models are compared: \lstminputfusion and \gatedfusion. The columns from left to right are the ground truth yield map, the predicted yield map, prediction and target scatter, plot of prediction (blue) and target (green) distribution.}\label{fig:fieldex:1}
\end{figure*}   
\paragraph{Yield Map Prediction Visualization}  \label{sec:analysis:visualization}
Since we use a pixel-wise prediction approach for the field images, we visualize the field prediction to qualitative inspect if the model learns the in-field variability. 
In Fig.~\ref{fig:fieldex:1} we compare the yield map predictions of two fusion models (\lstminputfusion and \gatedfusion) for a single field in ARG-S data.
The proposed \gatedfusion (Fig.~\ref{fig:fieldex:1:mvgf}) predicts a better yield map (first and second column in the plot) than the \lstminputfusion model (Fig.~\ref{fig:fieldex:1:if}). 
Indeed, the \lstminputfusion tends to predict higher yield values for the selected field (third column in the plot), while the predictions of the \gatedfusion are closer to the target data.
This is also observed with the better yield distribution alignment (fourth column in the plot) by the \gatedfusion model (Fig.~\ref{fig:fieldex:1:mvgf}).
Furthermore, we calculated the Bhattacharyya coefficient ($\rho$, \cite{ray_theoretical_1989}) 
between the target and predicted yield distribution of each field. The $\rho \in [0,1]$ provides a measure of the overlap between two distributions. The aggregated $\rho$ value for all the validation fields is $0.808$ and $0.824$ for \lstminputfusion and \gatedfusion models respectively.
This illustrates that, in the proposed pixel-wise mapping approach, an acceptable in-field variability can be learned. The evidence is also observed in the other datasets and fields (see Fig.~\ref{fig:fieldex:ap1}, \ref{fig:fieldex:ap2}, \ref{fig:fieldex:ap3}, and \ref{fig:fieldex:ap4} for some examples).

\input{attention_analysis/raw_att_weights}
\paragraph{Gated Fusion Weights Visualization}  \label{sec:analysis:weights}
To acquire insights about what the \gatedfusion model learned for the crop yield prediction, we visualize the \shortGFweights $\alpha_v$ computed by the GU (from \eqref{eq:attention_weights} in Sec.~\ref{sec:methods:gatedfusion}).
To get a general overview, we compare the \shortGFweights distribution across the 10 validation folds in Fig.~\ref{fig:attention_weights}.
We observe that the weights are not perfectly aligned across folds, mainly because each validation fold contains a unique set of fields, and the \shortGFweights are distributed differently from one field to another (Fig.~\ref{fig:att_field_level:app}).
Nevertheless, there is a tendency to assign higher weights to certain input views depending on the country and crop-type, also evident when we examine individual folds (Fig.~\ref{fig:att_dataset_level}).
This visualization suggests that the \gatedfusion model assigns higher \shortGFweights to different input views, within and across the datasets: \sentsource-based \sent view is particularly high in the Argentina fields, \dem view is predominant in Uruguay, while \soil and \weat views share high weights in the crops of German fields.  This crop-dependent behavior has also been observed in the attention weights computed along the temporal dimension for a crop classification use-case \citep{garnot2020lightweight}.

\subsection{Gated Fusion Weights Analysis with Linear Regression} \label{sec:analysis:mvgf_lr}

We further investigate the influence of the \GFweights in the GU, as well as the contribution of each view in the final prediction, by retraining a variation of the \gatedfusion model.
For this, we replaced the MLP in the prediction head with a linear layer, with weight parameters $\vect{w} \in \mathbb{R}^d$ and bias parameter $b \in \mathbb{R}^1$, that is fed with the fused representation $\vect{z}_{\text{F}} \in \mathbb{R}^d$. We refer to this model as \gatedfusion-LR (MVGF with Linear Regression). 
The prediction head (now a linear regression) allows us to interpret the contribution of each view. Given the extracted features from each view-encoder $\vect{z}_{\text{S2}},  \vect{z}_\text{W}, \vect{z}_\text{D}, \vect{z}_\text{S}$ (the view-representation) and the corresponding \GFweights $\alpha_{\text{S2}}, \alpha_{\text{W}}, \alpha_{\text{D}}, \alpha_{\text{S}}$, the expression for the crop yield prediction can be written as follows:
\begin{equation}\label{eq:pred_yield}
    \hat{y} = \vect{w}^{\top} \cdot \vect{z}_{\text{F}} + b = \sum_{i=1}^{d} w_i \times z_{\text{F}, i} + b ,
\end{equation}
where, if we expand $\vect{z}_{\text{F}}$ as the weighted sum of the view-representations \eqref{eq:weighted_sum}, we obtain:
\begin{align}
    \hat{y} &= \sum_{v} \vect{w}^{\top} \cdot \alpha_{v} \vect{z}_{v} + b \\
    &= \alpha_{\text{S2}} \vect{w}^{\top} \cdot \vect{z}_{\text{S2}}+ \alpha_{\text{W}} \vect{w}^{\top} \cdot \vect{z}_{\text{W}} + \alpha_{\text{D}} \vect{w}^{\top}\cdot \vect{z}_{\text{D}} + \alpha_{\text{S}} \vect{w}^{\top} \cdot  \vect{z}_{\text{S}} + b.
\end{align}

\begin{figure*}[t!]
\centering
\includegraphics[width=0.55\linewidth]{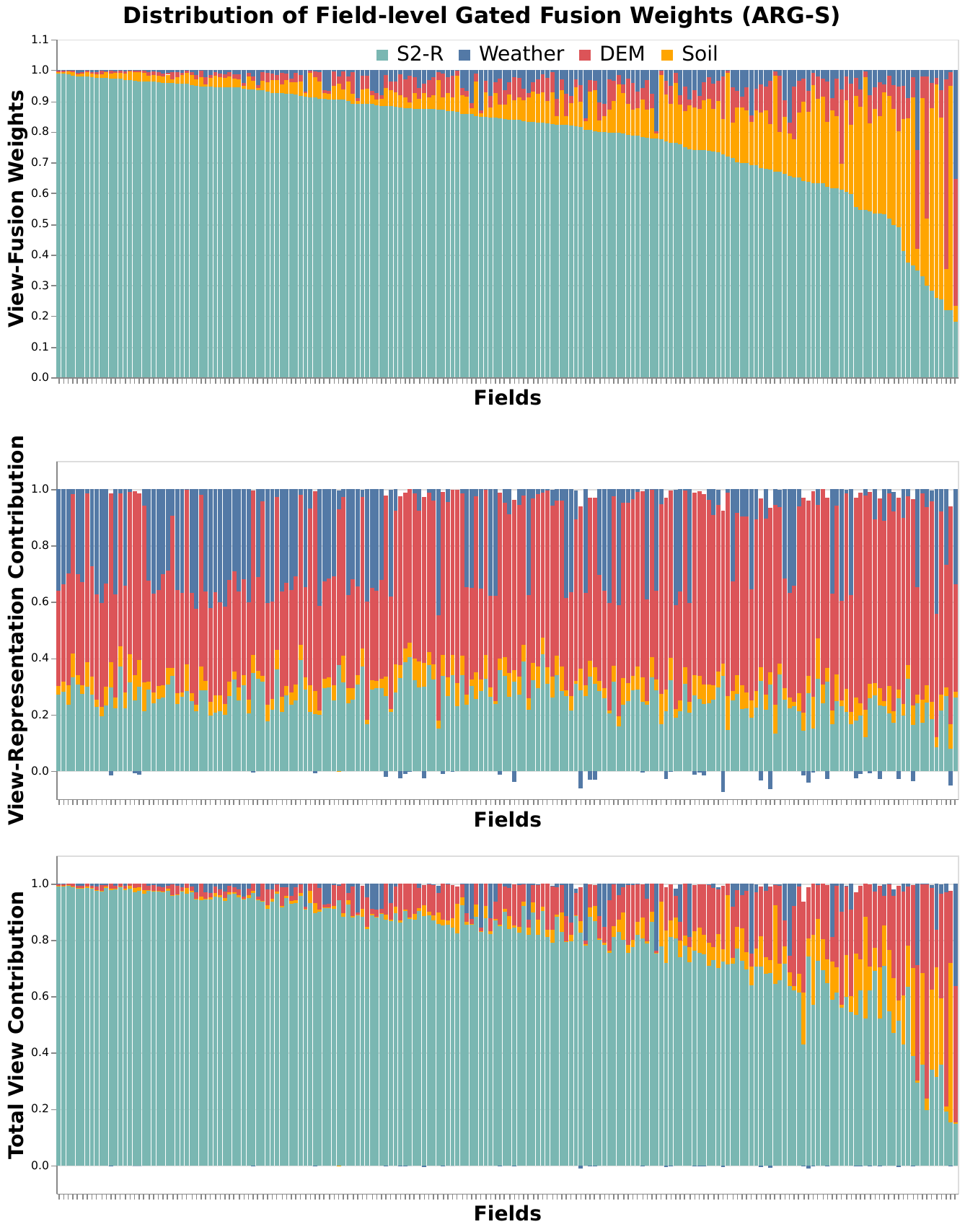}
\caption{Gated fusion weights from the GU ($\alpha_v$) at the top, the contribution before applying the \shortGFweights ($C_v$) at the middle, and the final contribution in the predicted yield ($\alpha_v C_v$) at the bottom. The values of the \gatedfusion-LR model in \textbf{ARG-S} data are aggregated (averaged) at field level using the best performance split. The $C_{v}$ and $\alpha_v C_{v}$ are scaled by $1/\sum_v|C_v|$ and $1/\sum_v|\alpha_v C_v|$ respectively for better visualization.
The field bars are ordered (from left to right) in descending order of the weight given to the predominant view (\sentraw).}\label{fig:att_weight_comparison:app}
\end{figure*} 

Consider $ C_{v} = \vect{w}^{\top} \cdot \vect{z}_{v}$ and $\alpha_v C_{v}$ as the contribution to the prediction by a particular view $v$ before and after applying the \shortGFweights respectively.
We train this model with the same cross-validation settings used for the ARG-S data. However, to mitigate the variability in the results stemming from the multiple splits, we focus solely on a single split --- the one yielding the highest sub-field $R^2$ score.
We illustrate the $\alpha_v$, $C_{v}$, and $\alpha_v C_{v}$ aggregated at field level\footnote{We assume that within each field, fusion weights for different views are roughly aligned, as validated in Fig.~\ref{fig:att_field_level:app}.} for different fields in Figure~\ref{fig:att_weight_comparison:app}. 
This analysis suggests that the learned \shortGFweights ($\alpha_v$) linearly scales the contribution of each view ($C_v$) and yet allows each view-encoders to learn representations ($\vect{z}_v$) with distinct scales.
This experiment demonstrates that the GU module has a tendency to assigns higher weights to the \sent view compared to other views. 
However, we can see how the learned \shortGFweights complement each other in some fields. For instance, when the GU assigns a small value to the \sent view (right side of Fig.~\ref{fig:att_weight_comparison:app}), \soil view shares the largest portion of the \shortGFweights (with \dem view in some cases).
This is also reflected in the total view contribution ($\alpha_v C_{v}$) despite a significant view-representation contribution ($C_{v}$) of \dem and \weat views (Fig.~\ref{fig:att_weight_comparison:app} middle and last row). 
We observe a similar pattern in the other fields (see Fig.~\ref{fig:att_weight_comparison:germany} for GER-W and GER-S).

\begin{figure*}[t!]
\centering
\subfloat[Order based on ascending field spatial coverage. \label{fig:att:weight:comparison:order:spatial}]{\includegraphics[width=0.48\linewidth]{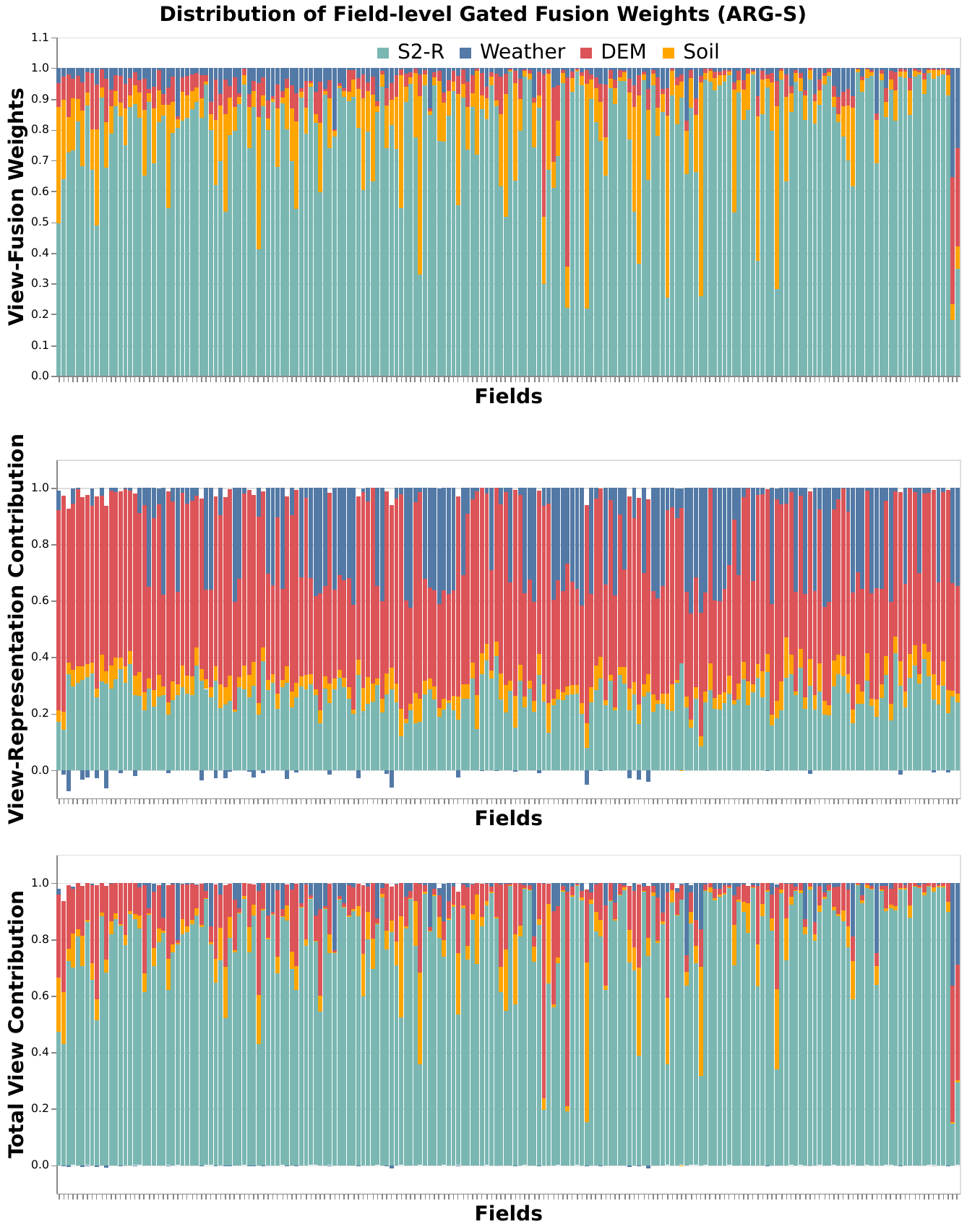}}
\quad
\subfloat[Order based on ascending yield prediction value. \label{fig:att:weight:comparison:order:yield}]{\includegraphics[width=0.48\linewidth]{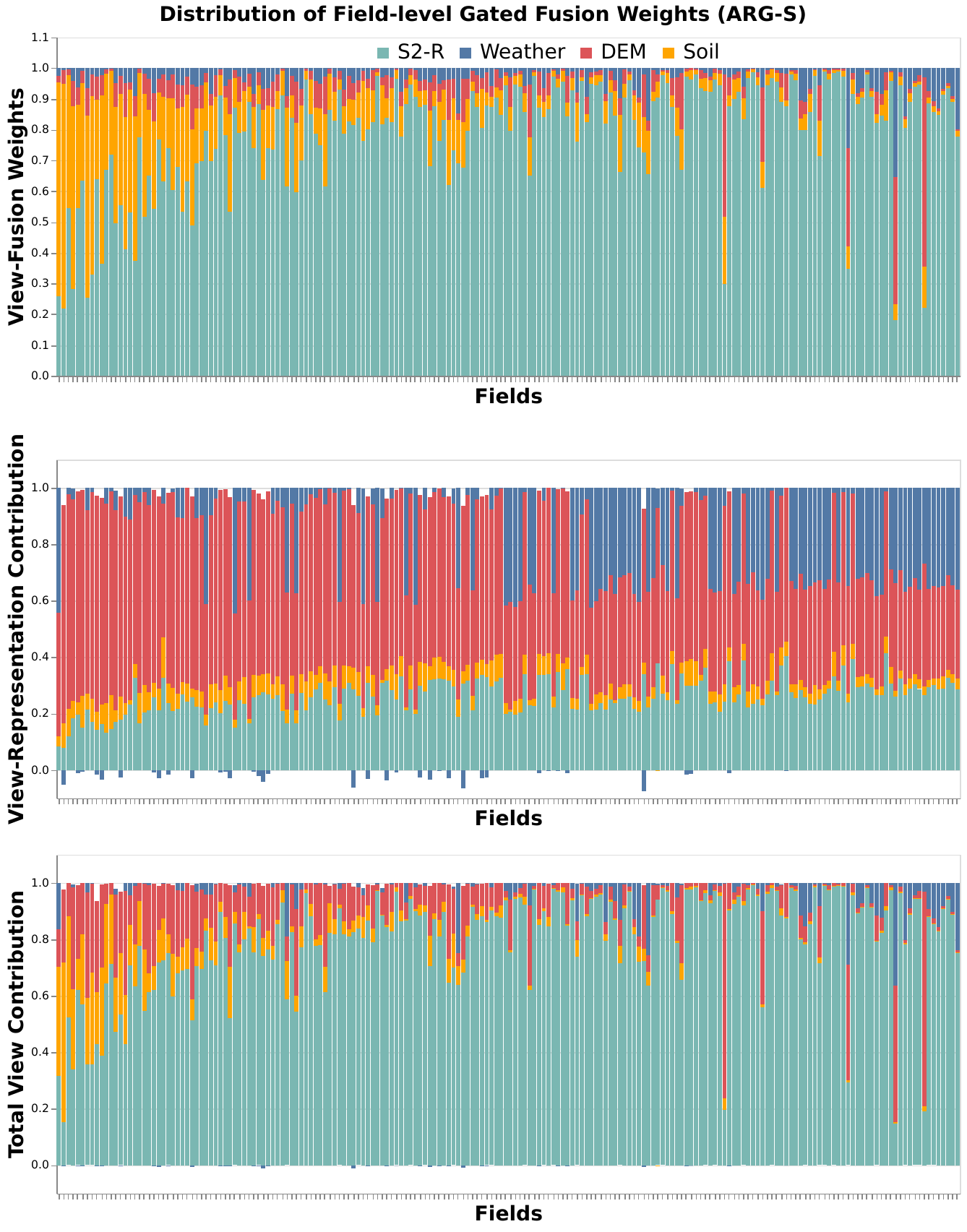}}
\caption{Same plot as Fig.~\ref{fig:att_weight_comparison:app} where the bars (x-axis) are re-ordered from left to right based on different criteria. The values of the \gatedfusion-LR model for \textbf{ARG-S} data are aggregated at field level using the best performance split.}\label{fig:att_weight_comparison:order}
\end{figure*} 
To verify if the field spatial coverage (\sent data availability) has an effect on the learned \shortGFweights assigned to the \sentraw view ($\alpha_{\text{S2}}$), we reorder the fields according to their ascending spatial coverage in Fig.~\ref{fig:att:weight:comparison:order:spatial}, and the prediction value in Fig.~\ref{fig:att:weight:comparison:order:yield}.
We observe that while most of the fields with higher spatial coverage are indeed getting higher \shortGFweights to the \sentraw view, the fields with the lowest spatial coverage are also getting high values. 
However, there is a stronger relation between the yield prediction value and the static views in the ARG-S fields. When the model predicts a low yield value, it tends to assign a high \shortGFweights to the \soil view and high view contribution ($C_v$) to the \dem view. 
Overall, this simplified model version (MVGF-LR) allows us to express the crop yield predictions in terms of individual view-representations and fusion weights, making the model more interpretable for the MVL task.

\subsection{Ablation Study} \label{sec:analysis:ablation}
In the following, we present experiments varying the modeling components of the proposed \gatedfusion. The main purpose is to assess the impact of the individual components that contribute to the overall model performance. 
We focus on the cross-validation setting with the ARG-S fields.

\begin{table*}[!t]
    \centering
    \caption{\textbf{Input view combinations}. Crop yield prediction performance in ARG-S fields with different combinations of views as input data. The LSTM indicates the single-view model with the optical view (S2).
    The best results are in bold.
    }\label{tab:ablation:modalities}
    \footnotesize
    \begin{tabularx}{\linewidth}{Cc|CCC|CCC} \cline{3-8}
          & & \multicolumn{3}{c|}{\textit{Field}}  & \multicolumn{3}{c}{\textit{Sub-Field}}\\ \hline
         Model & Input Views & MAE & MAPE & $R^2$ & MAE & MAPE & $R^2$ \\ \hline
         \multirow{4}{*}{LSTM} & \sentmonth without SCL & $0.40$ & $11$ & $0.74$ & $0.69$ & $25$ & $0.61$  \\ 
          & \sentmonth with SCL & $0.39$ & $11$ & $0.75$ & $0.68$ & $25$ & $0.61$ \\ 
         & \sentraw without SCL & $0.31$ & $9$ & $0.82$& $0.62$& $23$& $0.67$  \\ 
         & \sentraw with SCL & $0.31$ & $9$ & $\highest{0.84}$ & $0.62$ & $23$ & $0.67$  \\ \hline
         \multirow{4}{*}{\gatedfusion} & \sentraw, \weat &  $0.32$ & $9$ & $0.83$ & $0.62$ & $\highest{22}$ & $0.67$  \\ 
          & \sentraw, \dem  & $\highest{0.30}$ & $9$ & $\highest{0.84}$ & $0.62$ & $\highest{22}$ & $0.67$  \\ 
         & \sentraw, \soil & $0.32$ & $9$ & $0.83$ & $0.63$ & $23$ & $0.66$  \\ 
          & \sentraw, \weat, \dem, \soil & $\highest{0.30}$ & $\highest{8}$ & $\highest{0.84}$ & $\highest{0.61}$ & $\highest{22}$ & $\highest{0.68}$ \\ 
         \hline
    \end{tabularx}
\end{table*} 
\begin{table*}[!t]
    \centering
    \caption{\textbf{Merge function alternatives}. Crop yield prediction performance in ARG-S fields with different types of merge functions in the feature-level fusion strategy. \textit{Sigmoid} and \textit{Softmax} indicate the activation function applied to the weights before the weighted sum. The \gatedfusion configuration is the one in the last row. The best results are in bold.}\label{tab:ablation:attention}
    \footnotesize
    \begin{tabularx}{\linewidth}{Lc|CCC|CCC} \cline{3-8}
          &  & \multicolumn{3}{c|}{\textit{Field}}  & \multicolumn{3}{c}{\textit{Sub-Field}}\\ \hline
         Approach & Merge & MAE  & MAPE & $R^2$ & MAE & MAPE & $R^2$ \\ \hline
         \multirow{4}{\hsize}{Feature Fusion} & Product & $0.34$ & $10$ & $0.81$ & $0.64$ & $23$ & $0.65$ \\ 
         & Maximum & $0.34$ & $10$ & $0.81$ & $0.63$ & $23$ & $0.67$ \\ 
         & Concat & $0.32$ & $9$ & $0.83$ & $0.62$ & $22$ & $0.67$  \\  
         & Uniform Sum   & $0.32$ & $9$ & $0.82$ & $0.62$ & $23$ & $0.66$  \\ 
         \hline
         \multirow{2}{\hsize}{Adaptive Fusion}  &  (\textit{Sigmoid}) Weighted Sum & $\highest{0.30}$ & $9$ & $0.83$ & $0.62$ & $23$ & $0.67$ \\
                            & (\textit{Softmax}) Weighted Sum  &  $\highest{0.30}$ & $\highest{8}$ & $\highest{0.84}$ & $\highest{0.61}$ & $\highest{22}$ & $\highest{0.68}$ \\ 
         \hline
    \end{tabularx}
\end{table*}
\begin{table*}[!t]
    \centering
    \caption{\textbf{Gated Unit variations}. Crop yield prediction performance in ARG-S fields for different ways of modeling the GU in the \gatedfusion model. The selected \gatedfusion configuration is the one in the last row. The best results are in bold.}\label{tab:ablation:gate}
    \footnotesize
    \begin{tabularx}{\linewidth}{Cc|CCC|CCC} \cline{3-8}
         &  & \multicolumn{3}{c|}{\textit{Field}}  & \multicolumn{3}{c}{\textit{Sub-Field}}\\ \hline
          GU Input & Across & MAE & MAPE & $R^2$ & MAE & MAPE & $R^2$ \\ \hline
         Average & Views+Features & $0.34$& $9$& $0.81$& $0.63$& $23$& $0.66$  \\ 
         Concat & Views+Features  & $0.33$& $9$& $0.81$& $0.63$& $23$& $0.66$  \\  
         Concat & Views & $\highest{0.30}$ & $\highest{8}$ & $\highest{0.84}$ & $\highest{0.61}$ & $\highest{22}$ & $\highest{0.68}$ \\ 
         \hline
    \end{tabularx}
\end{table*}
\begin{table*}[!t]
    \centering
    \caption{\textbf{Regularization contributions}. Crop yield prediction performance in ARG-S fields by varying ML techniques in the \gatedfusion model. The selected \gatedfusion configuration is the one in the last row. The best results are in bold.
    }\label{tab:ablation:model_comp}
    \footnotesize
    \begin{tabularx}{\linewidth}{c|CCC|CCC} \cline{2-7}
         & \multicolumn{3}{c|}{\textit{Field}}  & \multicolumn{3}{c}{\textit{Sub-Field}}\\ \hline 
         Technique & MAE & MAPE & $R^2$ & MAE & MAPE & $R^2$ \\ \hline 
         Without Regularization & $0.36$ & $10$ & $0.77$ & $0.64$ & $23$ & $0.64$ \\ 
         With Dropout & $0.35$& $10$& $0.79$& $0.65$& $24$& $0.64$ \\ 
         With BN & $0.35$& $10$& $0.80$& $0.63$& $23$& $0.66$ \\ 
         With BN and Dropout & $\highest{0.30}$ & $\highest{8}$ & $\highest{0.84}$ & $\highest{0.61}$ & $\highest{22}$ & $\highest{0.68}$ \\ 
         \hline
    \end{tabularx}
\end{table*}

In Table~\ref{tab:ablation:modalities} the performance obtained with different input views is compared.
The results illustrate that the \gatedfusion model is stable when feeding different (and heterogeneous) input views. There are other view combinations that also obtain the best performance in some metrics. However, the best result across all metrics is obtained by feeding all the views. This effect is observed in the other datasets as well.
In contrast to previous results in RS literature \citep{bocca2016effect,kang2020comparative,miranda2023}, the \gatedfusion model does not saturate through giving more input data.
This effect could be attributed to the GU in the gated fusion approach, since it could select which views to merge depending on each sample. The motivation of the GU comes from whether it allows passing the information through a channel \citep{arevalo2020gated}, so it can reduce non-relevant information and focus on the most important one.

Motivated by \cite{mena2023} work in crop classification, different merge functions are compared in Table~\ref{tab:ablation:attention}. The product, maximum, concatenation \citep{shahhosseini2021corn} and uniform sum replaced the weighted sum ($\mathsf{M}$ in \eqref{eq:static_merge}) as static merge functions. Additionally, we include the vanilla GU proposal with the sigmoid activation function \citep{arevalo2020gated}.
Nonetheless, our gated fusion version (with the softmax function) obtains the best prediction performance compared to these alternative merge functions, and illustrates the effectiveness of the \gatedfusion in combining the learned features in this MVL scenario. 

In Table~\ref{tab:ablation:gate} different ways to model the GU in the \gatedfusion are compared. Two alternatives are presented: i) calculate and apply \shortGFweights on each feature for each view (feature-specific) instead of the global \shortGFweights for each view, and ii) replace the concatenation function by the average in $\mathsf{M}$ (from \eqref{eq:attention_weights}).
Among these, the best combination (with minor improvements) is the global \shortGFweights with $\mathsf{M}$ as concatenation. The feature-specific \shortGFweights may suffer from overfitting since there is a larger number of parameters in the modeling, as the GU parameters increased from 2.1K (in the global) to 262K (in the feature-specific).

In Table~\ref{tab:ablation:model_comp} different regularization techniques are compared. The results prove that different techniques are essential to avoid model overfitting, especially when having several parameters in the \gatedfusion model.
Concretely, BN contributes more to improving prediction performance than Dropout. 
However, when both technique are combined, the regularization techniques boost their individual improvements and obtains the best result. 

\begin{table*}[!t]
    \centering
    \caption{\textbf{View-encoder alternatives}. Crop yield prediction performance in ARG-S fields by varying the temporal view-encoder between LSTM and Transformer in the \gatedfusion model. 
    The selected \gatedfusion configuration is the one with LSTM. The best results are in bold.
    }\label{tab:ablation:transformer}
    \footnotesize
    \begin{tabularx}{\linewidth}{Cc|CCC|CCC} \cline{3-8}
        & & \multicolumn{3}{c|}{\textit{Field}}  & \multicolumn{3}{c}{\textit{Sub-Field}}\\ \hline
         Model & Views & MAE & MAPE & $R^2$ & MAE & MAPE & $R^2$ \\ \hline
         Transformer & S2-R+SCL & $0.35$ & $10$ & $0.78$ & $0.63$ & $23$ & $0.66$ \\
         $\rightarrow$ \gatedfusion & all views & $0.33$ & $9$ & $0.79$ & $\highest{0.61}$ & $\highest{22}$ & $0.67$ \\  \hline
         LSTM & S2-R+SCL & $0.31$ & $9$ & $\highest{0.84}$ & $0.62$ & $23$ & $0.67$  \\
         $\rightarrow$ \gatedfusion & all views & $\highest{0.30}$ & $\highest{8}$ & $\highest{0.84}$ & $\highest{0.61}$ & $\highest{22}$ & $\highest{0.68}$ \\ \hline
    \end{tabularx}
\end{table*}
Table~\ref{tab:ablation:transformer} shows the results of the LSTM being replaced with the Transformer model \citep{vaswani2017attention} as the view-encoders of \sentsource and \weat views. Similar to Vision Transformer \citep{dosovitskiy2020image}, we use a class token to extract a single representation from the complete time series.
In sub-field level metrics, the \gatedfusion with Transformer has a similar performance to LSTM, contrary to field-level, where the LSTM has the best performance. The similarity in performance between Transformer-based model and neural networks without attention is something already observed in MVL with RS data \citep{zhao2022cnn}.

%% file: attention_analysis/raw_att_weights.tex
\begin{figure*}[t!]
\centering
\subfloat[Fields in ARG-S data.\label{fig:attention_weights:a}]{\includegraphics[width=0.25\linewidth]{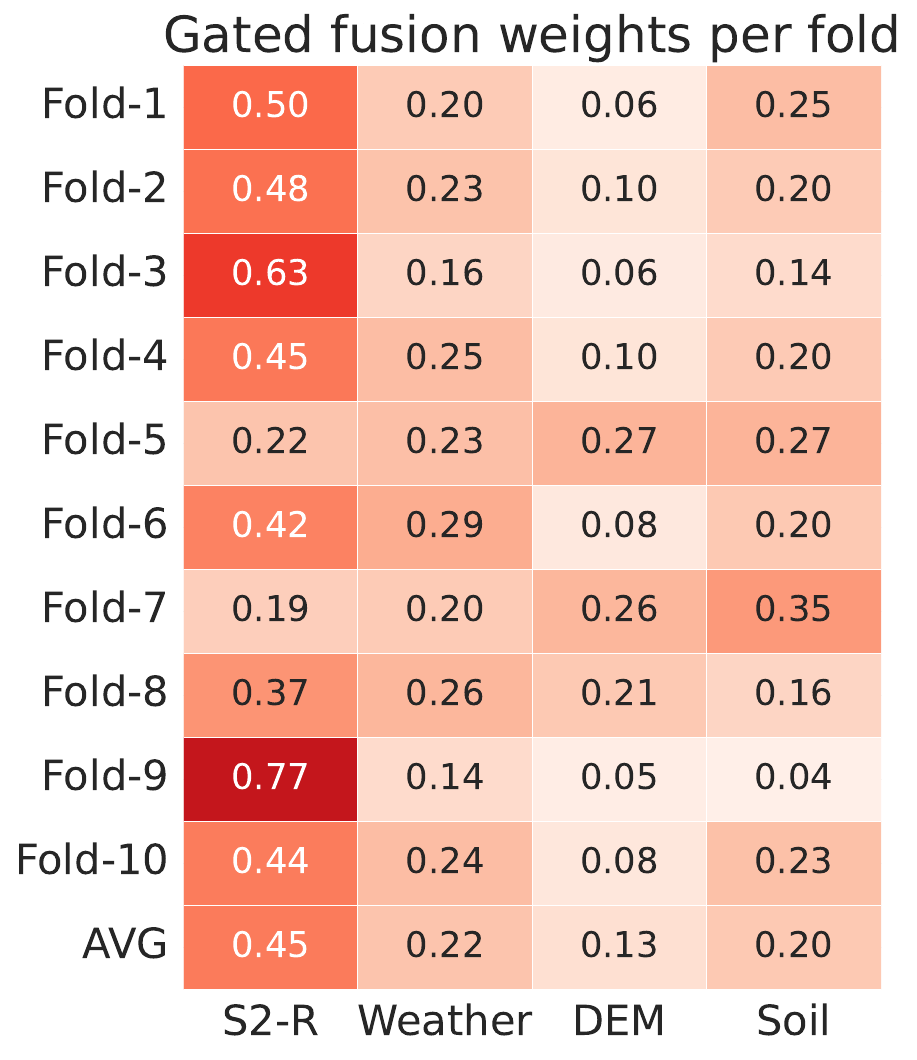}}
\hfill
\subfloat[Fields in URU-S data.\label{fig:attention_weights:b}]{\includegraphics[width=0.25\linewidth]{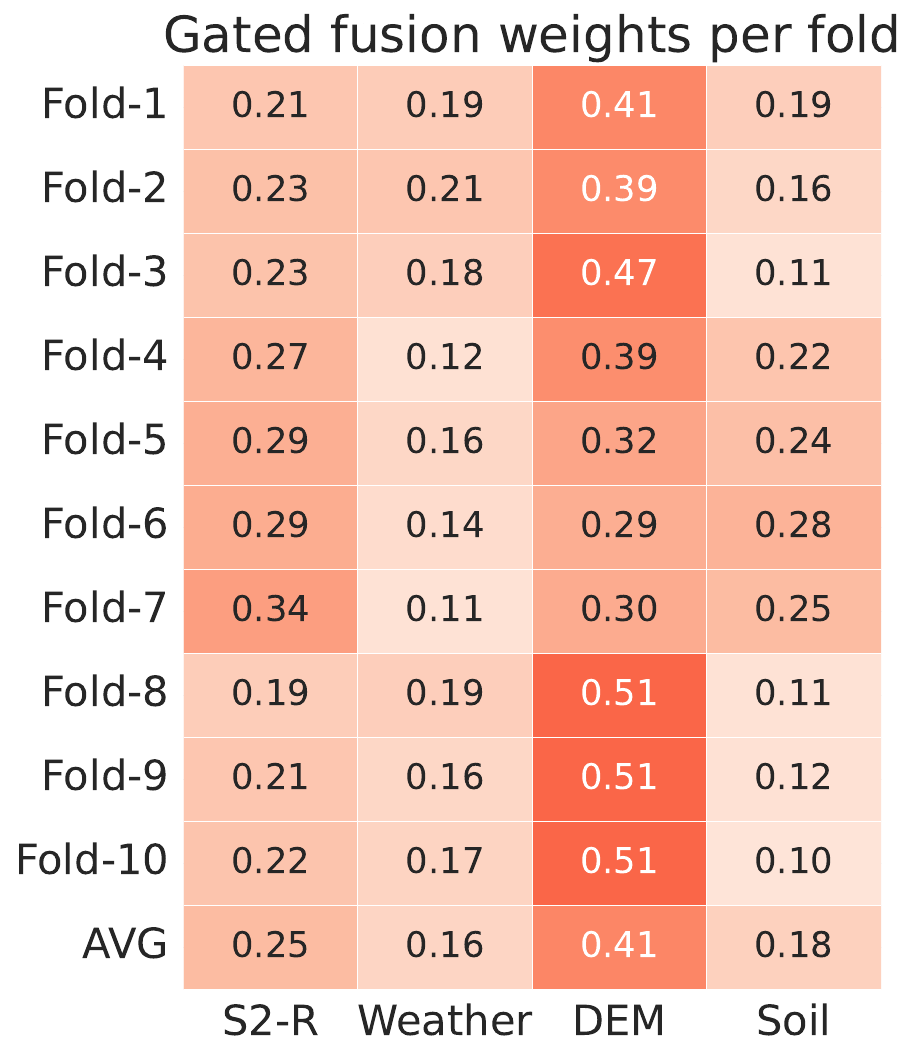}}
\hfill 
\subfloat[Fields in GER-R data.\label{fig:attention_weights:c}]{\includegraphics[width=0.25\linewidth]{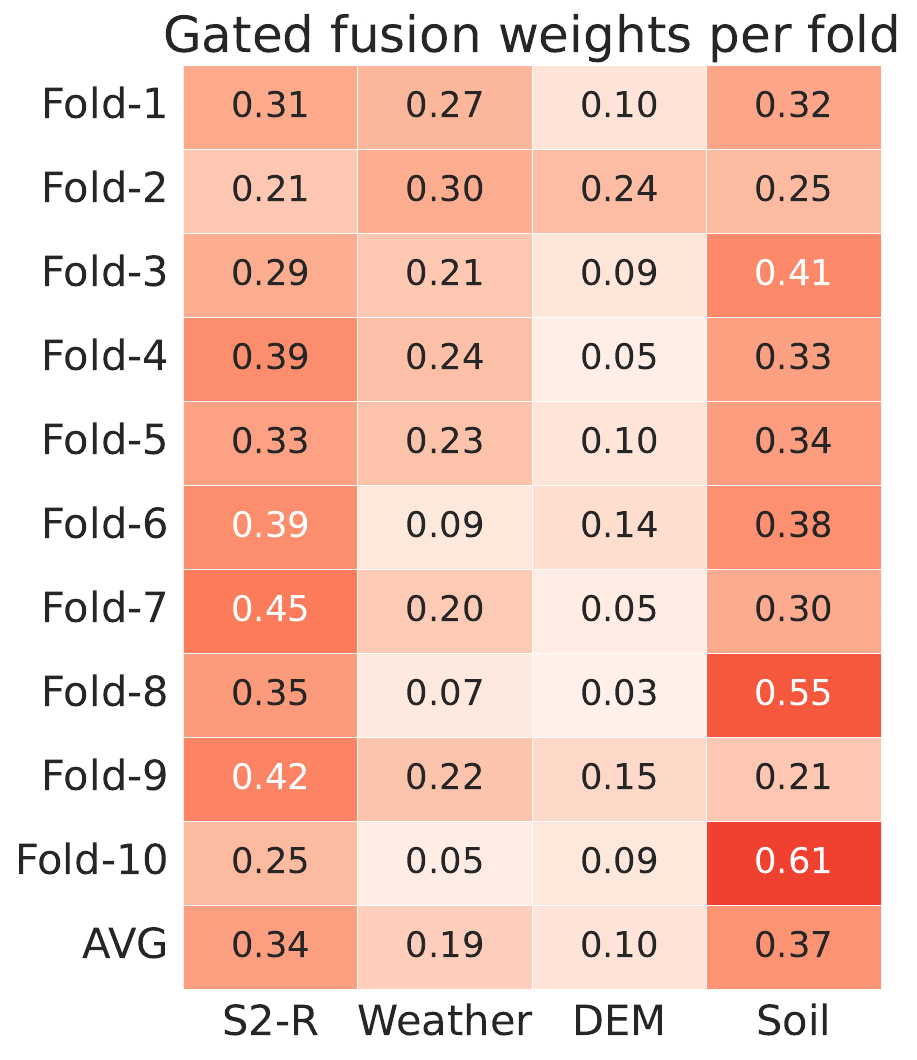}}
\hfill
\subfloat[Fields in GER-W data.\label{fig:attention_weights:d}]{\includegraphics[width=0.25\linewidth]{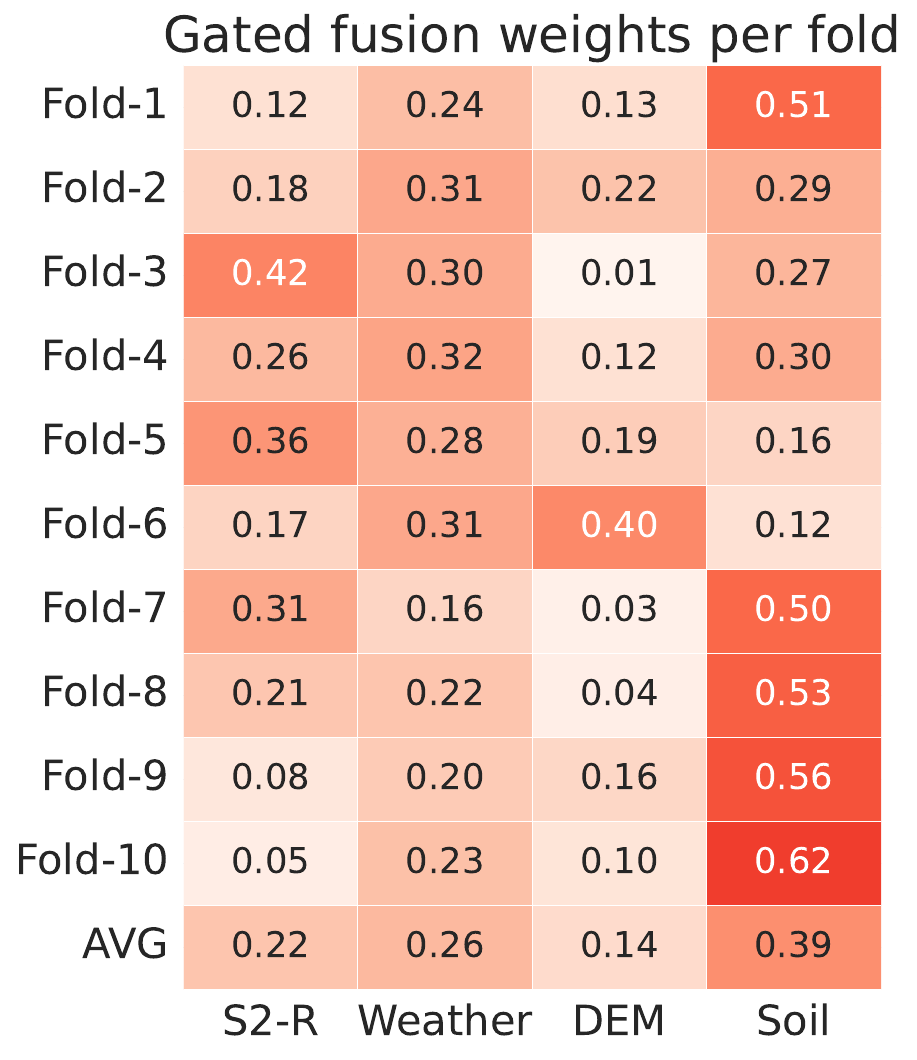}}
\caption{Summary of fusion weights {averaged across all pixels in each fold}. The 10 folds used for evaluation in the cross-validation are displayed for the fields in each dataset. The intensity of the color increases from 0 to 1.}
\label{fig:attention_weights}
\end{figure*} 

%% file: content/conclusion.tex
\section{Final Remarks} \label{sec:conclusion}

In this paper, we have presented a comprehensive case study within the MVL scenario, centered around RS data. The objective is to predict crop yield at sub-field level (yield data at a 10 m spatial resolution) by leveraging a variety of RS sources: \sentsource-based \sent images, \weat, \soil, and \dem maps.
Our proposed model, named Multi-View Gated Fusion (\gatedfusion), comprises two integral components. The first one involves feature-level learning, where dedicated view-encoders learn high-level representations for each input view, effectively handling varying temporal resolutions and data distributions. The second component involves gated fusion, where an attention-like mechanism (the GU) learns weights to fuse multi-view features adaptively, allowing customized aggregations depending on input information.
We have rigorously validated our regression task across fields from four country-crop combinations: soybeans in Argentina and Uruguay, as well as wheat and rapeseed in Germany. Our evaluation employed cross-validation and standard regression metrics, leading to the consistent observation that the proposed \gatedfusion achieves optimal crop yield predictions when all views are used. Such a consistent outcome is relatively uncommon in the RS domain, where factors such as region, data, model, views used and task often shape the effectiveness of the approach \citep{mena2022common}.

To support our findings, we have conducted diverse analyses and visualizations to highlight both the strengths and limitations of our proposed approach. Notably, results at the sub-field level underscored the increased complexity of the crop yield prediction scenario as compared to the field level. This insight prompts future researchers to prioritize the refining of models for this scenario, potentially through image-based mapping.
Furthermore, given the valuable insights provided by the \GFweights, future research will explore this avenue. Investigating disparities in \shortGFweights distribution across different countries and crops holds promise for enhancing model interpretability. Notably, certain researchers \citep{jain-wallace-2019-attention, wiegreffe-pinter-2019-attention} have already challenged the conventional notion of using attention weights as feature importance scores. Their suggested experimental framework to assess the reliability of attention mechanism presents a significant step towards explainability in neural networks.

%% file: content/appendix.tex
\section{Further Visualization}  \label{sec:app_vis}
Some additional charts are included in this section to highlight some patterns in the data.

\subsection{Data Visualization}

\begin{figure*}[t!]
\centering
\subfloat[Fields in ARG-S data.\label{fig:farm_seedtoharv:args}]{\includegraphics[width=0.49\linewidth, angle=0]{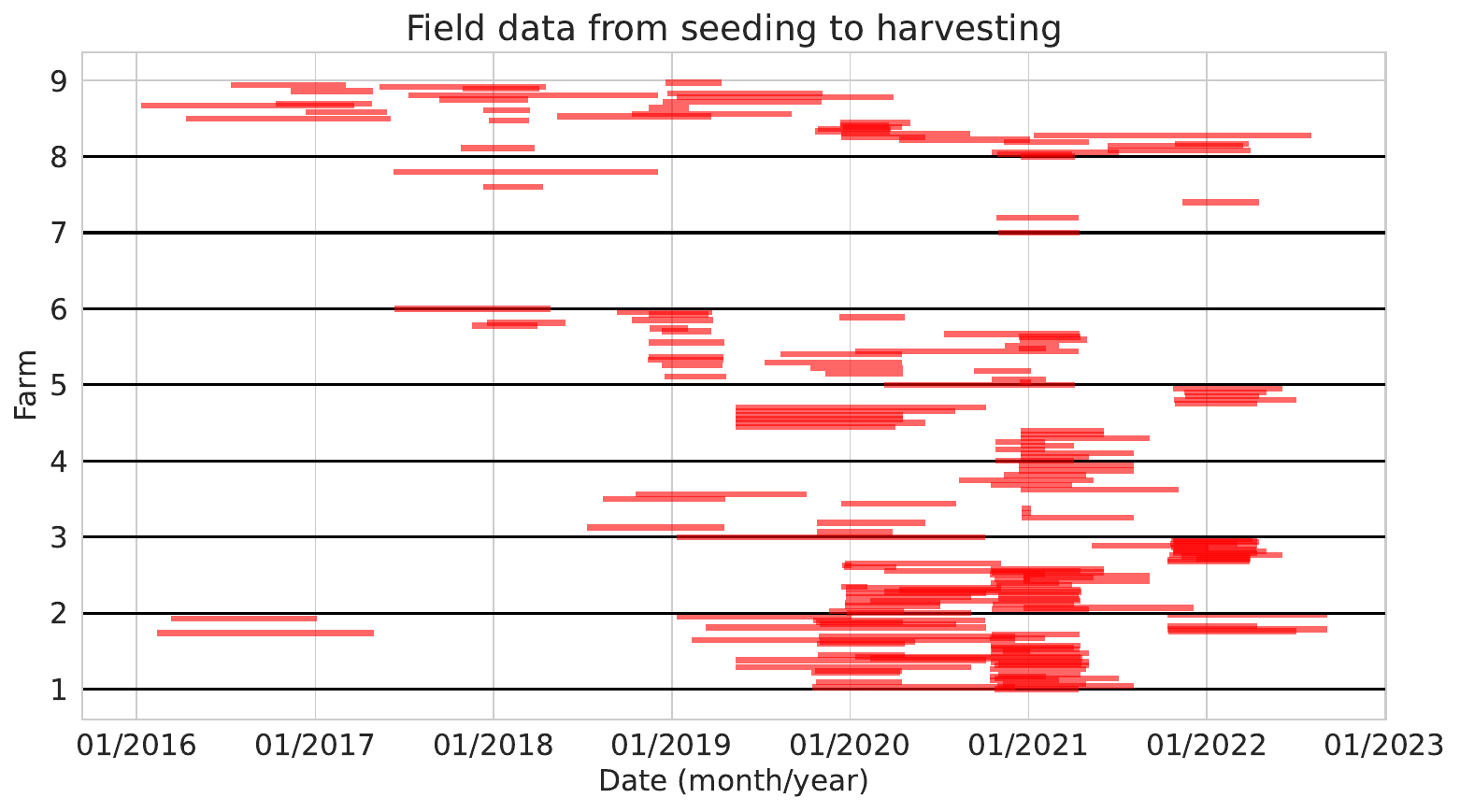}}
\hfill
\subfloat[Fields in URU-S data.\label{fig:farm_seedtoharv:urus}]{\includegraphics[width=0.49\linewidth, angle=0]{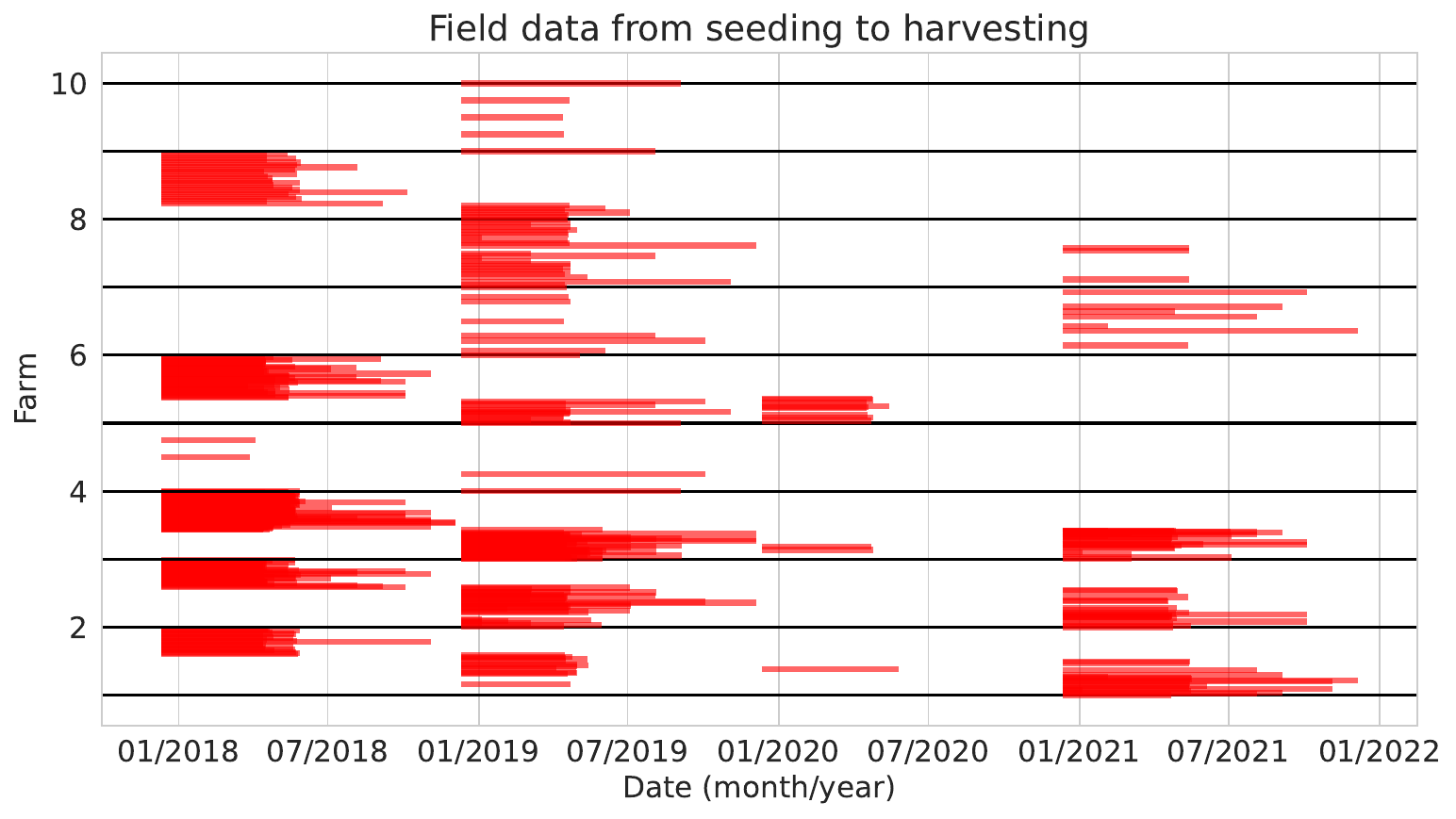}}\\
\subfloat[Fields in GER-R data.\label{fig:farm_seedtoharv:gerr}]{\includegraphics[width=0.49\linewidth, angle=0]{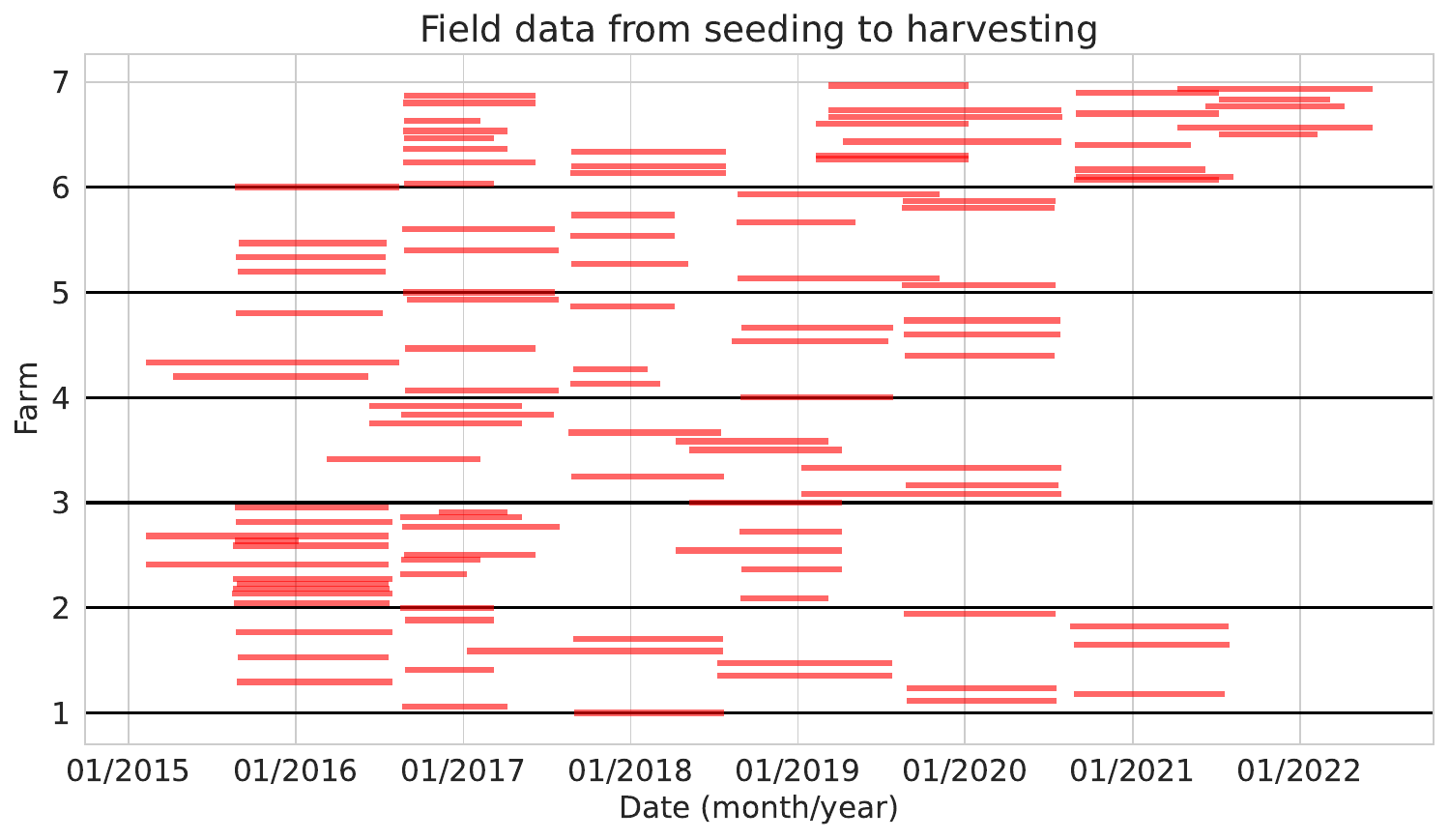}}
\hfill
\subfloat[Fields in GER-W data.\label{fig:farm_seedtoharv:gerw}]{\includegraphics[width=0.49\linewidth,angle=0]{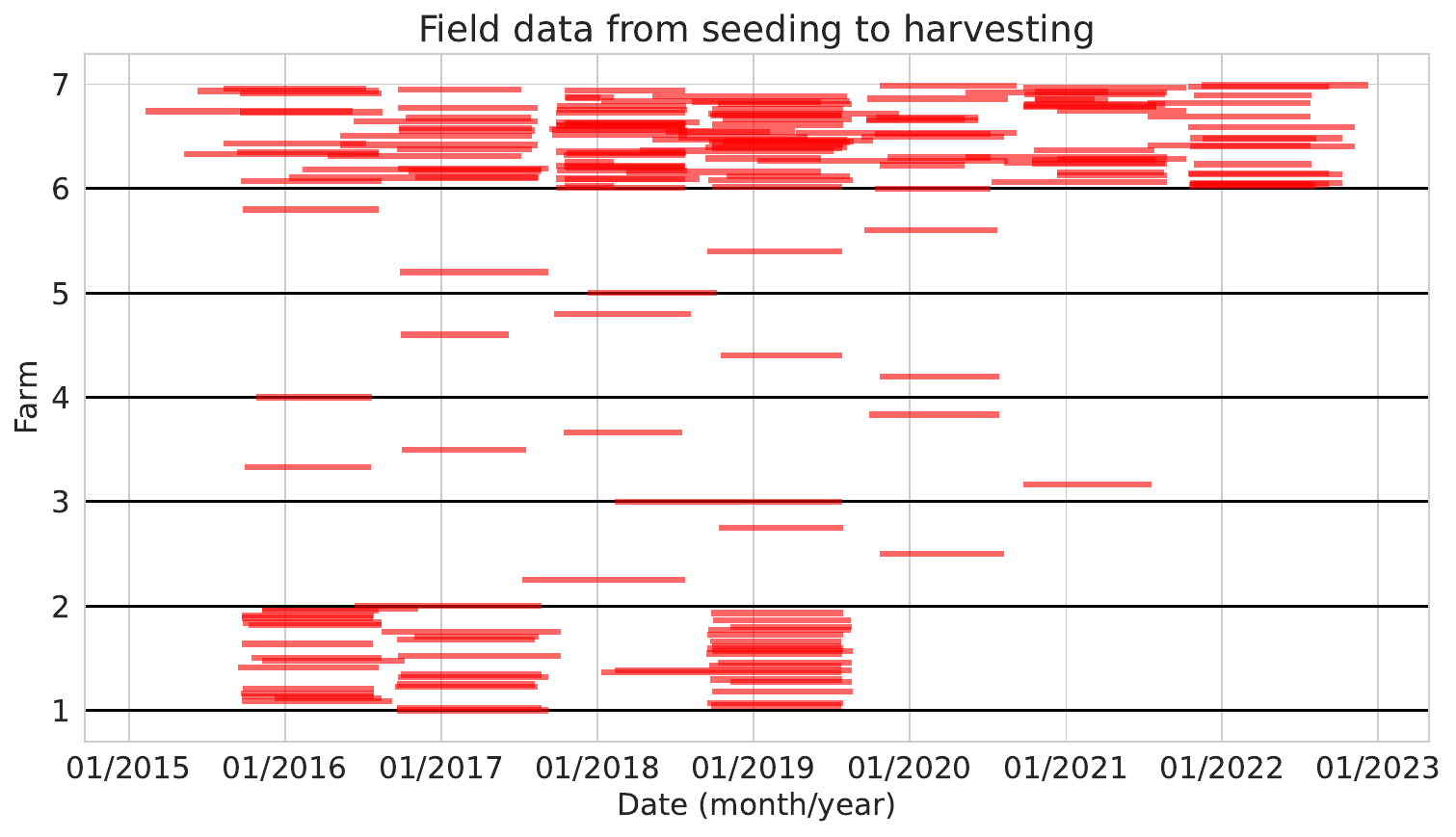}}\\
\caption{Fields from seeding to harvesting grouped by farms. Here, farm represents either a set of fields operated by a farmer or geographically nearby field. There are a different number of fields per year with different growing seasons.}\label{fig:farm_seedtoharv}
\end{figure*}

\begin{figure*}[t!]
\centering
\subfloat[Fields in ARG-S data.\label{fig:coverage_time_year:a}]{\includegraphics[width=0.25\linewidth]{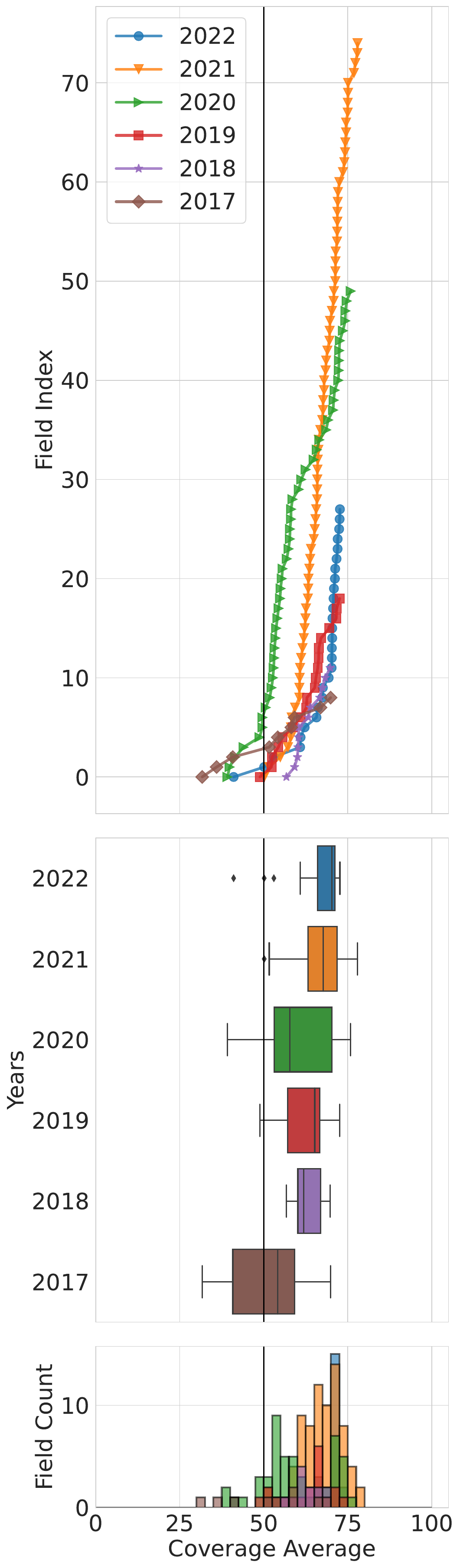}}
\hfill
\subfloat[Fields in URU-S data.\label{fig:coverage_time_year:b}]{\includegraphics[width=0.25\linewidth]{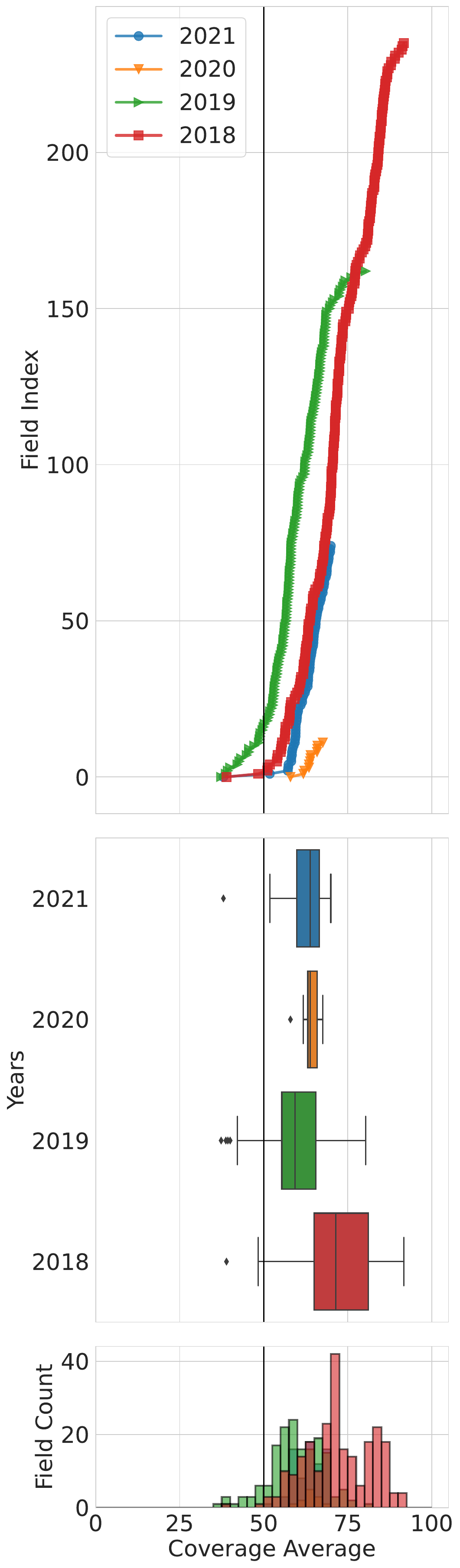}}
\hfill
\subfloat[Fields in GER-R data.\label{fig:coverage_time_year:c}]{\includegraphics[width=0.25\linewidth]{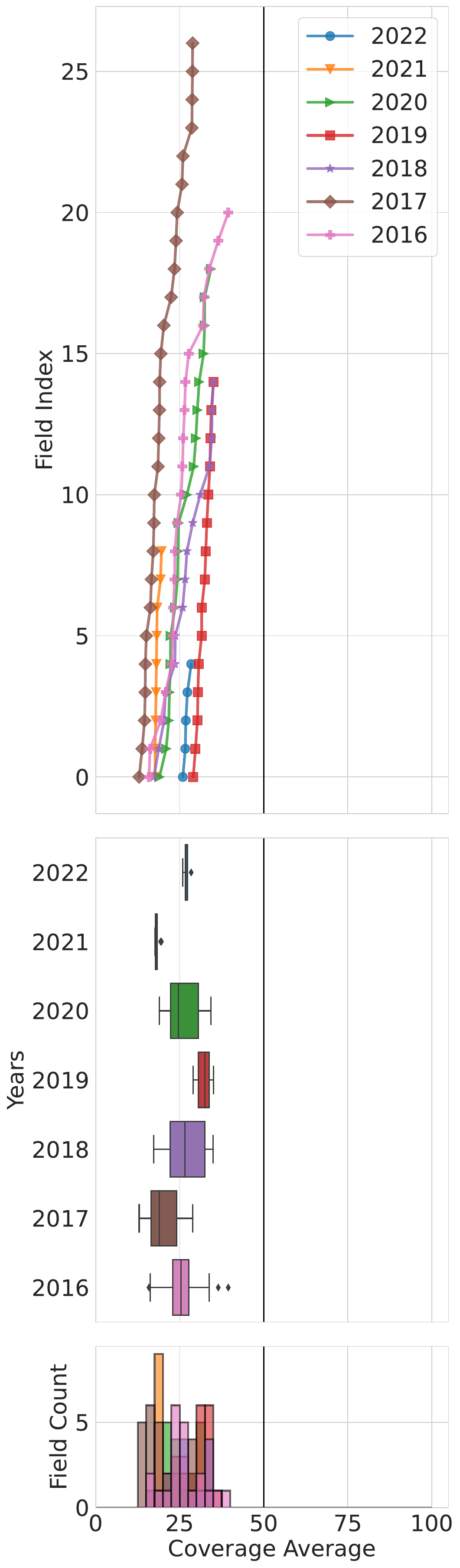}}
\hfill
\subfloat[Fields in GER-W data.\label{fig:coverage_time_year:d}]{\includegraphics[width=0.25\linewidth]{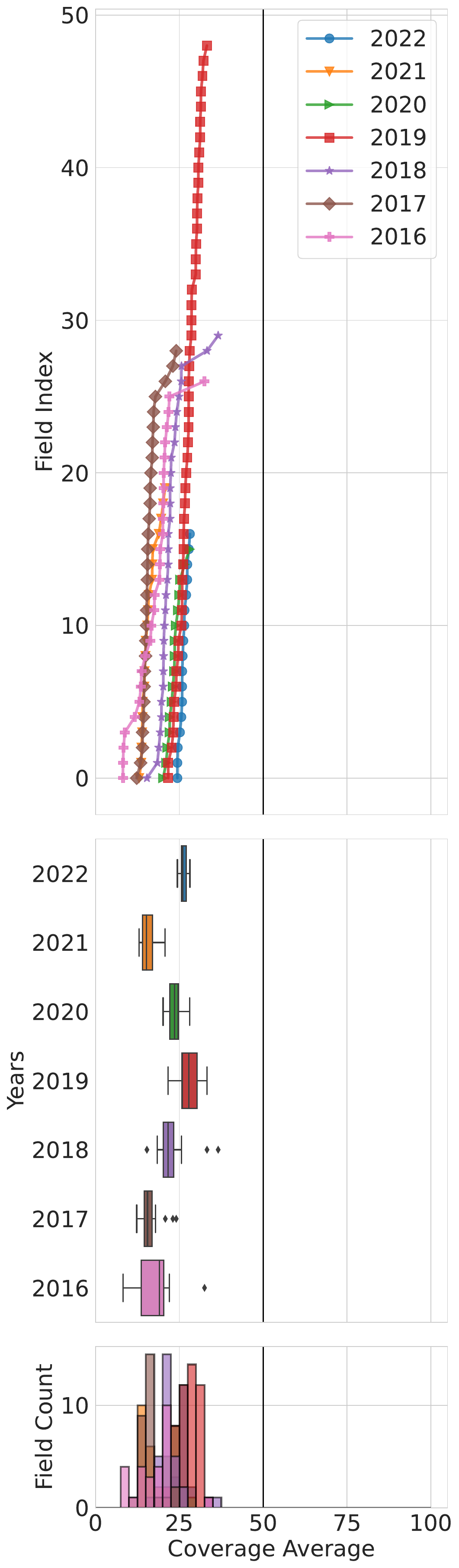}}
\\
\caption{Field spatial coverage for different years. Different sub-figures display different datasets used in this study. The spatial coverage is calculated (and averaged through the growing season) for each field and grouped by year. }\label{fig:coverage_time_year}
\end{figure*}

\paragraph{Input Data} Fig.~\ref{fig:farm_seedtoharv} illustrates the crop growing season in each field (from seeding to harvesting), to illustrate how diverse the collected data is regarding the temporal axis.
Fig.~\ref{fig:coverage_time_year} displays the field spatial coverage in different years of each dataset. This illustrates the variability in the \sentsource-based \sent images not only through regions (as Fig.~\ref{fig:coverage_time} illustrates) but also through years.


\subsection{Model-based Visualization}

\begin{figure*}[t!]
\centering
\subfloat[Predictions of \lstminputfusion model with \sentmonth and \dem input-views.]{\includegraphics[width=0.85\linewidth]{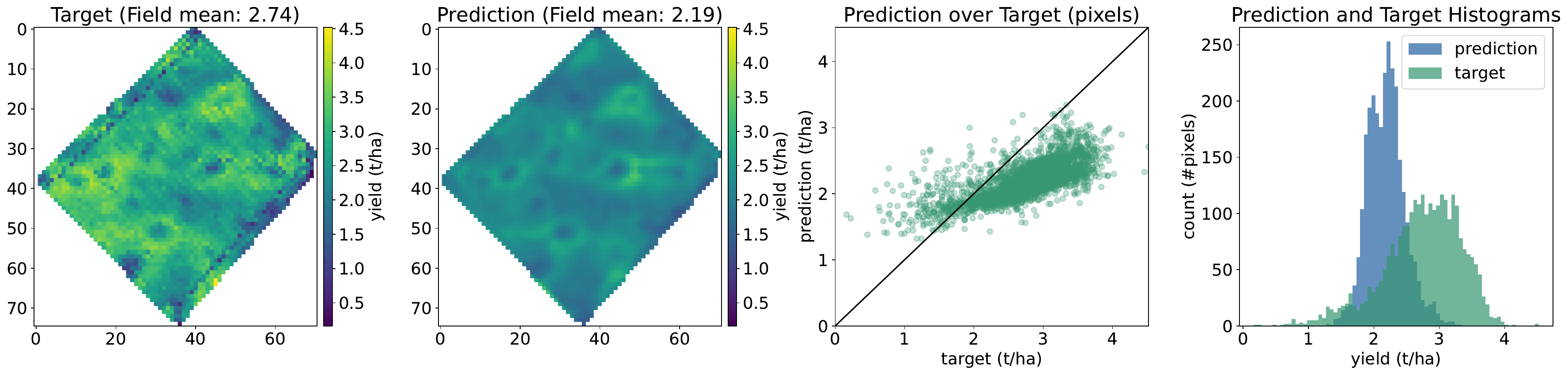}} \\
\subfloat[Predictions of \gatedfusion model with \sentraw, \weat, \dem, and \soil input-views.]{\includegraphics[width=0.85\linewidth]{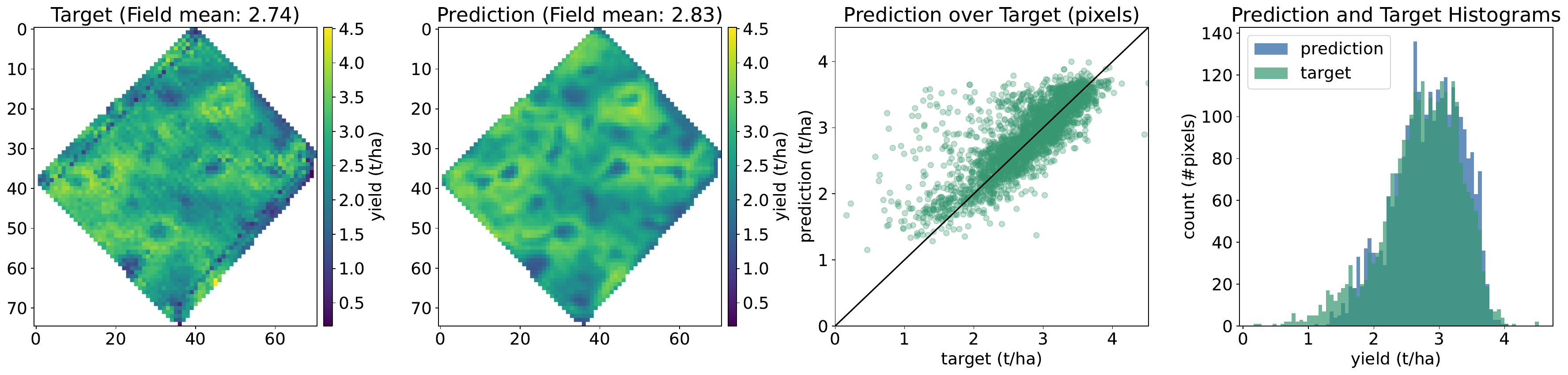}}
\caption{Crop yield prediction map for a sample field in \textbf{ARG-S} data. The columns from left to right are the ground truth yield map, the predicted yield map, prediction and target scatter, and prediction-target (blue-green) distributions.}\label{fig:fieldex:ap1}
\end{figure*}

\begin{figure*}[t!]
\centering
\subfloat[Predictions of \lstminputfusion model with \sentraw, \weat, \dem, and \soil input-views.]{\includegraphics[width=0.85\linewidth]{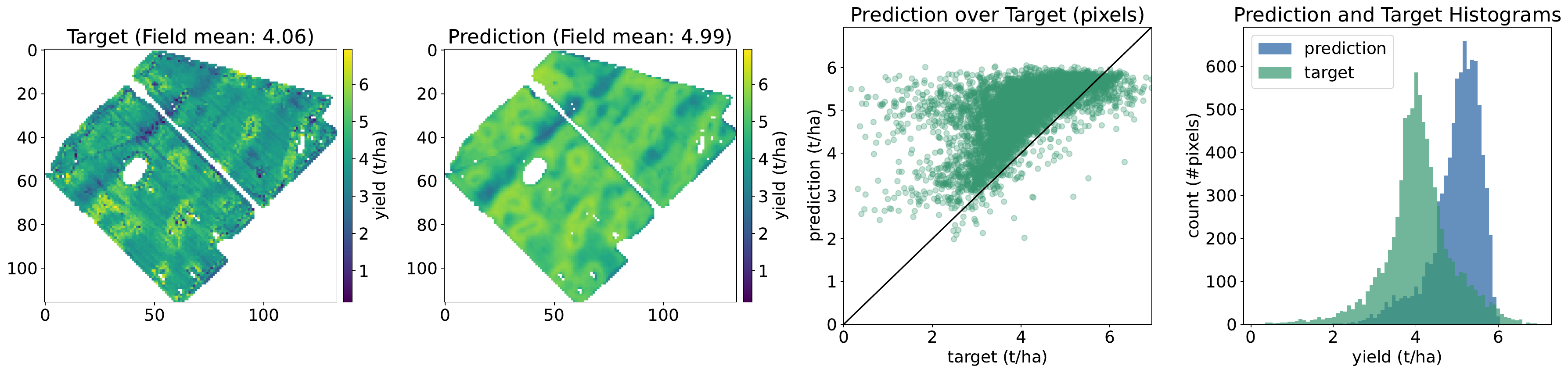}} \\
\subfloat[Predictions of \gatedfusion model with \sentraw, \weat, \dem, and \soil input-views.]{\includegraphics[width=0.85\linewidth]{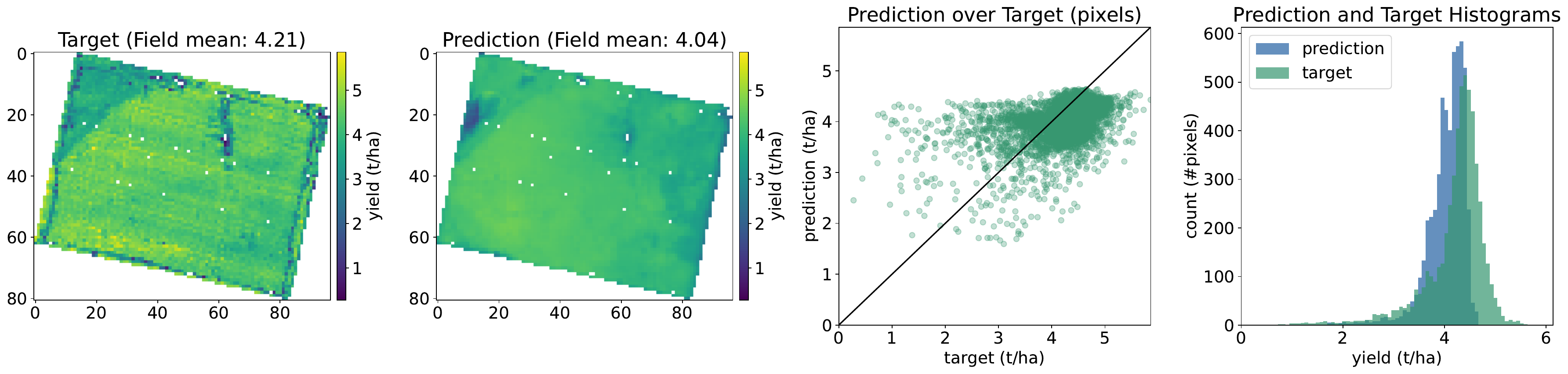}}
\caption{Crop yield prediction map for a sample field in \textbf{URU-S} data. The columns from left to right are the ground truth yield map, the predicted yield map, prediction and target scatter, and prediction-target (blue-green) distributions.}\label{fig:fieldex:ap2}
\end{figure*}

\begin{figure*}[t!]
\centering
\subfloat[Predictions of \lstminputfusion model with \sentmonth and \soil input-views.]{\includegraphics[width=0.85\linewidth]{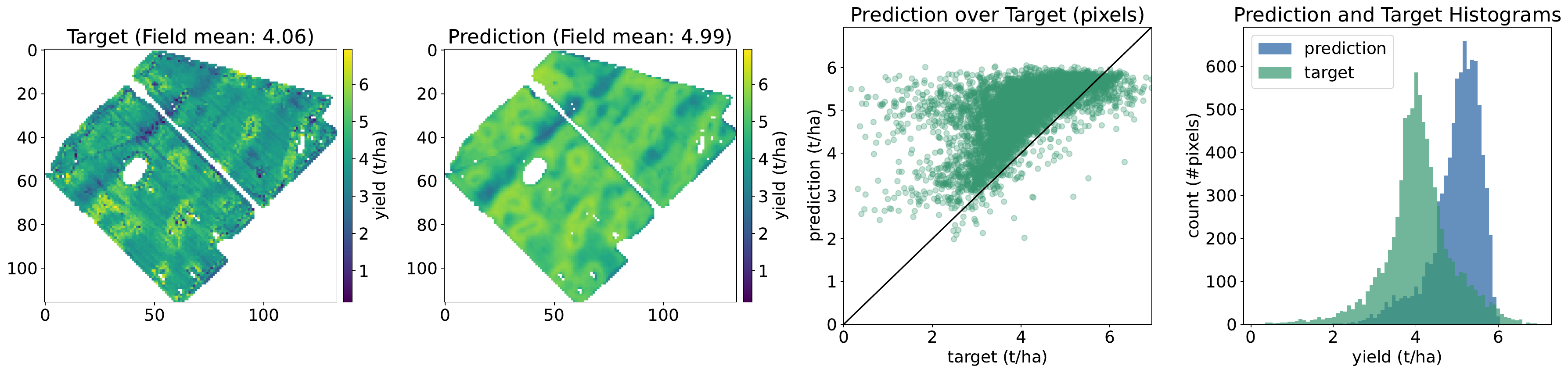}} \\
\subfloat[Predictions of \gatedfusion model with \sentraw, \weat, \dem, and \soil input-views.]{\includegraphics[width=0.85\linewidth]{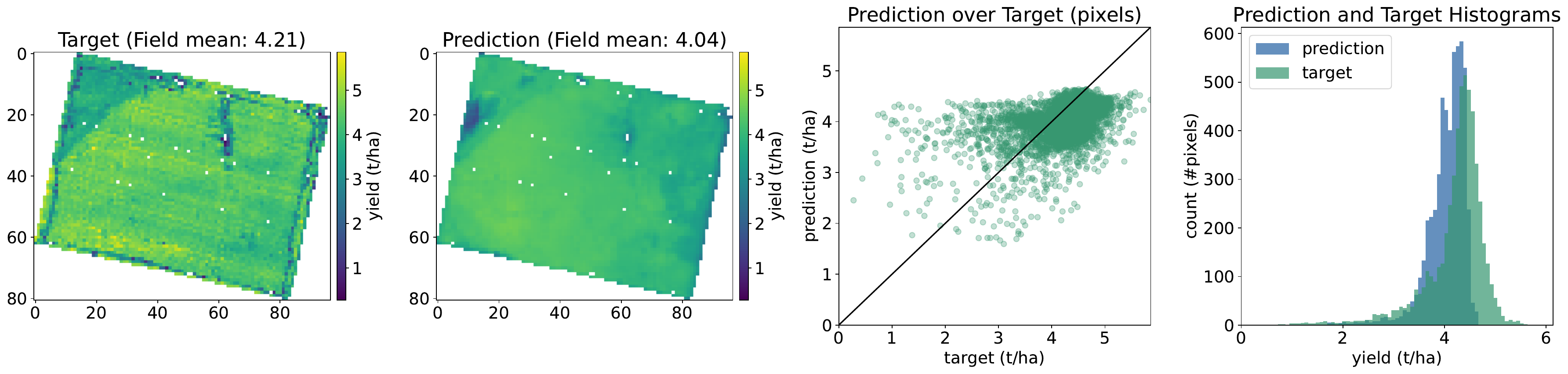}}
\caption{Crop yield prediction map for a sample field in \textbf{GER-R} data. The columns from left to right are the ground truth yield map, the predicted yield map, prediction and target scatter, and prediction-target (blue-green) distributions.}\label{fig:fieldex:ap3}
\end{figure*}

\begin{figure*}[t!]
\centering
\subfloat[Predictions of \lstminputfusion model with \sentraw, \weat, \dem, and \soil input-views.]{\includegraphics[width=0.85\linewidth]{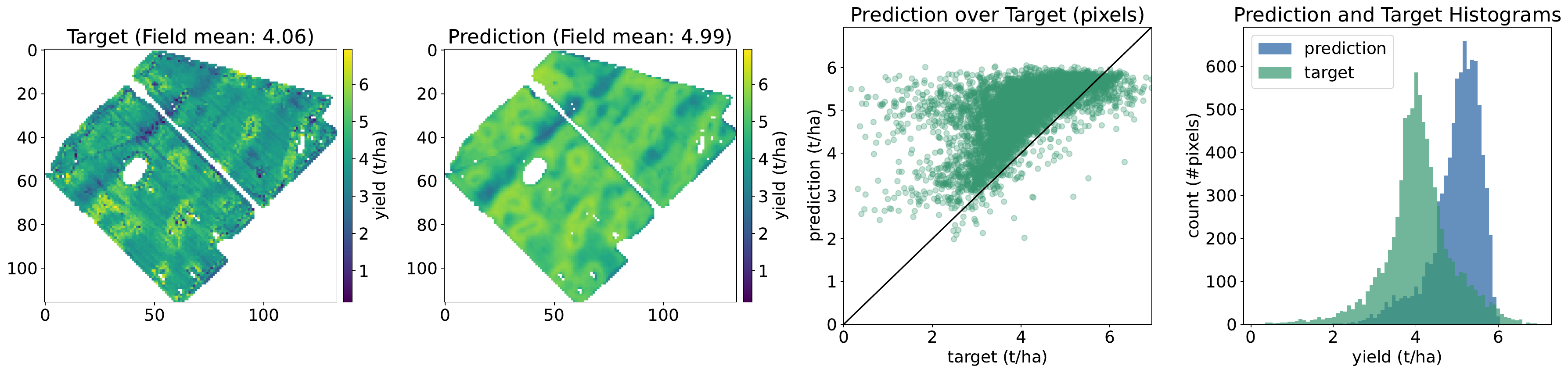}} \\
\subfloat[Predictions of \gatedfusion model with \sentraw, \weat, \dem, and \soil input-views.]{\includegraphics[width=0.85\linewidth]{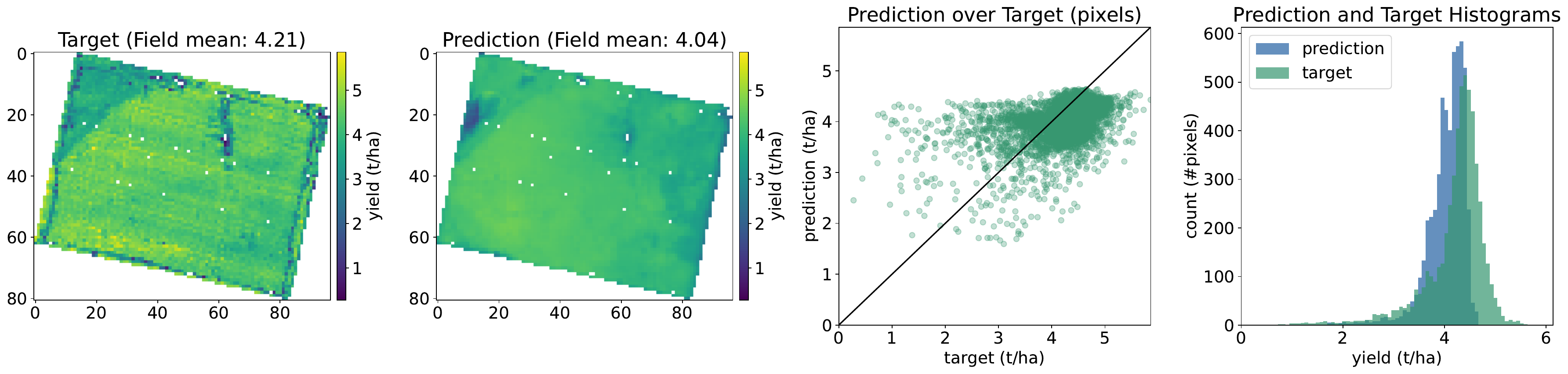}}
\caption{Crop yield prediction map for a sample field in \textbf{GER-W} data. The columns from left to right are the ground truth yield map, the predicted yield map, prediction and target scatter, and prediction-target (blue-green) distributions.}\label{fig:fieldex:ap4}
\end{figure*}

\paragraph{Crop Yield Predictions} With the sub-field predictions of two models (\lstminputfusion and \gatedfusion), the yield map prediction in a field is created.
Fig.~\ref{fig:fieldex:ap1}, \ref{fig:fieldex:ap2}, \ref{fig:fieldex:ap3}, and \ref{fig:fieldex:ap4} show examples of yield map prediction in random fields of the ARG-S, URU-S, GER-R, GER-W data.

\begin{figure*}[t!]
\centering
\subfloat[Fields in ARG-S data.]{\includegraphics[width=0.8\linewidth]{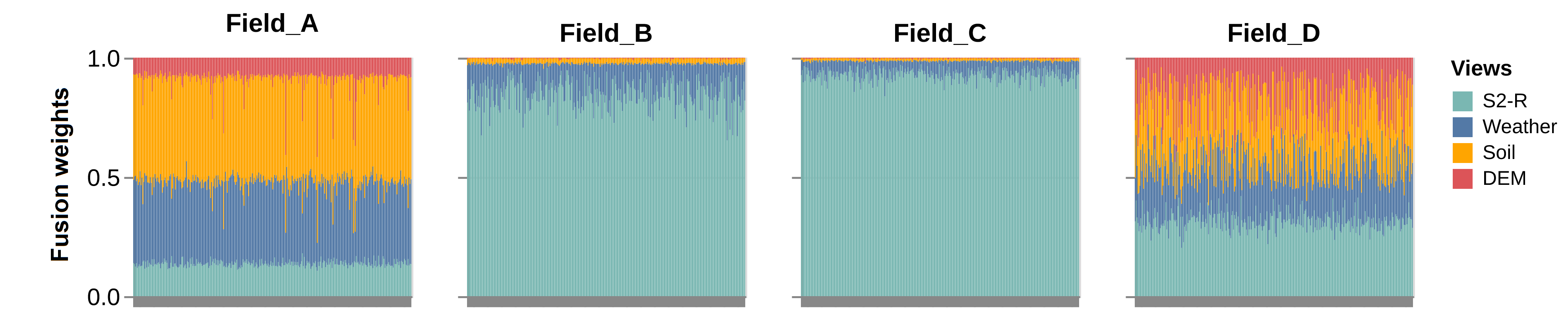}}
\\
\subfloat[Fields in URU-S data.]{\includegraphics[width=0.8\linewidth]{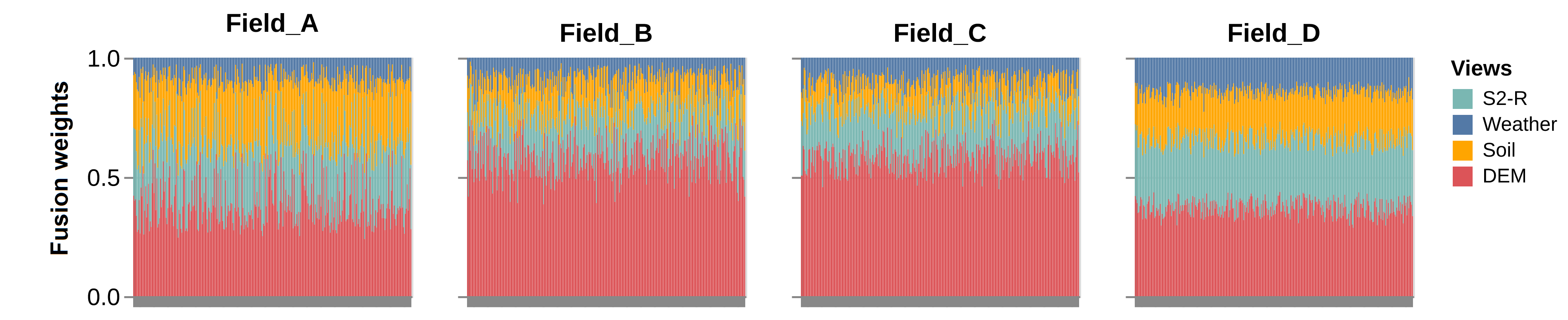}}
\\
\subfloat[Fields in GER-R data.]{\includegraphics[width=0.8\linewidth]{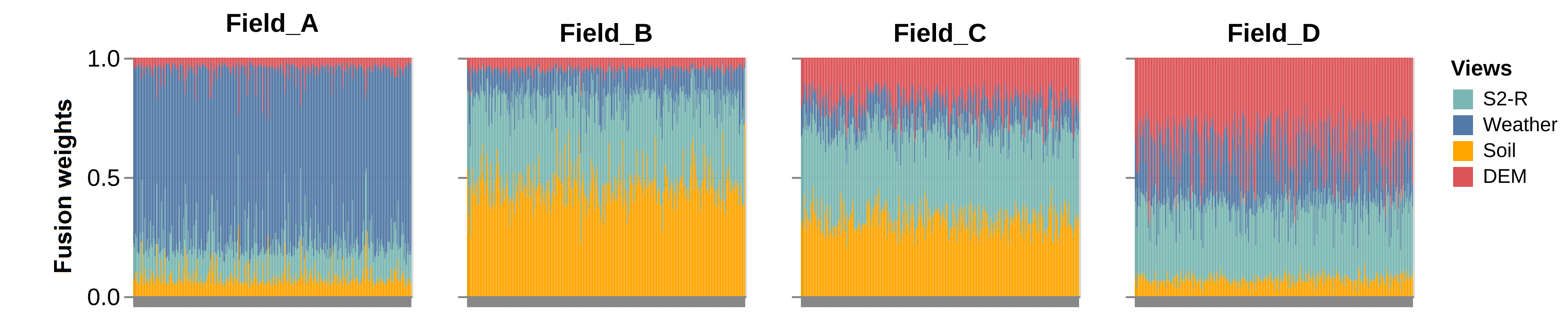}}
\\
\subfloat[Fields in GER-W data.]{\includegraphics[width=0.8\linewidth]{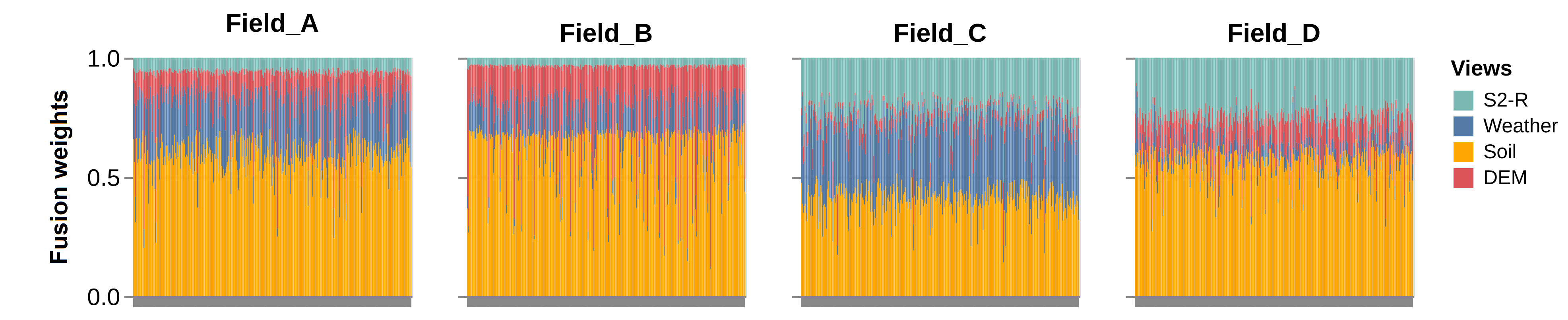}}
\caption{The \GFweights distribution of 300 randomly sampled pixels from 4 random fields of the different datasets used. The x-axis displays the different pixels, while the y-axis the \shortGFweights.
}\label{fig:att_field_level:app}
\end{figure*}

\paragraph{\gatedfusion-based \shortGFweights at sub-field level}\label{sec:app_vis:field_att_weights} Fig.~\ref{fig:att_field_level:app} displays a stacked bar plot on four randomly selected fields per data and 300 pixels within each. Spatially close pixels (pixels within the same field) have similar distribution of weights. These results were consistent across a larger set of fields, which we inspected separately. 

\paragraph{\gatedfusion-based \shortGFweights of a single-fold}\label{sec:app_vis:att_weights}
\input{attention_analysis/att_dataset_level}
To reduce the fold-split variability, we focus on the best performing model across folds (with highest sub-field $R^2$).
Fig.~\ref{fig:att_dataset_level} displays the \shortGFweights distribution for each dataset aggregated at field level (average within a field). This includes fields from the training and validation to get a general overview of each dataset.  
There is an interesting pattern in the Argentina fields with two visually distinguishable groups (Fig.~\ref{fig:att_dataset_level:a}): one group has \sentsource \shortGFweights higher than $0.7$, while the other group has values below $0.3$. 
These plots show the variable and complementary weight that the \gatedfusion model learned to assign depending on the crop-type and region.

\begin{figure*}[t!]
\centering
\subfloat[Fields from GER-R data.]{\includegraphics[width=0.48\linewidth]{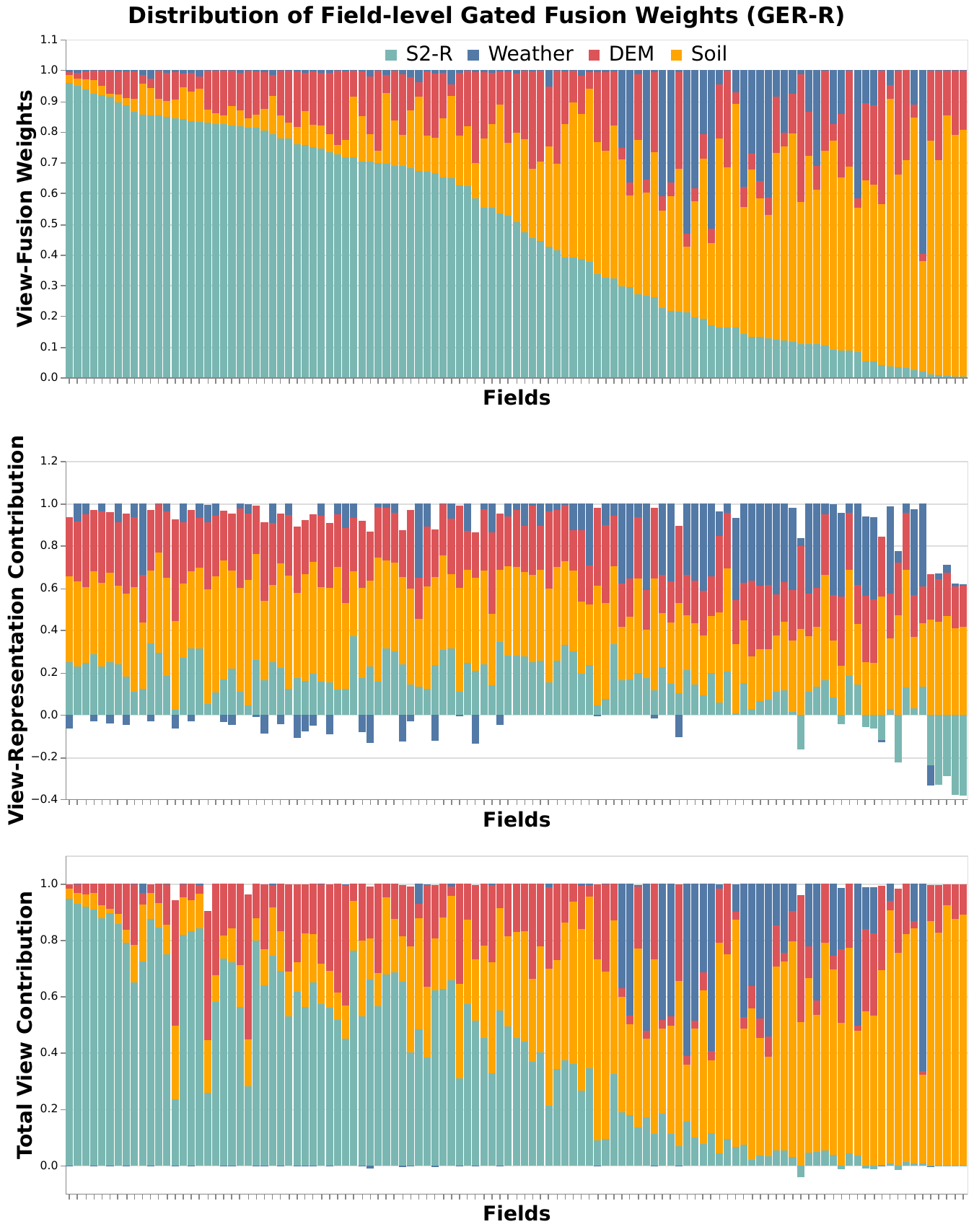}}
\quad
\subfloat[Fields from GER-W data.]{\includegraphics[width=0.48\linewidth]{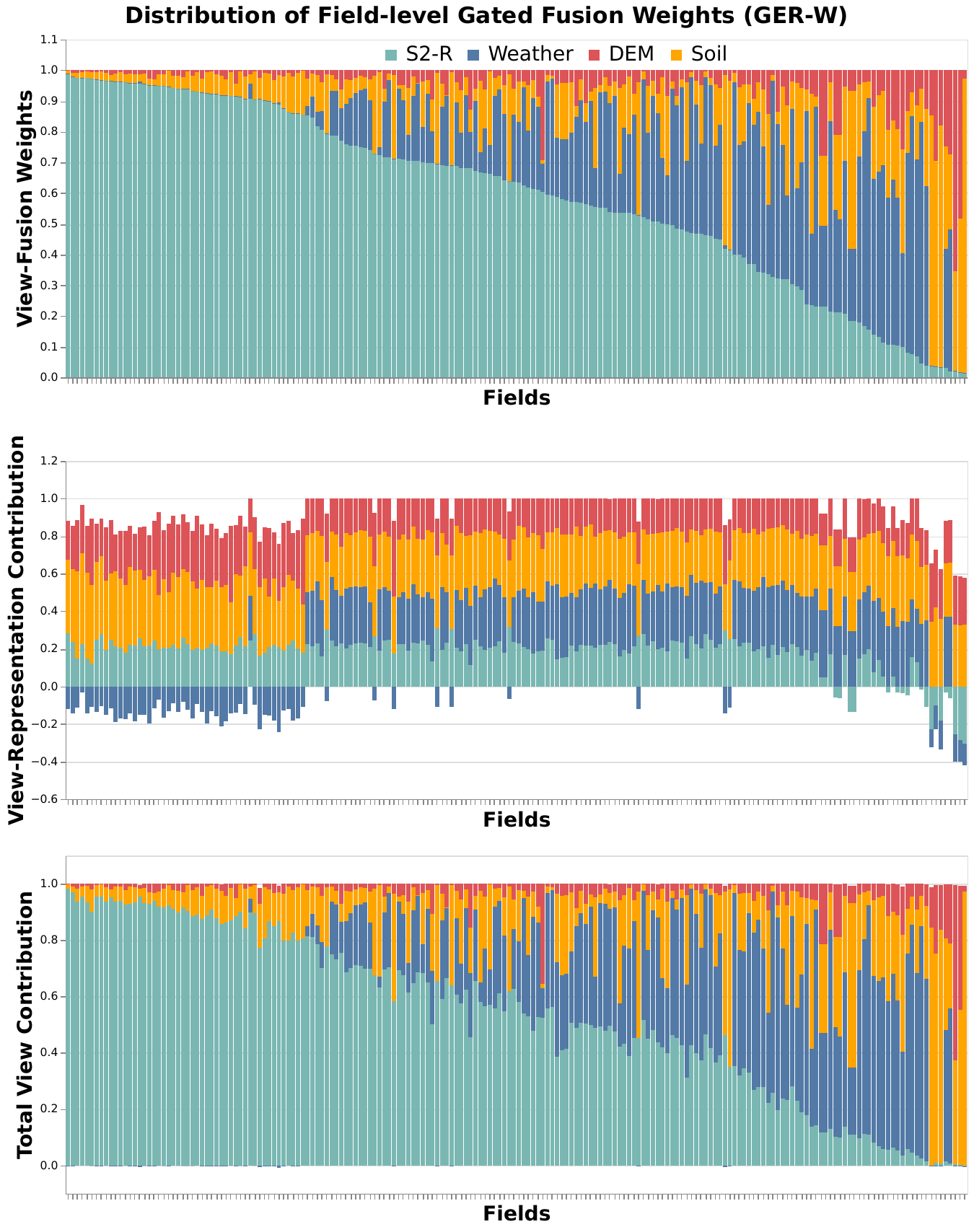}}
\caption{Field level gated fusion weights from the GU ($\alpha_v$) at the top, the contribution before applying the \shortGFweights ($C_v$) at the middle, and the final contribution in the predicted yield ($\alpha_v C_v$) at the bottom. The values are from the \gatedfusion-LR model. The field bars are ordered (from left to right) in descending order of the weight given to the predominant view (\sentraw). The $C_{v}$ and $\alpha_v C_{v}$ are scaled by $1/\sum_v|C_v|$ and $1/\sum_v|\alpha_v C_v|$ respectively for better visualization.}\label{fig:att_weight_comparison:germany}
\end{figure*} 
\paragraph{MVGF-LR-based \shortGFweights analysis}\label{sec:app_vis:mvgf_lr}
In Fig.~\ref{fig:att_weight_comparison:germany} we show the $\alpha_v$, $C_{v}$, and $\alpha_v C_{v}$ (explained in Section~\ref{sec:analysis:mvgf_lr}) aggregated at field level for the German fields (rapeseed and wheat crops).

\section{Extended Results} \label{sec:app_results}
Extension of some results from the main document content are presented in this section.

\paragraph{Per dataset}\label{sec:app_results:perdata} The crop yield prediction performance for different metrics (MAE, MAPE, $R^2$) at field and sub-field level is presented in Table~\ref{tab:performance:arg_s} for ARG-S fields, Table~\ref{tab:performance:uru_s} for URU-S fields, Table~\ref{tab:performance:ger_r} for GER-R fields, and Table~\ref{tab:performance:ger_w} for GER-W fields. Depending on the metric, different models get better or worse performance.
 
\begin{table*}[!t]
    \centering
    \caption{Crop yield prediction performance in the \textbf{ARG-S} fields for different models and combinations of views.  The result with the highest mean and lowest standard deviation is in bold.}\label{tab:performance:arg_s}
    \scriptsize
    \begin{tabularx}{\linewidth}{Cc|CCC|CCC} \hline
          &  & \multicolumn{3}{|c|}{Field}  & \multicolumn{3}{|c}{Sub-Field}\\
         Model & Views & MAE & MAPE & $R^2$ & MAE & MAPE & $R^2$ \\ \hline
         LSTM & \sentmonth & {$0.40 \pm 0.09$} & $11 \pm 3$ & $0.74 \pm 0.12$ & $0.69 \pm0.07$ & $25 \pm4$ & $0.61 \pm 0.11$ \\  
         \lgbminputfusion & \sentmonth, \dem & $0.41 \pm 0.07$ & $12 \pm 3$ & $0.72 \pm 0.14$ & $0.72 \pm 0.07$ & $27 \pm 5$ & $0.58 \pm 0.11$ \\ 
         \lstminputfusion & \sentmonth, \dem & $0.33 \pm 0.08$ & $9 \pm 3$ & $0.82 \pm 0.12$ & $0.65 \pm \highest{0.05}$ & $24 \pm \highest{3}$ & $0.65 \pm 0.08$ \\ \hline
         LSTM & \sentraw & $0.31 \pm \highest{0.07}$ & $9 \pm 3$ & $\highest{0.84} \pm \highest{0.08}$ & $0.62 \pm 0.06$ & $23 \pm \highest{3}$ & $0.67 \pm \highest{0.05}$  \\
        \textbf{\gatedfusion} & \sentraw, \weat, \dem, \soil & $\highest{{0.30}} \pm 0.08$ & $\highest{8} \pm \highest{2}$ & $\highest{0.84} \pm 0.11$ & $\highest{0.61} \pm 0.07$ & $\highest{22} \pm \highest{3}$ & $\highest{0.68} \pm \highest{0.05}$ \\ 
         \hline
    \end{tabularx}
\end{table*}

\begin{table*}[!t]
    \centering
    \caption{Crop yield prediction performance in the \textbf{URU-S } fields for different models and combinations of views. The result with the highest mean and lowest standard deviation is in bold.}\label{tab:performance:uru_s}
    \scriptsize
    \begin{tabularx}{\linewidth}{Cc|CCC|CCC} \hline
         &  & \multicolumn{3}{|c|}{Field}  & \multicolumn{3}{|c}{Sub-Field}\\
         Model & Views & MAE & MAPE & $R^2$ & MAE & MAPE & $R^2$ \\ \hline
         LSTM & \sentmonth & $0.40 \pm 0.06$ & $ 21 \pm 3$ & $0.69 \pm 0.14$ & $0.80 \pm 0.07$ & $100 \pm 18$ & $0.38 \pm 0.08$   \\
         \lgbminputfusion & \sentmonth, \weat, \dem, \soil & $\highest{0.35} \pm 0.06$ & $20 \pm 4$ & $0.77 \pm 0.08$ & $0.78 \pm0.07$ & $102 \pm 17$ & $\highest{0.42} \pm 0.07$ \\
         \lstminputfusion & \sentmonth, \weat, \dem, \soil & $0.37 \pm 0.06$ & $20 \pm \highest{2}$ & $0.74 \pm 0.12$ & $0.78 \pm \highest{0.06}$ & $99 \pm 15$ & $0.41 \pm 0.07$ \\
         \hline 
          LSTM & \sentraw & $\highest{0.35}\pm 0.06$ & $\highest{18} \pm 3$ & $0.77 \pm 0.09$ & $\highest{0.77} \pm \highest{0.06}$ & $95 \pm \highest{14}$ & $0.41\pm \highest{0.06}$ \\
         \textbf{\gatedfusion} & \sentraw, \weat, \dem, \soil & $\highest{0.35} \pm \highest{0.05}$ & $19 \pm 3$ & $\highest{0.78} \pm \highest{0.07}$ & $\highest{0.77} \pm \highest{0.06}$ & $\highest{94} \pm \highest{14}$ & $\highest{0.42} \pm \highest{0.06}$ \\ 
         \hline
    \end{tabularx}
\end{table*}

\begin{table*}[!t]
    \centering
    \caption{Crop yield prediction performance in the \textbf{GER-R} fields for different models and combinations of views. The result with the highest mean and lowest standard deviation is in bold.}\label{tab:performance:ger_r}
    \scriptsize
    \begin{tabularx}{\linewidth}{Cc|CCC|CCC} \hline
          &  & \multicolumn{3}{|c|}{Field}  & \multicolumn{3}{|c}{Sub-Field}\\
         Model & Views & MAE  & MAPE & $R^2$ & MAE & MAPE & $R^2$ \\ \hline
         LSTM & \sentmonth & $0.60 \pm 0.16$ & $18 \pm 11$ & $0.65 \pm 0.18$ & $1.01 \pm 0.21$ & $42 \pm 16$ & $0.35 \pm 0.13$ \\ 
         \lgbminputfusion & \sentmonth, \soil & $0.58 \pm 0.18$ & $18 \pm11$ & $0.69 \pm 0.09$ & $0.94 \pm 0.19$ & $40\pm 14$ & $0.42 \pm 0.08$ \\ 
         \lstminputfusion & \sentmonth, \soil & $0.47 \pm 0.16$ & $15 \pm 9$ & $0.78 \pm \highest{0.09}$ & $0.93 \pm \highest{0.17}$ & $39 \pm 12$ & $0.45 \pm \highest{0.10}$ \\ 
         \hline  
          LSTM & \sentraw & $0.51 \pm 0.17$ & $15 \pm 8$ & $0.77 \pm 0.13$  & $\highest{0.90} \pm 0.18$ & $\highest{36} \pm 12$ & $\highest{0.46} \pm 0.11$ \\ 
         \textbf{\gatedfusion} &\sentraw, \weat, \dem, \soil & $\highest{0.43} \pm \highest{0.15}$ & $\highest{13} \pm \highest{7}$ & $\highest{0.80} \pm 0.13$ & $\highest{0.90} \pm \highest{0.17}$ & $\highest{36} \pm \highest{9}$ & $\highest{0.46} \pm 0.15$ \\ 
         \hline
    \end{tabularx}
\end{table*}

\begin{table*}[!t]
    \centering
    \caption{Crop yield prediction performance in the \textbf{GER-W} fields for different models and combinations of views. The result with the highest mean and lowest standard deviation is in bold.}\label{tab:performance:ger_w}
    \scriptsize
    \begin{tabularx}{\linewidth}{Cc|CCC|CCC} \hline
         &  & \multicolumn{3}{|c|}{Field}  & \multicolumn{3}{|c}{Sub-Field}\\
         Model & Views & MAE & MAPE & $R^2$ & MAE & MAPE & $R^2$ \\ \hline
         LSTM & \sentmonth & $0.91 \pm 0.23$ & $10 \pm 4$ & $0.60 \pm 0.25$ & $1.79 \pm 0.22$ & $30 \pm 5$ & $0.32 \pm 0.09$  \\ 
         \lgbminputfusion & \sentmonth, \weat, \dem, \soil & $0.86 \pm 0.22$& $10 \pm 2$& $0.68 \pm 0.12$ & $1.76 \pm 0.24$& $30 \pm 4$& $0.37\pm 0.10$\\ 
         \lstminputfusion & \sentmonth, \weat, \dem, \soil & $0.84 \pm 0.19$ & $9 \pm 3$ & $0.68 \pm 0.28$ & $1.71 \pm 0.28$ & $29 \pm 6$ & $0.37 \pm 0.12$ \\ 
         \hline  
          LSTM & \sentraw & $0.72 \pm 0.24$ & $8 \pm3 $ & $0.72 \pm 0.11$ & $1.59 \pm \highest{0.26}$ & $\highest{27} \pm \highest{4}$ & $0.41 \pm \highest{0.08}$  \\ 
         \textbf{\gatedfusion} & \sentraw, \weat, \dem, \soil & $\highest{0.64} \pm \highest{0.16}$ & $\highest{7} \pm \highest{2}$ & $\highest{0.80} \pm \highest{0.09}$ & $\highest{1.58} \pm \highest{0.26}$ & $\highest{27} \pm 5$ & $\highest{0.44} \pm 0.10$  \\
         \hline
    \end{tabularx}
\end{table*}

\paragraph{LOYO experiment} \label{sec:app_results:loyo} The different regression metrics on the leave-one-year-out (LOYO) cross-validation for the ARG-S fields is displayed in Table~\ref{tab:performance:arg_s:yearly}.

\begin{table*}[!t]
    \centering
    \caption{Crop yield prediction performance on \textbf{ARG-S} fields at different evaluation years. The best input views configuration is used, for \lstminputfusion is \sentmonth with \dem, while for \gatedfusion are all views (\sentraw, \weat, \dem, \soil).}\label{tab:performance:arg_s:yearly}
    \footnotesize
    \begin{tabularx}{\linewidth}{CC|CCC|CCC} \hline
         &  & \multicolumn{3}{c|}{Field}  & \multicolumn{3}{c}{Sub-Field}\\
         Year & Model & MAE & MAPE & $R^2$ & MAE & MAPE & $R^2$ \\ \hline
         \multirow{2}{*}{2017} & \lstminputfusion & $\highest{0.29}$ & $\highest{7}$ & $\highest{0.78}$ & $\highest{0.68}$ & $\highest{20}$ & $\highest{0.32}$  \\   
         & \textbf{\gatedfusion} & $0.63$ & $14$ & $-0.32$ & $0.89$ & $24$ & $-0.14$ \\  \hline 
         \multirow{2}{*}{2018} & \lstminputfusion & $0.91$ & $30$ & $-0.16$ & $0.95$ & $36$ & $0.17$  \\   
         & \textbf{\gatedfusion} &  $\highest{0.65}$ & $\highest{21}$ & $\highest{0.46}$ & $\highest{0.80}$ & $\highest{29}$ & $\highest{0.44}$ \\  \hline 
         \multirow{2}{*}{2019} & \lstminputfusion & $0.59$ & $15$ & $-0.77$ & $0.85$ & $33$ & $0.13$  \\   
         & \textbf{\gatedfusion} & $\highest{0.41}$ & $\highest{10}$ & $\highest{0.09}$ & $\highest{0.67}$ & $\highest{23}$ & $\highest{0.41}$ \\  \hline 
         \multirow{2}{*}{2020} & \lstminputfusion & $0.49$ & $13$ & $0.24$ & $0.76$ & $27$ & $0.29$  \\   
         & \textbf{\gatedfusion} & $\highest{0.41}$ & $\highest{11}$ & $\highest{0.54}$ & $\highest{0.67}$ & $\highest{23}$ & $\highest{0.45}$ \\  \hline 
         \multirow{2}{*}{2021} & \lstminputfusion & $\highest{0.48}$ & $\highest{16}$ & $\highest{0.62}$ & $0.79$ & $34$ & $0.57$  \\   
         & \textbf{\gatedfusion} & $0.51$ & $17$ & $0.59$ & $\highest{0.73}$ & $\highest{30}$ & $\highest{0.63}$ \\  \hline 
         \multirow{2}{*}{2022} & \lstminputfusion &  $0.85$ & $16$ & $-0.72$ & $1.11$ & $26$ & $0.11$ \\   
         & \textbf{\gatedfusion} & $\highest{0.50}$ & $\highest{10}$ & $\highest{0.26}$ & $\highest{0.85}$ & $\highest{22}$ & $\highest{0.37}$ \\ 
         \hline
    \end{tabularx}
\end{table*}

%% file: attention_analysis/att_dataset_level.tex
\begin{figure*}[t!]
\centering
    \subfloat[Fields in ARG-S data. \label{fig:att_dataset_level:a}]{\includegraphics[width=0.48\linewidth]{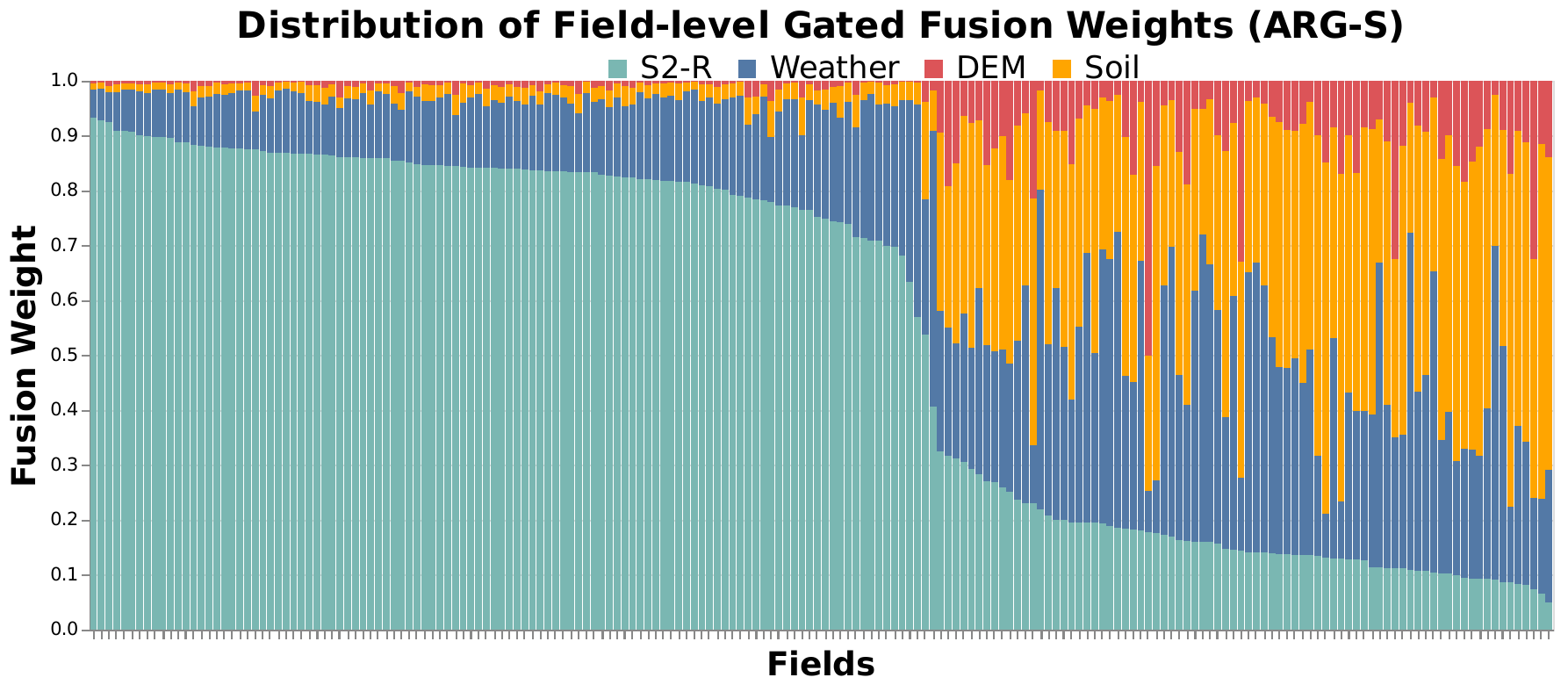}} 
\quad
    \subfloat[200 random fields from the URU-S data. \label{fig:att_dataset_level:b}]{\includegraphics[width=0.48\linewidth]{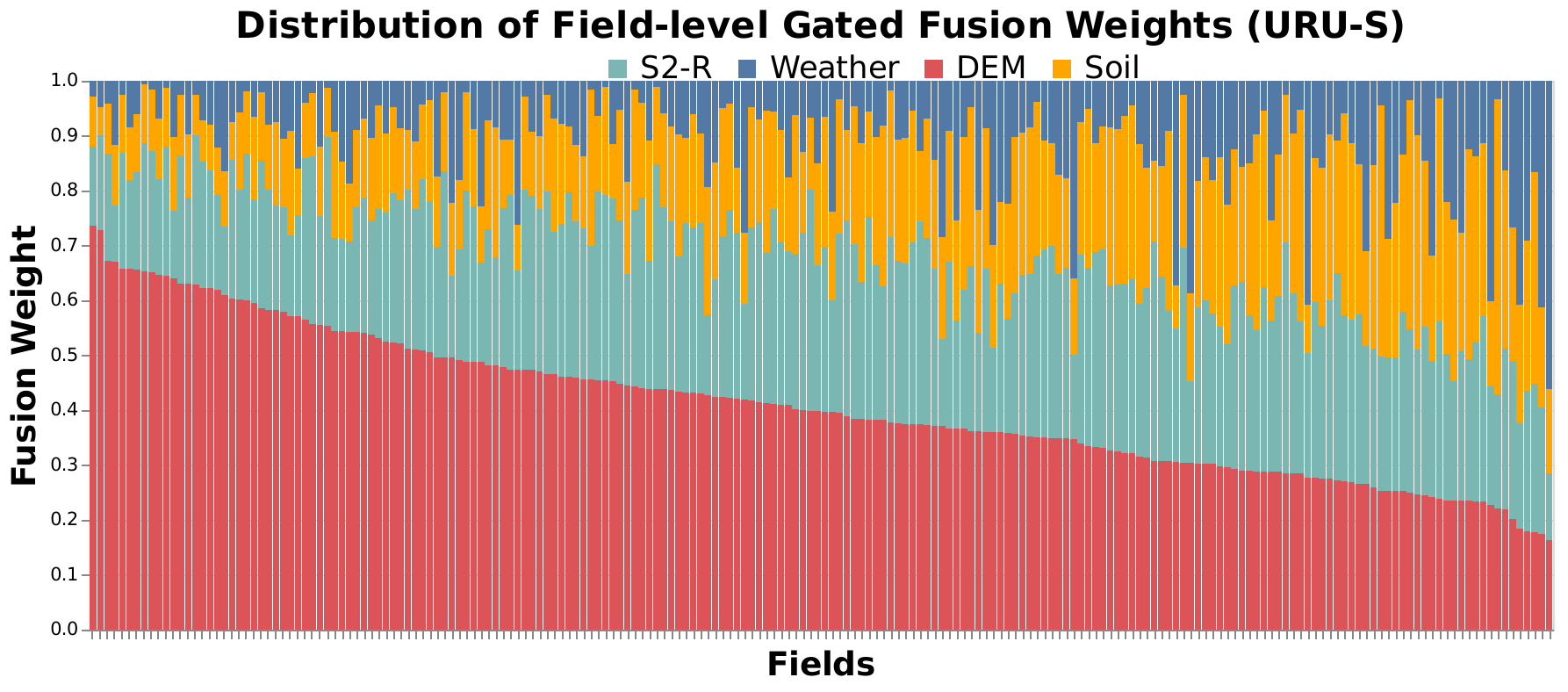}} 
\\
    \subfloat[Fields in GER-R data. \label{fig:att_dataset_level:c}]{\includegraphics[width=0.48\linewidth]{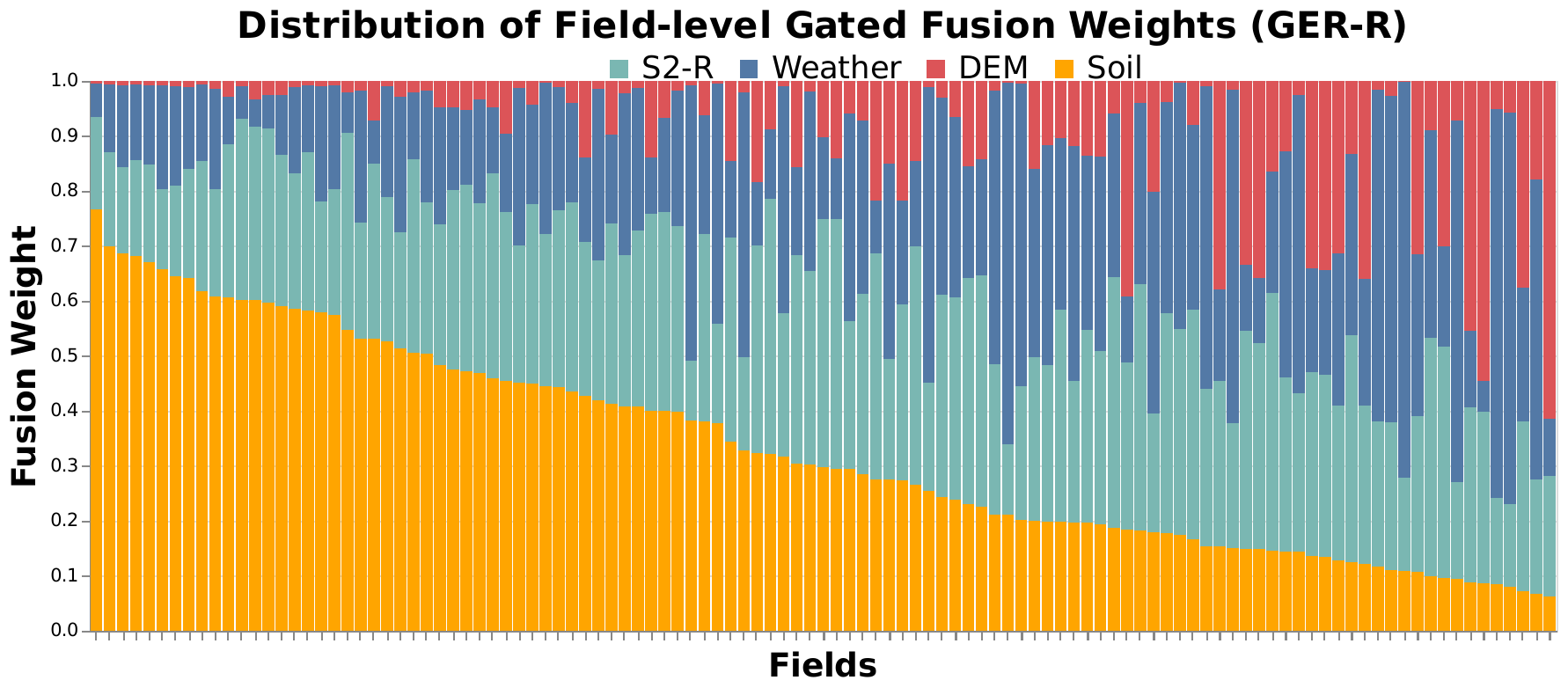}} 
\quad
    \subfloat[Fields in GER-W data. \label{fig:att_dataset_level:d}]{\includegraphics[width=0.48\linewidth]{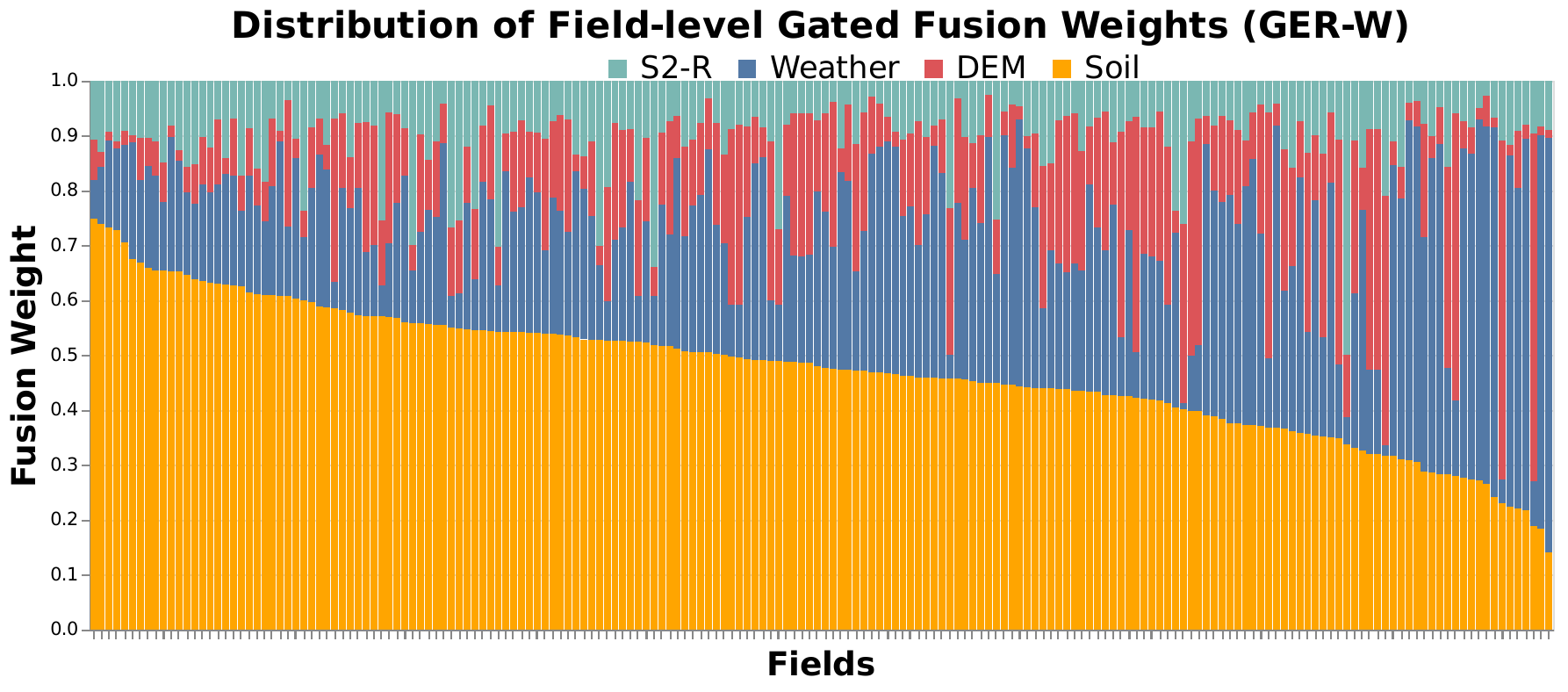}} \\
\caption{{Field level} fusion weights of the \gatedfusion model, where each bar represents one field.
The field bars are ordered (from left to right) in descending order of the weight given to the predominant view in each dataset.}
\label{fig:att_dataset_level}
\end{figure*} 